\documentclass[pdflatex,sn-mathphys-num,Numbered]{sn-jnl} 
\usepackage{fontspec}

\RequirePackage{xeCJK}

\usepackage{graphicx}%
\usepackage{multirow}%
\usepackage{amsmath,amssymb,amsfonts}%
\usepackage{amsthm}
\usepackage{mathrsfs}%
\usepackage[title]{appendix}%
\usepackage{xcolor}%
\usepackage{textcomp}%
\usepackage{manyfoot}%
\usepackage{booktabs}%
\usepackage{algorithm}%
\usepackage{algorithmicx}%
\usepackage{algpseudocode}%
\usepackage{listings}%

\usepackage{longtable}
\usepackage{array}

\usepackage{tabularx}
\usepackage{hyperref}  
\usepackage{xurl}

\usepackage[most]{tcolorbox}
\usepackage{fvextra}
\usepackage{xltabular}
\usepackage{seqsplit}

\tcbuselibrary{breakable}
\definecolor{terminalbg}{RGB}{245,246,248}
\definecolor{terminalframe}{RGB}{55,65,81}
\definecolor{terminaltitle}{RGB}{31,41,55}
\definecolor{terminaltext}{RGB}{33,37,41}
\definecolor{terminalaccent}{RGB}{107,114,128}

\lstdefinestyle{terminalpromptstyle}{
  basicstyle=\ttfamily\footnotesize\color{terminaltext},
  backgroundcolor=\color{terminalbg},
  breaklines=true,
  breakatwhitespace=false,
  columns=fullflexible,
  keepspaces=true,
  showstringspaces=false,
  frame=none,
  xleftmargin=0.6em,
  xrightmargin=0.6em,
  aboveskip=0.4em,
  belowskip=0.2em
}

\newtcolorbox{agentstep}[1]{
  enhanced,
  breakable,
  colback=white,
  colframe=black!70,
  colbacktitle=black!85,
  coltitle=white,
  fonttitle=\bfseries,
  title=#1,
  boxrule=0.6pt,
  arc=3pt,
  left=6pt,
  right=6pt,
  top=6pt,
  bottom=6pt
}

\newtcolorbox{humanprompt}[1][]{
  enhanced,
  breakable,
  colback=blue!2!white,
  colframe=blue!75!black,
  colbacktitle=blue!10!white,
  coltitle=black,
  fonttitle=\bfseries\large,
  title={Human Researcher},
  boxrule=0.8pt,
  arc=3pt,
  left=6pt, right=6pt, top=6pt, bottom=6pt,
  before skip=1.5em,
  after skip=1em,
  titlerule=0.5pt,
  #1
}

\newcommand{\newcheck}{\checkmark}
\newcommand{\newcross}{\ensuremath{\times}}
\newcommand{\y}{\newcheck}
\newcommand{\n}{\newcross}

\usepackage{multibib}
\newcites{methods}{References}

\usepackage{mdframed}
\newcounter{bxcomm}
\definecolor{aqua}{rgb}{0.00,0.67,0.80}

\newcounter{todocomm}

\newcommand{\code}[1]{\texttt{#1}}
\newcommand{\eg}{\textit{e}.\textit{g}.}

\newcommand{\vf}{VenusFactory2}

\theoremstyle{thmstyleone}%
\theoremstyle{thmstyletwo}%

\theoremstyle{thmstylethree}%

\begin{document}

\title[Venus]{Self-evolving AI agents for protein discovery and directed evolution}

\author[1,2,3,4]{\fnm{Yang} \sur{Tan}}
\equalcont{These authors contributed equally to this work.}

\author[1,3,4]{\fnm{Lingrong} \sur{Zhang}}
\equalcont{These authors contributed equally to this work.}

\author[1,4]{\fnm{Mingchen} \sur{Li}}
\equalcont{These authors contributed equally to this work.}

\author[1,4]{\fnm{Yuanxi} \sur{Yu}}

\author[1,4]{\fnm{Bozitao} \sur{Zhong}}

\author*[1,4]{\fnm{Bingxin} \sur{Zhou}}\email{bingxin.zhou@sjtu.edu.cn}

\author*[2]{\fnm{Nanqing} \sur{Dong}}\email{nanqing.dong@sii.edu.cn}

\author*[1,3,4]{\fnm{Liang} \sur{Hong}}\email{hongl3liang@sjtu.edu.cn}

\affil[1]{\orgdiv{Shanghai National Center for Applied Mathematics (SJTU Center) \& Institute of Natural Sciences}, \orgname{Shanghai Jiao Tong University}, \orgaddress{\city{Shanghai}, \postcode{200240}, \country{China}}}

\affil[2]{\orgname{Shanghai Innovation Institute}, \orgaddress{\city{Shanghai}, \postcode{200231}, \country{China}}}

\affil[3]{\orgdiv{School of Computer Science}, \orgname{Shanghai Jiao Tong University}, \orgaddress{\city{Shanghai}, \postcode{200240}, \country{China}}}

\affil[4]{\orgdiv{Zhangjiang Institute for Advanced Study}, \orgname{Shanghai Jiao Tong University}, \orgaddress{\city{Shanghai}, \postcode{201203}, \country{China}}}



\abstract{
Protein scientific discovery is bottlenecked by the manual orchestration of information and algorithms, while general agents are insufficient in complex domain projects. \vf~provides an autonomous framework that shifts from static tool usage to dynamic workflow synthesis via a self-evolving multi-agent infrastructure to address protein-related demands. It outperforms a set of well-known agents on the VenusAgentEval benchmark, and autonomously organizes the discovery and optimization of proteins from a single natural language prompt.
}

\maketitle

\begin{figure*}[t]
    \centering
    \includegraphics[width=\textwidth]{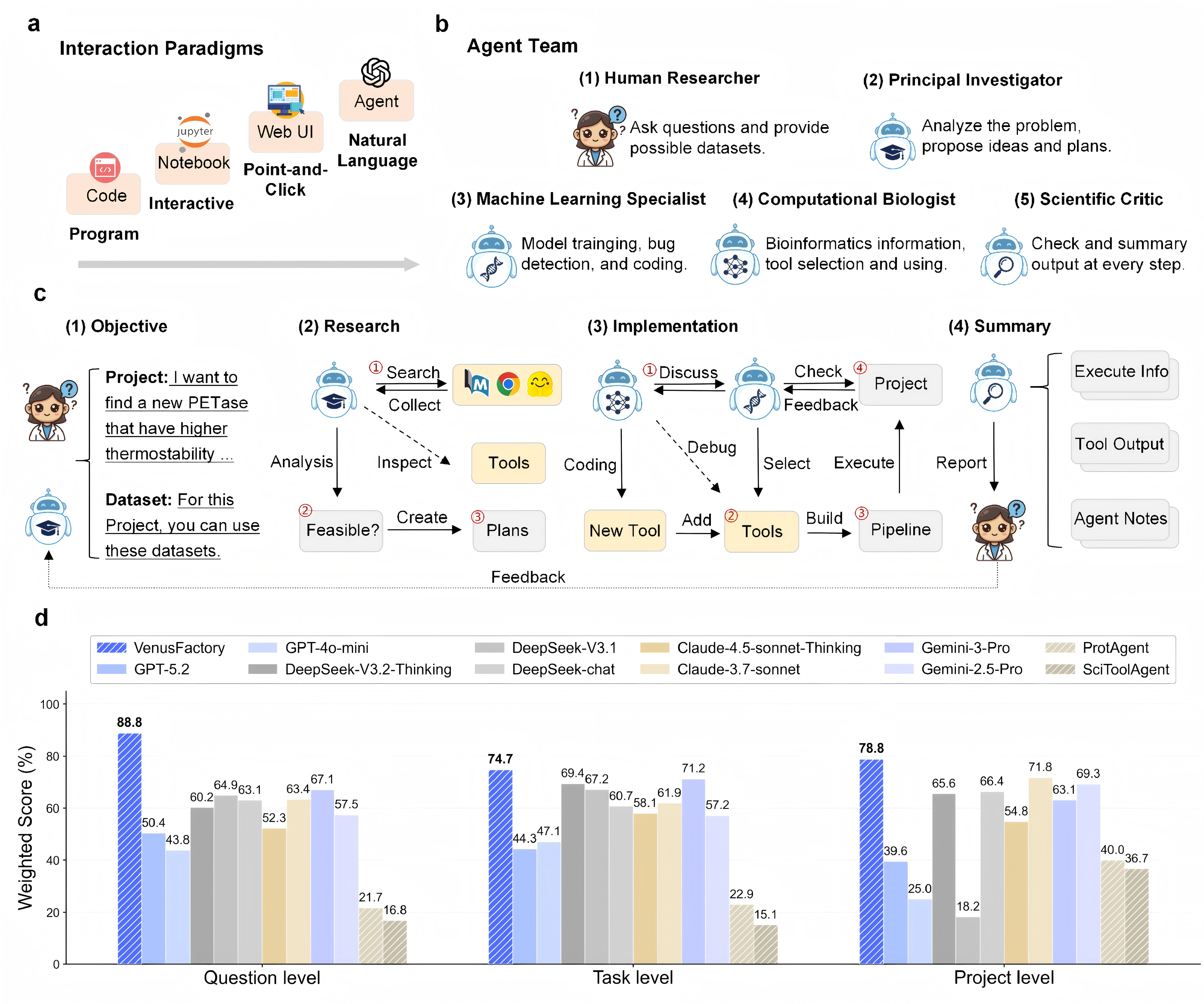}
    \caption{\textbf{The \vf~architecture and evaluation.} \textbf{a}, Evolution of interaction paradigms of AI tools in protein science, shifting from static programmatic interfaces to dynamic, natural language-driven autonomous orchestration. \textbf{b}, The multi-agent architecture comprising five specialized functional roles for collaborative reasoning and execution. \textbf{c}, The four-phase operational workflow illustrates the iterative process of task decomposition and tool invocation. \textbf{d}, Comparative performance evaluation of \vf~against a suite of state-of-the-art general-purpose LLMs and domain-specific agents across three complexity tiers of the VenusAgentEval benchmark.}
    \label{fig:framework}
\end{figure*}

Scientific discovery is propelled by the synergistic co-evolution of tools and intelligence \cite{asai2026openscholar}, while protein serves as the primary substrate for this technological evolution within life sciences \cite{gao2024bio_discovery_agent}. In this context, the integration of deep learning into protein science has yielded substantial improvements in specific tasks, such as structure prediction \cite{jumper2021alphafold2,abramson2024alphafold3}, functional annotation \cite{yu2023clean,wang2025dpfunc}, and directed evolution \cite{yang2025alde,tan2025venusrem}. While these models have achieved unprecedented accuracy, they are typically deployed as isolated, static command-line interfaces with incompatible data formats \cite{su2025saprothub}. This disconnection between conceptual biological intent and low-level programmatic execution forces researchers to dedicate disproportionate cognitive resources to low-level pipeline orchestration and software dependency resolution, thereby decoupling algorithmic intelligence from high-level biological intuition.

Recent efforts such as Jupyter notebooks \cite{mirdita2022colabfold,su2025saprothub} or graphical user interfaces \cite{abramson2024alphafold3,van2024foldseek} have simplified the execution of individual tools; however, they primarily simplify the execution of individual tools rather than resolving the fundamental challenge of knowledge-driven orchestration (Fig.~\ref{fig:framework}a). Protein engineering campaigns are rarely isolated algorithmic tasks; rather, they necessitate the seamless synthesis of domain knowledge—mined from vast literature—with heterogeneous computational predictions (e.g., using AlphaFold prediction for ESM designed mutations) \cite{swanson2025virtuallab}. In existing frameworks, bridging the gap between conceptual biological reasoning and technical execution demands extensive manual intervention \cite{jiang2024evolvepro,seo2025pet_evo}. Researchers are burdened not only with data wrangling but also with the cognitive load of contextually aligning algorithmic outputs with biological hypotheses. Consequently, scalability is constrained not by computational power, but by the temporal overhead and cognitive burden of orchestrating these disjointed workflows \cite{m2024chemcrow, ding2025scitoolagent}.

To address these limitations, we introduce \vf, an autonomous framework that elevates the interaction paradigm from rigid tool execution to collaborative scientific reasoning specifically tailored for protein engineering and discovery. \vf~mimics a digital research laboratory (Fig.~\ref{fig:framework}b) populated by specialized agents acting as \textit{Principal Investigator}, \textit{Machine Learning Specialist}, \textit{Computational Biologist}, and \textit{Scientific Critic} \cite{swanson2025virtuallab}. By simulating the collaboration in real human research teams, the system does not merely execute pipelines but proactively translates abstract biological objectives into rigorous experimental plans. Crucially, \vf~achieves self-evolving capability expansion: it detects methodological gaps and autonomously synthesizes—then deploys—executable computational tools to resolve unforeseen challenges, enabling open-ended discovery without manual intervention.

As illustrated in Figure~\ref{fig:framework}c, we simplify the research workflow of~\vf~into a four-phase protocol orchestrated by role-specific agents. (1) \textbf{Objective}: The process initiates with the human researcher specifying high-level biological objectives and optional dataset constraints. (2) \textbf{Research}: The \textit{Principal Investigator} agent conducts a feasibility analysis by cross-referencing external literature with internal tool capabilities, engaging in iterative dialogue to deeply research the experimental design. (3) \textbf{Implementation}: This phase is synergized by the \textit{Machine Learning Specialist} and \textit{Computational Biologist}. The former focuses on code review, execution, and new computational tools development, while the latter selects appropriate utilities to check and assemble the analysis pipeline. (4) \textbf{Summary}: Finally, the \textit{Scientific Critic} agent aggregates outputs, execution logs, and reasoning traces to audit biological validity and generate the final report. 

To systematically assess procedural planning and reasoning capabilities within the domains of protein discovery and directed evolution, we established VenusAgentEval (Fig.~\ref{fig:framework}d), an expert-reviewed benchmark comprising $148$ validated instances stratified into Question-, Task-, and Project-level complexities. Formally, each evaluation task is defined as a tuple $T=(Q, P, \mathcal{C})$, where the query $Q$ is paired with a multi-modal prompt context $P$—encompassing amino acid sequences, local file paths, and database identifiers—and a set of ground-truth constraints $\mathcal{C}$ for objective scoring.

Comparative evaluation on the stratified VenusAgentEval benchmark reveals distinct failure modes in existing paradigms (Fig.~\ref{fig:framework}d). \vf~demonstrates superior robustness across all complexity tiers, achieving a weighted Project-level score of 78.8\%. In sharp contrast, generalist models such as DeepSeek-V3.1, while competitive in atomic reasoning tasks (Task-level: 67.2\%), suffer a catastrophic performance collapse in long-horizon workflows (plummeting to 18.2\% at Project-level). This trajectory indicates that the implicit Chain-of-Thought capabilities of general-purpose LLMs degrade rapidly under the accumulation of biological context. Conversely, domain-specific baselines like SciToolAgent \cite{ding2025scitoolagent} and ProtAgent \cite{ghafarollahi2024protagents} exhibit architectural rigidity, failing significantly in open-ended retrieval ($<25\%$ Question-level) due to hard-coded pipelines. \vf~overcomes both the contextual fragility of generalists and the procedural rigidity of specialists, employing its explicit role-playing architecture to maintain logical consistency across multi-step discovery campaigns.

\begin{figure*}[t]
    \centering
    \includegraphics[width=\textwidth]{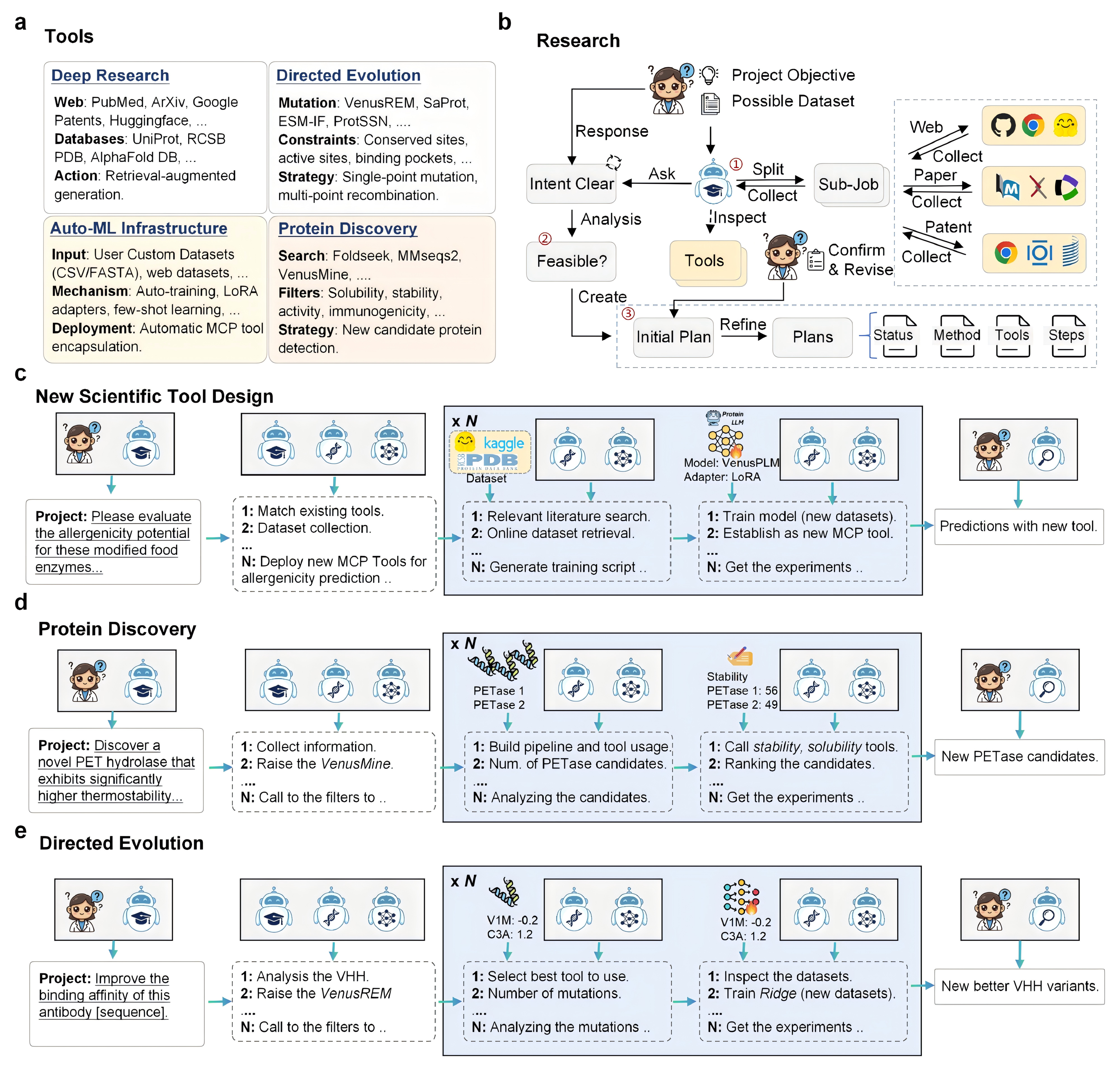}
    \caption{\textbf{Implementation mechanisms and biological validation.} \textbf{a}, Overview of the integrated computational infrastructure, stratified into four functional quadrants ranging from structural mining to generative design. \textbf{b}, Workflow of the Research module for generating structured execution protocols via hierarchical retrieval and constraint verification. \textbf{c}, Mechanism of autonomous tool instantiation, exemplified by the runtime synthesis of an allergenicity predictor. \textbf{d}, Validation in targeted enzyme retrieval, recapitulating the identification of KbPETase via semantic search and multiparametric filtration ($T_m$, pH). \textbf{e}, Validation in VHH antibody directed evolution, demonstrating adaptive code generation to model combinatorial epistatic effects utilizing single-site priors.}
    \label{fig:detail}
\end{figure*}

To address the incremental demands of protein engineering, the execution capability of \vf~integrates over 100 high-level computational utilities (Fig.~\ref{fig:detail}a), stratified into four functional quadrants (Methods). The workflow is anchored by the Research module, which implements retrieval-augmented generation across major academic and biological databases \cite{sayers2025ncbi,uniprot2025uniprot} to ground experimental hypotheses. For candidate identification, the Protein Discovery quadrant orchestrates structural or embedding alignment algorithms with biophysical filters \cite{van2024foldseek,wu2025venusmine}, while the Directed Evolution quadrant employs generative mutation models to optimize functional properties under epistatic constraints \cite{su2023saprot,tan2025venusrem}. Complementing these pre-defined capabilities, the Auto-ML Infrastructure serves as an adaptive engine \cite{tan2025venusfactory}, enabling the runtime encapsulation of user-specific datasets into deployable tools when standard methods are insufficient.

The operational logic of \vf~initiates with the Research module (Fig.~\ref{fig:detail}b), where the \textit{Principal Investigator} agent translates high-level biological objectives into structured execution protocols. Upon receiving project constraints, the agent decomposes the objective into specific retrieval sub-tasks, executing a hierarchical search strategy across web content, academic literature, and patent databases to ground the scientific hypothesis. To guarantee experimental feasibility, the module implements a constraint verification loop: the agent engages in structured dialogue to resolve ambiguities and performs a resource feasibility analysis by cross-referencing retrieved methodologies against the system's internal tool library. This process culminates in a formalized research plan—comprising current status, methodological workflow, and tool invocation sequences—that serves as a deterministic blueprint to minimize hallucinatory deviations during downstream execution.

The framework overcomes the limitations of static libraries by enabling open-ended adaptability through autonomous tool synthesis (Fig.~\ref{fig:detail}c). When the Research module identifies a missing function, exemplified by a requirement for allergenicity prediction, it aggregates training data from external repositories, user uploads, or conversation history. The agents then synthesize reproducible scripts to fine-tune foundation models using parameter-efficient methods applied to architectural templates (Methods). This workflow culminates in the automatic encapsulation of the new model as a persistent and shareable Model Context Protocol utility. Consequently, the system dynamically expands its operational scope to address specific and evolving research requirements.

To validate the utility for protein discovery, we deployed \vf~on a targeted polyethylene terephthalate hydrolase retrieval campaign (Fig.~\ref{fig:detail}d). Leveraging the VenusMine \cite{wu2025venusmine}, the system maps natural language functional descriptors including thermostable PETase directly to protein sequence space via high-dimensional embedding retrieval. In cases where template sequences are absent, the Research module initiates a literature-based search to establish a seed profile. The subsequent workflow executes a hierarchical filtration protocol wherein candidates are mined from metagenomic databases and subjected to multiparametric profiling by the \textit{Computational Biologist} agent. This agent dynamically invokes predictive assays to screen for physicochemical constraints including melting temperature ($T_m$) and pH stability. In a blind validation benchmark, \vf~successfully recapitulated the identification of KbPETase from a vast space of unannotated candidates. This confirms the system's ability to execute precision retrieval by synthesizing semantic intent with biophysical property filtering.

For complex optimization tasks, we evaluated \vf~in a directed evolution scenario targeting a VHH antibody (Fig.~\ref{fig:detail}e). The objective involved enhancing binding affinity through combinatorial mutagenesis. The system orchestrated a multi-stage pipeline, integrating AlphaFold2 \cite{jumper2021alphafold2} for structural resolution and VenusREM \cite{tan2025venusrem} for zero-shot mutational scanning. Crucially, the system demonstrated adaptive reasoning to bridge a methodological gap: while the initial tool output was limited to single-point fitness scores, the project required multi-site predictions. Identifying this discrepancy, the \textit{Machine Learning Specialist} agent autonomously synthesized and executed a Ridge Regression \cite{hsu2022few_shot_ridge_regression} algorithm, utilizing single-point priors to model combinatorial epistatic effects. The resulting top-ranked variants, including double and quadruple mutants, exhibited high correlation with ground-truth wet-lab data. This case validates the system's ability to expand beyond static tool definitions by dynamically constructing computational logic for specific biological objectives.

Collectively, \vf~advances the transition from static tool execution to intent-centric protein engineering. By resolving the cognitive complexity of workflow orchestration and enabling runtime capability expansion, the framework empowers experimentalists to deploy sophisticated computational strategies without requiring specialized programming expertise. This work establishes a foundation for self-evolving scientific agents capable of driving autonomous and open-ended discovery in biomedicine.

\bibliography{1reference}


\begin{thebibliography}{23}
\ifx \bisbn   \undefined \def \bisbn  #1{ISBN #1}\fi
\ifx \binits  \undefined \def \binits#1{#1}\fi
\ifx \bauthor  \undefined \def \bauthor#1{#1}\fi
\ifx \batitle  \undefined \def \batitle#1{#1}\fi
\ifx \bjtitle  \undefined \def \bjtitle#1{#1}\fi
\ifx \bvolume  \undefined \def \bvolume#1{\textbf{#1}}\fi
\ifx \byear  \undefined \def \byear#1{#1}\fi
\ifx \bissue  \undefined \def \bissue#1{#1}\fi
\ifx \bfpage  \undefined \def \bfpage#1{#1}\fi
\ifx \blpage  \undefined \def \blpage #1{#1}\fi
\ifx \burl  \undefined \def \burl#1{\textsf{#1}}\fi
\ifx \doiurl  \undefined \def \doiurl#1{\url{https://doi.org/#1}}\fi
\ifx \betal  \undefined \def \betal{\textit{et al.}}\fi
\ifx \binstitute  \undefined \def \binstitute#1{#1}\fi
\ifx \binstitutionaled  \undefined \def \binstitutionaled#1{#1}\fi
\ifx \bctitle  \undefined \def \bctitle#1{#1}\fi
\ifx \beditor  \undefined \def \beditor#1{#1}\fi
\ifx \bpublisher  \undefined \def \bpublisher#1{#1}\fi
\ifx \bbtitle  \undefined \def \bbtitle#1{#1}\fi
\ifx \bedition  \undefined \def \bedition#1{#1}\fi
\ifx \bseriesno  \undefined \def \bseriesno#1{#1}\fi
\ifx \blocation  \undefined \def \blocation#1{#1}\fi
\ifx \bsertitle  \undefined \def \bsertitle#1{#1}\fi
\ifx \bsnm \undefined \def \bsnm#1{#1}\fi
\ifx \bsuffix \undefined \def \bsuffix#1{#1}\fi
\ifx \bparticle \undefined \def \bparticle#1{#1}\fi
\ifx \barticle \undefined \def \barticle#1{#1}\fi
\bibcommenthead
\ifx \bconfdate \undefined \def \bconfdate #1{#1}\fi
\ifx \botherref \undefined \def \botherref #1{#1}\fi
\ifx \url \undefined \def \url#1{\textsf{#1}}\fi
\ifx \bchapter \undefined \def \bchapter#1{#1}\fi
\ifx \bbook \undefined \def \bbook#1{#1}\fi
\ifx \bcomment \undefined \def \bcomment#1{#1}\fi
\ifx \oauthor \undefined \def \oauthor#1{#1}\fi
\ifx \citeauthoryear \undefined \def \citeauthoryear#1{#1}\fi
\ifx \endbibitem  \undefined \def \endbibitem {}\fi
\ifx \bconflocation  \undefined \def \bconflocation#1{#1}\fi
\ifx \arxivurl  \undefined \def \arxivurl#1{\textsf{#1}}\fi
\csname PreBibitemsHook\endcsname

\bibitem[\protect\citeauthoryear{Asai et~al.}{2026}]{asai2026openscholar}
\begin{botherref}
\oauthor{\bsnm{Asai}, \binits{A.}},
\oauthor{\bsnm{He}, \binits{J.}},
\oauthor{\bsnm{Shao}, \binits{R.}},
\oauthor{\bsnm{Shi}, \binits{W.}},
\oauthor{\bsnm{Singh}, \binits{A.}},
\oauthor{\bsnm{Chang}, \binits{J.C.}},
\oauthor{\bsnm{Lo}, \binits{K.}},
\oauthor{\bsnm{Soldaini}, \binits{L.}},
\oauthor{\bsnm{Feldman}, \binits{S.}},
\oauthor{\bsnm{D’Arcy}, \binits{M.}}, et al.:
Synthesizing scientific literature with retrieval-augmented language models.
Nature,
1--7
(2026)
\end{botherref}
\endbibitem

\bibitem[\protect\citeauthoryear{Gao et~al.}{2024}]{gao2024bio_discovery_agent}
\begin{barticle}
\bauthor{\bsnm{Gao}, \binits{S.}},
\bauthor{\bsnm{Fang}, \binits{A.}},
\bauthor{\bsnm{Huang}, \binits{Y.}},
\bauthor{\bsnm{Giunchiglia}, \binits{V.}},
\bauthor{\bsnm{Noori}, \binits{A.}},
\bauthor{\bsnm{Schwarz}, \binits{J.R.}},
\bauthor{\bsnm{Ektefaie}, \binits{Y.}},
\bauthor{\bsnm{Kondic}, \binits{J.}},
\bauthor{\bsnm{Zitnik}, \binits{M.}}:
\batitle{Empowering biomedical discovery with ai agents}.
\bjtitle{Cell}
\bvolume{187}(\bissue{22}),
\bfpage{6125}--\blpage{6151}
(\byear{2024})
\end{barticle}
\endbibitem

\bibitem[\protect\citeauthoryear{Jumper et~al.}{2021}]{jumper2021alphafold2}
\begin{barticle}
\bauthor{\bsnm{Jumper}, \binits{J.}},
\bauthor{\bsnm{Evans}, \binits{R.}},
\bauthor{\bsnm{Pritzel}, \binits{A.}},
\bauthor{\bsnm{Green}, \binits{T.}},
\bauthor{\bsnm{Figurnov}, \binits{M.}},
\bauthor{\bsnm{Ronneberger}, \binits{O.}},
\bauthor{\bsnm{Tunyasuvunakool}, \binits{K.}},
\bauthor{\bsnm{Bates}, \binits{R.}},
\bauthor{\bsnm{{\v{Z}}{\'\i}dek}, \binits{A.}},
\bauthor{\bsnm{Potapenko}, \binits{A.}}, \betal:
\batitle{Highly accurate protein structure prediction with {AlphaFold}}.
\bjtitle{Nature}
\bvolume{596}(\bissue{7873}),
\bfpage{583}--\blpage{589}
(\byear{2021})
\end{barticle}
\endbibitem

\bibitem[\protect\citeauthoryear{Abramson et~al.}{2024}]{abramson2024alphafold3}
\begin{botherref}
\oauthor{\bsnm{Abramson}, \binits{J.}},
\oauthor{\bsnm{Adler}, \binits{J.}},
\oauthor{\bsnm{Dunger}, \binits{J.}},
\oauthor{\bsnm{Evans}, \binits{R.}},
\oauthor{\bsnm{Green}, \binits{T.}},
\oauthor{\bsnm{Pritzel}, \binits{A.}},
\oauthor{\bsnm{Ronneberger}, \binits{O.}},
\oauthor{\bsnm{Willmore}, \binits{L.}},
\oauthor{\bsnm{Ballard}, \binits{A.J.}},
\oauthor{\bsnm{Bambrick}, \binits{J.}}, et al.:
Accurate structure prediction of biomolecular interactions with {AlphaFold} 3.
Nature,
1--3
(2024)
\end{botherref}
\endbibitem

\bibitem[\protect\citeauthoryear{Yu et~al.}{2023}]{yu2023clean}
\begin{barticle}
\bauthor{\bsnm{Yu}, \binits{T.}},
\bauthor{\bsnm{Cui}, \binits{H.}},
\bauthor{\bsnm{Li}, \binits{J.C.}},
\bauthor{\bsnm{Luo}, \binits{Y.}},
\bauthor{\bsnm{Jiang}, \binits{G.}},
\bauthor{\bsnm{Zhao}, \binits{H.}}:
\batitle{Enzyme function prediction using contrastive learning}.
\bjtitle{Science}
\bvolume{379}(\bissue{6639}),
\bfpage{1358}--\blpage{1363}
(\byear{2023})
\end{barticle}
\endbibitem

\bibitem[\protect\citeauthoryear{Wang et~al.}{2025}]{wang2025dpfunc}
\begin{barticle}
\bauthor{\bsnm{Wang}, \binits{W.}},
\bauthor{\bsnm{Shuai}, \binits{Y.}},
\bauthor{\bsnm{Zeng}, \binits{M.}},
\bauthor{\bsnm{Fan}, \binits{W.}},
\bauthor{\bsnm{Li}, \binits{M.}}:
\batitle{{DPFunc}: accurately predicting protein function via deep learning with domain-guided structure information}.
\bjtitle{Nature Communications}
\bvolume{16}(\bissue{1}),
\bfpage{70}
(\byear{2025})
\end{barticle}
\endbibitem

\bibitem[\protect\citeauthoryear{Yang et~al.}{2025}]{yang2025alde}
\begin{barticle}
\bauthor{\bsnm{Yang}, \binits{J.}},
\bauthor{\bsnm{Lal}, \binits{R.G.}},
\bauthor{\bsnm{Bowden}, \binits{J.C.}},
\bauthor{\bsnm{Astudillo}, \binits{R.}},
\bauthor{\bsnm{Hameedi}, \binits{M.A.}},
\bauthor{\bsnm{Kaur}, \binits{S.}},
\bauthor{\bsnm{Hill}, \binits{M.}},
\bauthor{\bsnm{Yue}, \binits{Y.}},
\bauthor{\bsnm{Arnold}, \binits{F.H.}}:
\batitle{Active learning-assisted directed evolution}.
\bjtitle{Nature Communications}
\bvolume{16}(\bissue{1}),
\bfpage{714}
(\byear{2025})
\end{barticle}
\endbibitem

\bibitem[\protect\citeauthoryear{Tan et~al.}{2025}]{tan2025venusrem}
\begin{botherref}
\oauthor{\bsnm{Tan}, \binits{Y.}},
\oauthor{\bsnm{Wang}, \binits{R.}},
\oauthor{\bsnm{Wu}, \binits{B.}},
\oauthor{\bsnm{Hong}, \binits{L.}},
\oauthor{\bsnm{Zhou}, \binits{B.}}:
From high-throughput evaluation to wet-lab studies: advancing mutation effect prediction with a retrieval-enhanced model.
Bioinformatics
\textbf{41}
(2025)
\end{botherref}
\endbibitem

\bibitem[\protect\citeauthoryear{Su et~al.}{2025}]{su2025saprothub}
\begin{botherref}
\oauthor{\bsnm{Su}, \binits{J.}},
\oauthor{\bsnm{Li}, \binits{Z.}},
\oauthor{\bsnm{Tao}, \binits{T.}},
\oauthor{\bsnm{Han}, \binits{C.}},
\oauthor{\bsnm{He}, \binits{Y.}},
\oauthor{\bsnm{Dai}, \binits{F.}},
\oauthor{\bsnm{Yuan}, \binits{Q.}},
\oauthor{\bsnm{Gao}, \binits{Y.}},
\oauthor{\bsnm{Si}, \binits{T.}},
\oauthor{\bsnm{Zhang}, \binits{X.}}, et al.:
Democratizing protein language model training, sharing and collaboration.
Nature Biotechnology,
1--7
(2025)
\end{botherref}
\endbibitem

\bibitem[\protect\citeauthoryear{Mirdita et~al.}{2022}]{mirdita2022colabfold}
\begin{barticle}
\bauthor{\bsnm{Mirdita}, \binits{M.}},
\bauthor{\bsnm{Sch{\"u}tze}, \binits{K.}},
\bauthor{\bsnm{Moriwaki}, \binits{Y.}},
\bauthor{\bsnm{Heo}, \binits{L.}},
\bauthor{\bsnm{Ovchinnikov}, \binits{S.}},
\bauthor{\bsnm{Steinegger}, \binits{M.}}:
\batitle{Colabfold: making protein folding accessible to all}.
\bjtitle{Nature Methods}
\bvolume{19}(\bissue{6}),
\bfpage{679}--\blpage{682}
(\byear{2022})
\end{barticle}
\endbibitem

\bibitem[\protect\citeauthoryear{Van~Kempen et~al.}{2024}]{van2024foldseek}
\begin{barticle}
\bauthor{\bsnm{Van~Kempen}, \binits{M.}},
\bauthor{\bsnm{Kim}, \binits{S.S.}},
\bauthor{\bsnm{Tumescheit}, \binits{C.}},
\bauthor{\bsnm{Mirdita}, \binits{M.}},
\bauthor{\bsnm{Lee}, \binits{J.}},
\bauthor{\bsnm{Gilchrist}, \binits{C.L.}},
\bauthor{\bsnm{S{\"o}ding}, \binits{J.}},
\bauthor{\bsnm{Steinegger}, \binits{M.}}:
\batitle{Fast and accurate protein structure search with {Foldseek}}.
\bjtitle{Nature Biotechnology}
\bvolume{42}(\bissue{2}),
\bfpage{243}--\blpage{246}
(\byear{2024})
\end{barticle}
\endbibitem

\bibitem[\protect\citeauthoryear{Swanson et~al.}{2025}]{swanson2025virtuallab}
\begin{barticle}
\bauthor{\bsnm{Swanson}, \binits{K.}},
\bauthor{\bsnm{Wu}, \binits{W.}},
\bauthor{\bsnm{Bulaong}, \binits{N.L.}},
\bauthor{\bsnm{Pak}, \binits{J.E.}},
\bauthor{\bsnm{Zou}, \binits{J.}}:
\batitle{The virtual lab of ai agents designs new sars-cov-2 nanobodies}.
\bjtitle{Nature}
\bvolume{646}(\bissue{8085}),
\bfpage{716}--\blpage{723}
(\byear{2025})
\end{barticle}
\endbibitem

\bibitem[\protect\citeauthoryear{Jiang et~al.}{2024}]{jiang2024evolvepro}
\begin{barticle}
\bauthor{\bsnm{Jiang}, \binits{K.}},
\bauthor{\bsnm{Yan}, \binits{Z.}},
\bauthor{\bsnm{Di~Bernardo}, \binits{M.}},
\bauthor{\bsnm{Sgrizzi}, \binits{S.R.}},
\bauthor{\bsnm{Villiger}, \binits{L.}},
\bauthor{\bsnm{Kayabolen}, \binits{A.}},
\bauthor{\bsnm{Kim}, \binits{B.}},
\bauthor{\bsnm{Carscadden}, \binits{J.K.}},
\bauthor{\bsnm{Hiraizumi}, \binits{M.}},
\bauthor{\bsnm{Nishimasu}, \binits{H.}}, \betal:
\batitle{Rapid in silico directed evolution by a protein language model with {EVOLVEpro}}.
\bjtitle{Science}
\bvolume{387}(\bissue{6732}),
\bfpage{6006}
(\byear{2024})
\end{barticle}
\endbibitem

\bibitem[\protect\citeauthoryear{Seo et~al.}{2025}]{seo2025pet_evo}
\begin{barticle}
\bauthor{\bsnm{Seo}, \binits{H.}},
\bauthor{\bsnm{Hong}, \binits{H.}},
\bauthor{\bsnm{Park}, \binits{J.}},
\bauthor{\bsnm{Lee}, \binits{S.H.}},
\bauthor{\bsnm{Ki}, \binits{D.}},
\bauthor{\bsnm{Ryu}, \binits{A.}},
\bauthor{\bsnm{Sagong}, \binits{H.-Y.}},
\bauthor{\bsnm{Kim}, \binits{K.-J.}}:
\batitle{Landscape profiling of pet depolymerases using a natural sequence cluster framework}.
\bjtitle{Science}
\bvolume{387}(\bissue{6729}),
\bfpage{5637}
(\byear{2025})
\end{barticle}
\endbibitem

\bibitem[\protect\citeauthoryear{M.~Bran et~al.}{2024}]{m2024chemcrow}
\begin{barticle}
\bauthor{\bsnm{M.~Bran}, \binits{A.}},
\bauthor{\bsnm{Cox}, \binits{S.}},
\bauthor{\bsnm{Schilter}, \binits{O.}},
\bauthor{\bsnm{Baldassari}, \binits{C.}},
\bauthor{\bsnm{White}, \binits{A.D.}},
\bauthor{\bsnm{Schwaller}, \binits{P.}}:
\batitle{Augmenting large language models with chemistry tools}.
\bjtitle{Nature Machine Intelligence}
\bvolume{6}(\bissue{5}),
\bfpage{525}--\blpage{535}
(\byear{2024})
\end{barticle}
\endbibitem

\bibitem[\protect\citeauthoryear{Ding et~al.}{2025}]{ding2025scitoolagent}
\begin{botherref}
\oauthor{\bsnm{Ding}, \binits{K.}},
\oauthor{\bsnm{Yu}, \binits{J.}},
\oauthor{\bsnm{Huang}, \binits{J.}},
\oauthor{\bsnm{Yang}, \binits{Y.}},
\oauthor{\bsnm{Zhang}, \binits{Q.}},
\oauthor{\bsnm{Chen}, \binits{H.}}:
{SciToolAgent}: a knowledge-graph-driven scientific agent for multitool integration.
Nature Computational Science,
1--11
(2025)
\end{botherref}
\endbibitem

\bibitem[\protect\citeauthoryear{Ghafarollahi and Buehler}{2024}]{ghafarollahi2024protagents}
\begin{barticle}
\bauthor{\bsnm{Ghafarollahi}, \binits{A.}},
\bauthor{\bsnm{Buehler}, \binits{M.J.}}:
\batitle{{ProtAgents}: protein discovery via large language model multi-agent collaborations combining physics and machine learning}.
\bjtitle{Digital Discovery}
\bvolume{3}(\bissue{7}),
\bfpage{1389}--\blpage{1409}
(\byear{2024})
\end{barticle}
\endbibitem

\bibitem[\protect\citeauthoryear{Sayers et~al.}{2025}]{sayers2025ncbi}
\begin{barticle}
\bauthor{\bsnm{Sayers}, \binits{E.W.}},
\bauthor{\bsnm{Beck}, \binits{J.}},
\bauthor{\bsnm{Bolton}, \binits{E.E.}},
\bauthor{\bsnm{Brister}, \binits{J.R.}},
\bauthor{\bsnm{Chan}, \binits{J.}},
\bauthor{\bsnm{Connor}, \binits{R.}},
\bauthor{\bsnm{Feldgarden}, \binits{M.}},
\bauthor{\bsnm{Fine}, \binits{A.M.}},
\bauthor{\bsnm{Funk}, \binits{K.}},
\bauthor{\bsnm{Hoffman}, \binits{J.}}, \betal:
\batitle{Database resources of the national center for biotechnology information in 2025}.
\bjtitle{Nucleic Acids Research}
\bvolume{53}(\bissue{D1}),
\bfpage{20}--\blpage{29}
(\byear{2025})
\end{barticle}
\endbibitem

\bibitem[\protect\citeauthoryear{Consortium}{2025}]{uniprot2025uniprot}
\begin{barticle}
\bauthor{\bsnm{Consortium}, \binits{U.}}:
\batitle{{UniProt}: the universal protein knowledgebase in 2025}.
\bjtitle{Nucleic Acids Research}
\bvolume{53}(\bissue{D1}),
\bfpage{609}--\blpage{617}
(\byear{2025})
\end{barticle}
\endbibitem

\bibitem[\protect\citeauthoryear{Wu et~al.}{2025}]{wu2025venusmine}
\begin{barticle}
\bauthor{\bsnm{Wu}, \binits{B.}},
\bauthor{\bsnm{Zhong}, \binits{B.}},
\bauthor{\bsnm{Zheng}, \binits{L.}},
\bauthor{\bsnm{Huang}, \binits{R.}},
\bauthor{\bsnm{Jiang}, \binits{S.}},
\bauthor{\bsnm{Li}, \binits{M.}},
\bauthor{\bsnm{Hong}, \binits{L.}},
\bauthor{\bsnm{Tan}, \binits{P.}}:
\batitle{Harnessing protein language model for structure-based discovery of highly efficient and robust pet hydrolases}.
\bjtitle{Nature Communications}
\bvolume{16}(\bissue{1}),
\bfpage{6211}
(\byear{2025})
\end{barticle}
\endbibitem

\bibitem[\protect\citeauthoryear{Su et~al.}{2023}]{su2023saprot}
\begin{bchapter}
\bauthor{\bsnm{Su}, \binits{J.}},
\bauthor{\bsnm{Han}, \binits{C.}},
\bauthor{\bsnm{Zhou}, \binits{Y.}},
\bauthor{\bsnm{Shan}, \binits{J.}},
\bauthor{\bsnm{Zhou}, \binits{X.}},
\bauthor{\bsnm{Yuan}, \binits{F.}}:
\bctitle{{SaProt}: protein language modeling with structure-aware vocabulary}.
In: \bbtitle{The Twelfth International Conference on Learning Representations}
(\byear{2023})
\end{bchapter}
\endbibitem

\bibitem[\protect\citeauthoryear{Tan et~al.}{2025}]{tan2025venusfactory}
\begin{bchapter}
\bauthor{\bsnm{Tan}, \binits{Y.}},
\bauthor{\bsnm{Liu}, \binits{C.}},
\bauthor{\bsnm{Gao}, \binits{J.}},
\bauthor{\bsnm{Banghao}, \binits{W.}},
\bauthor{\bsnm{Li}, \binits{M.}},
\bauthor{\bsnm{Wang}, \binits{R.}},
\bauthor{\bsnm{Zhang}, \binits{L.}},
\bauthor{\bsnm{Yu}, \binits{H.}},
\bauthor{\bsnm{Fan}, \binits{G.}},
\bauthor{\bsnm{Hong}, \binits{L.}}, \betal:
\bctitle{{VenusFactory}: An integrated system for protein engineering with data retrieval and language model fine-tuning}.
In: \bbtitle{Proceedings of the 63rd Annual Meeting of the Association for Computational Linguistics (Volume 3: System Demonstrations)},
pp. \bfpage{230}--\blpage{241}
(\byear{2025})
\end{bchapter}
\endbibitem

\bibitem[\protect\citeauthoryear{Hsu et~al.}{2022}]{hsu2022few_shot_ridge_regression}
\begin{barticle}
\bauthor{\bsnm{Hsu}, \binits{C.}},
\bauthor{\bsnm{Nisonoff}, \binits{H.}},
\bauthor{\bsnm{Fannjiang}, \binits{C.}},
\bauthor{\bsnm{Listgarten}, \binits{J.}}:
\batitle{Learning protein fitness models from evolutionary and assay-labeled data}.
\bjtitle{Nature Biotechnology}
\bvolume{40}(\bissue{7}),
\bfpage{1114}--\blpage{1122}
(\byear{2022})
\end{barticle}
\endbibitem

\end{thebibliography}


\begin{thebibliography}{10}

\bibitem{paysan2023interpro}
Typhaine Paysan-Lafosse, Matthias Blum, Sara Chuguransky, Tiago Grego, Beatriz~L{\'a}zaro Pinto, Gustavo~A Salazar, Maxwell~L Bileschi, Peer Bork, Alan Bridge, Lucy Colwell, et~al.
\newblock {InterPro} in 2022.
\newblock {\em Nucleic Acids Research}, 51(D1):D418--D427, 2023.

\bibitem{burley2019rcsb}
Stephen~K Burley, Helen~M Berman, Charmi Bhikadiya, Chunxiao Bi, Li~Chen, Luigi Di~Costanzo, Cole Christie, Ken Dalenberg, Jose~M Duarte, Shuchismita Dutta, et~al.
\newblock {RCSB Protein Data Bank}: biological macromolecular structures enabling research and education in fundamental biology, biomedicine, biotechnology and energy.
\newblock {\em Nucleic Acids Research}, 47(D1):D464--D474, 2019.

\bibitem{varadi2022alphafolddb}
Mihaly Varadi, Stephen Anyango, Mandar Deshpande, Sreenath Nair, Cindy Natassia, Galabina Yordanova, David Yuan, Oana Stroe, Gemma Wood, Agata Laydon, et~al.
\newblock Alphafold protein structure database: massively expanding the structural coverage of protein-sequence space with high-accuracy models.
\newblock {\em Nucleic Acids Research}, 50(D1):D439--D444, 2022.

\bibitem{hauenstein2026brenda}
Julia Hauenstein, Lisa Jeske, Antje J{\"a}de, Mathias Krull, Katrin D{\"u}mmer, Julia Koblitz, Anja Tietz, Dieter Jahn, Lorenz~Christian Reimer, and Boyke Bunk.
\newblock {BRENDA} in 2026: a {Global Core Biodata Resource} for functional enzyme and metabolic data within the {DSMZ Digital Diversity}.
\newblock {\em Nucleic Acids Research}, 54(D1):D527--D534, 2026.

\bibitem{zdrazil2024chembl}
Barbara Zdrazil, Eloy Felix, Fiona Hunter, Emma~J Manners, James Blackshaw, Sybilla Corbett, Marleen De~Veij, Harris Ioannidis, David~Mendez Lopez, Juan~F Mosquera, et~al.
\newblock The {ChEMBL Database} in 2023: a drug discovery platform spanning multiple bioactivity data types and time periods.
\newblock {\em Nucleic Acids Research}, 52(D1):D1180--D1192, 2024.

\bibitem{kanehisa2025kegg}
Minoru Kanehisa, Miho Furumichi, Yoko Sato, Yuriko Matsuura, and Mari Ishiguro-Watanabe.
\newblock {KEGG}: biological systems database as a model of the real world.
\newblock {\em Nucleic Acids Research}, 53(D1):D672--D677, 2025.

\bibitem{szklarczyk2025string}
Damian Szklarczyk, Katerina Nastou, Mikaela Koutrouli, Rebecca Kirsch, Farrokh Mehryary, Radja Hachilif, Dewei Hu, Matteo~E Peluso, Qingyao Huang, Tao Fang, et~al.
\newblock The {STRING} database in 2025: protein networks with directionality of regulation.
\newblock {\em Nucleic Acids Research}, 53(D1):D730--D737, 2025.

\bibitem{Zheng2026protssn_wet_lab}
Lirong Zheng, Dongqing Shi, Yixiao Yan, Bingxin Zhou, Jormay Lim, Yongjie Hou, Bobae An, Jason~K. Adhinarta, Michael Lin, BumJin Ko, William~C. Joesten, Mehul Gautam, Elie D.~M. Huez, Eung~Chang Kim, Emily~G. Klyder, Boxuan Chang, Sethuramasundaram Pitchiaya, Michael~T. Roberts, Denise~J. Cai, Edward~S. Boyden, Donglai Wei, Pietro Li{\`o}, and Changyang Linghu.
\newblock Scalable and multiplexed recorders of gene regulation dynamics across weeks.
\newblock {\em Nature}, jan 2026.

\bibitem{zhou2024fsfp}
Ziyi Zhou, Liang Zhang, Yuanxi Yu, Banghao Wu, Mingchen Li, Liang Hong, and Pan Tan.
\newblock Enhancing efficiency of protein language models with minimal wet-lab data through few-shot learning.
\newblock {\em Nature Communications}, 15(1):5566, 2024.

\bibitem{notin2024proteingym}
Pascal Notin, Aaron Kollasch, Daniel Ritter, Lood Van~Niekerk, Steffanie Paul, Han Spinner, Nathan Rollins, Ada Shaw, Rose Orenbuch, Ruben Weitzman, et~al.
\newblock {ProteinGym}: large-scale benchmarks for protein fitness prediction and design.
\newblock In {\em Advances in Neural Information Processing Systems}, volume~36, 2024.

\bibitem{zhang2025venusmuthub}
Liang Zhang, Hua Pang, Chenghao Zhang, Song Li, Yang Tan, Fan Jiang, Mingchen Li, Yuanxi Yu, Ziyi Zhou, Banghao Wu, et~al.
\newblock Venusmuthub: a systematic evaluation of protein mutation effect predictors on small-scale experimental data.
\newblock {\em Acta Pharmaceutica Sinica B}, 2025.

\bibitem{lin2023esm2}
Zeming Lin, Halil Akin, Roshan Rao, Brian Hie, Zhongkai Zhu, Wenting Lu, Nikita Smetanin, Robert Verkuil, Ori Kabeli, Yaniv Shmueli, et~al.
\newblock Evolutionary-scale prediction of atomic-level protein structure with a language model.
\newblock {\em Science}, 379(6637):1123--1130, 2023.

\bibitem{meier2021esm1v}
Joshua Meier, Roshan Rao, Robert Verkuil, Jason Liu, Tom Sercu, and Alex Rives.
\newblock Language models enable zero-shot prediction of the effects of mutations on protein function.
\newblock {\em Advances in Neural Information Processing Systems}, 34:29287--29303, 2021.

\bibitem{rives2021esm1b}
Alexander Rives, Joshua Meier, Tom Sercu, Siddharth Goyal, Zeming Lin, Jason Liu, Demi Guo, Myle Ott, C~Lawrence Zitnick, Jerry Ma, et~al.
\newblock Biological structure and function emerge from scaling unsupervised learning to 250 million protein sequences.
\newblock {\em Proceedings of the National Academy of Sciences}, 118(15):e2016239118, 2021.

\bibitem{li2024prosst}
Mingchen Li, Yang Tan, Xinzhu Ma, Bozitao Zhong, Huiqun Yu, Ziyi Zhou, Wanli Ouyang, Bingxin Zhou, Pan Tan, and Liang Hong.
\newblock Pro{SST}: Protein language modeling with quantized structure and disentangled attention.
\newblock In {\em The Thirty-eighth Annual Conference on Neural Information Processing Systems}, 2024.

\bibitem{tan2025protssn}
Yang Tan, Bingxin Zhou, Lirong Zheng, Guisheng Fan, and Liang Hong.
\newblock Semantical and geometrical protein encoding toward enhanced bioactivity and thermostability.
\newblock {\em eLife}, 13:RP98033, may 2025.

\bibitem{hsu2022esm-if1}
Chloe Hsu, Robert Verkuil, Jason Liu, Zeming Lin, Brian Hie, Tom Sercu, Adam Lerer, and Alexander Rives.
\newblock Learning inverse folding from millions of predicted structures.
\newblock In {\em International Conference on Machine Learning}, pages 8946--8970. PMLR, 2022.

\bibitem{mifst}
Kevin~K Yang, Niccol{\`o} Zanichelli, and Hugh Yeh.
\newblock Masked inverse folding with sequence transfer for protein representation learning.
\newblock {\em Protein Engineering, Design and Selection}, 36:gzad015, 2023.

\bibitem{tan2026venusx}
Yang Tan, Wenrui Gou, Bozitao Zhong, Huiqun Yu, Liang Hong, and Bingxin Zhou.
\newblock {VenusX}: Unlocking fine-grained functional understanding of proteins.
\newblock In {\em The Fourteenth International Conference on Learning Representations}, 2026.

\bibitem{carradec2018gopc}
Quentin Carradec, Eric Pelletier, Corinne Da~Silva, Adriana Alberti, Yoann Seeleuthner, Romain Blanc-Mathieu, Gipsi Lima-Mendez, Fabio Rocha, Leila Tirichine, Karine Labadie, et~al.
\newblock A global ocean atlas of eukaryotic genes.
\newblock {\em Nature Communications}, 9(1):373, 2018.

\bibitem{richardson2023mgnify}
Lorna Richardson, Ben Allen, Germana Baldi, Martin Beracochea, Maxwell~L Bileschi, Tony Burdett, Josephine Burgin, Juan Caballero-P{\'e}rez, Guy Cochrane, Lucy~J Colwell, et~al.
\newblock {MGnify}: the microbiome sequence data analysis resource in 2023.
\newblock {\em Nucleic Acids Research}, 51(D1):D753--D759, 2023.

\bibitem{cornman2025omg}
Andre Cornman, Jacob West-Roberts, Antonio~Pedro Camargo, Simon Roux, Martin Beracochea, Milot Mirdita, Sergey Ovchinnikov, and Yunha Hwang.
\newblock The {OMG} dataset: An open metagenomic corpus for mixed-modality genomic language modeling.
\newblock In {\em The Thirteenth International Conference on Learning Representations}, 2025.

\bibitem{altschul1990blast}
Stephen~F Altschul, Warren Gish, Webb Miller, Eugene~W Myers, and David~J Lipman.
\newblock Basic local alignment search tool.
\newblock {\em Journal of Molecular Biology}, 215(3):403--410, 1990.

\bibitem{kallenborn2025mmseqs2-gpu}
Felix Kallenborn, Alejandro Chacon, Christian Hundt, Hassan Sirelkhatim, Kieran Didi, Sooyoung Cha, Christian Dallago, Milot Mirdita, Bertil Schmidt, and Martin Steinegger.
\newblock {GPU}-accelerated homology search with {MMseqs2}.
\newblock {\em Nature Methods}, 22(10):2024--2027, 2025.

\bibitem{Khurana2018deepsol}
Sameer Khurana, Reda Rawi, Khalid Kunji, Gwo-Yu Chuang, Halima Bensmail, and Raghvendra Mall.
\newblock {DeepSol}: a deep learning framework for sequence-based protein solubility prediction.
\newblock {\em Bioinformatics}, 34(15):2605--2613, 03 2018.

\bibitem{tan2024protsolm}
Yang Tan, Jia Zheng, Liang Hong, and Bingxin Zhou.
\newblock {ProtSolM}: Protein solubility prediction with multi-modal features.
\newblock In {\em 2024 IEEE International Conference on Bioinformatics and Biomedicine (BIBM)}, pages 223--232. IEEE, 2024.

\bibitem{chen2021esol}
Jianwen Chen, Shuangjia Zheng, Huiying Zhao, and Yuedong Yang.
\newblock Structure-aware protein solubility prediction from sequence through graph convolutional network and predicted contact map.
\newblock {\em Journal of cheminformatics}, 13:1--10, 2021.

\bibitem{almagro2017deeploc}
Jos{\'e}~Juan Almagro~Armenteros, Casper~Kaae S{\o}nderby, S{\o}ren~Kaae S{\o}nderby, Henrik Nielsen, and Ole Winther.
\newblock {DeepLoc}: prediction of protein subcellular localization using deep learning.
\newblock {\em Bioinformatics}, 33(21):3387--3395, 2017.

\bibitem{thumuluri2022deeploc2}
Vineet Thumuluri, Jos{\'e}~Juan Almagro~Armenteros, Alexander~Rosenberg Johansen, Henrik Nielsen, and Ole Winther.
\newblock {DeepLoc 2.0}: multi-label subcellular localization prediction using protein language models.
\newblock {\em Nucleic Acids Research}, 50(W1):W228--W234, 2022.

\bibitem{odum2024deeploc2.1}
Marius~Thrane {\O}dum, Felix Teufel, Vineet Thumuluri, Jos{\'e}~Juan Almagro~Armenteros, Alexander~Rosenberg Johansen, Ole Winther, and Henrik Nielsen.
\newblock {DeepLoc 2.1}: multi-label membrane protein type prediction using protein language models.
\newblock {\em Nucleic Acids Research}, 52(W1):W215--W220, 2024.

\bibitem{tan2024sesadapter}
Yang Tan, Mingchen Li, Bingxin Zhou, Bozitao Zhong, Lirong Zheng, Pan Tan, Ziyi Zhou, Huiqun Yu, Guisheng Fan, and Liang Hong.
\newblock Simple, efficient, and scalable structure-aware adapter boosts protein language models.
\newblock {\em Journal of Chemical Information and Modeling}, 2024.

\bibitem{li2022deepet}
Gang Li, Filip Buric, Jan Zrimec, Sandra Viknander, Jens Nielsen, Aleksej Zelezniak, and Martin~KM Engqvist.
\newblock Learning deep representations of enzyme thermal adaptation.
\newblock {\em Protein Science}, 31(12):e4480, 2022.

\bibitem{gado2025ephod}
Japheth~E Gado, Matthew Knotts, Ada~Y Shaw, Debora Marks, Nicholas~P Gauthier, Chris Sander, and Gregg~T Beckham.
\newblock Machine learning prediction of enzyme optimum ph.
\newblock {\em Nature Machine Intelligence}, pages 1--14, 2025.

\bibitem{li2022dlkcat}
Feiran Li, Le~Yuan, Hongzhong Lu, Gang Li, Yu~Chen, Martin~KM Engqvist, Eduard~J Kerkhoven, and Jens Nielsen.
\newblock Deep learning-based k cat prediction enables improved enzyme-constrained model reconstruction.
\newblock {\em Nature Catalysis}, 5(8):662--672, 2022.

\bibitem{li2025VenusVaccine}
Song Li, Yang Tan, Song Ke, Liang Hong, and Bingxin Zhou.
\newblock Immunogenicity prediction with dual attention enables vaccine target selection.
\newblock In {\em The Thirteenth International Conference on Learning Representations}, 2025.

\bibitem{cock2009biopython}
Peter~JA Cock, Tiago Antao, Jeffrey~T Chang, Brad~A Chapman, Cymon~J Cox, Andrew Dalke, Iddo Friedberg, Thomas Hamelryck, Frank Kauff, Bartek Wilczynski, et~al.
\newblock Biopython: freely available python tools for computational molecular biology and bioinformatics.
\newblock {\em Bioinformatics}, 25(11):1422, 2009.

\bibitem{mangrulkar2023peft}
S~Mangrulkar, S~Gugger, L~Debut, Yo~Belkada, S~Paul, and B~Bossan.
\newblock {PEFT}: state-of-the-art parameter-efficient fine-tuning methods (2022).
\newblock {\em URL https://github. com/huggingface/peft}, 2023.

\bibitem{hu2022lora}
Edward~J Hu, yelong shen, Phillip Wallis, Zeyuan Allen-Zhu, Yuanzhi Li, Shean Wang, Lu~Wang, and Weizhu Chen.
\newblock Lo{RA}: Low-rank adaptation of large language models.
\newblock In {\em International Conference on Learning Representations}, 2022.

\bibitem{liu2024dora}
Shih-Yang Liu, Chien-Yi Wang, Hongxu Yin, Pavlo Molchanov, Yu-Chiang~Frank Wang, Kwang-Ting Cheng, and Min-Hung Chen.
\newblock Dora: Weight-decomposed low-rank adaptation.
\newblock In {\em Forty-first International Conference on Machine Learning}, 2024.

\bibitem{zhang2023adalora}
Qingru Zhang, Minshuo Chen, Alexander Bukharin, Pengcheng He, Yu~Cheng, Weizhu Chen, and Tuo Zhao.
\newblock Adaptive budget allocation for parameter-efficient fine-tuning.
\newblock In {\em The Eleventh International Conference on Learning Representations}, 2024.

\bibitem{dettmers2023qlora}
Tim Dettmers, Artidoro Pagnoni, Ari Holtzman, and Luke Zettlemoyer.
\newblock Qlora: Efficient finetuning of quantized llms.
\newblock {\em Advances in neural information processing systems}, 36:10088--10115, 2023.

\bibitem{liu2022lora_ia3}
Haokun Liu, Derek Tam, Mohammed Muqeeth, Jay Mohta, Tenghao Huang, Mohit Bansal, and Colin~A Raffel.
\newblock Few-shot parameter-efficient fine-tuning is better and cheaper than in-context learning.
\newblock {\em Advances in Neural Information Processing Systems}, 35:1950--1965, 2022.

\bibitem{kabsch1983dssp}
Wolfgang Kabsch and Christian Sander.
\newblock Dictionary of protein secondary structure: pattern recognition of hydrogen-bonded and geometrical features.
\newblock {\em Biopolymers: Original Research on Biomolecules}, 22(12):2577--2637, 1983.

\bibitem{elnaggar2021prottrans}
Ahmed Elnaggar, Michael Heinzinger, Christian Dallago, Ghalia Rehawi, Yu~Wang, Llion Jones, Tom Gibbs, Tamas Feher, Christoph Angerer, Martin Steinegger, et~al.
\newblock Prottrans: Toward understanding the language of life through self-supervised learning.
\newblock {\em IEEE Transactions on Pattern Analysis and Machine Intelligence}, 44(10):7112--7127, 2021.

\bibitem{tan2023peta}
Yang Tan, Mingchen Li, Ziyi Zhou, Pan Tan, Huiqun Yu, Guisheng Fan, and Liang Hong.
\newblock {PETA}: evaluating the impact of protein transfer learning with sub-word tokenization on downstream applications.
\newblock {\em Journal of Cheminformatics}, 16(1):92, 2024.

\bibitem{elnaggar2023ankh}
Ahmed Elnaggar, Hazem Essam, Wafaa Salah-Eldin, Walid Moustafa, Mohamed Elkerdawy, Charlotte Rochereau, and Burkhard Rost.
\newblock Ankh: Optimized protein language model unlocks general-purpose modelling.
\newblock {\em arXiv:2301.06568}, 2023.

\bibitem{alsamkary2025ankh3}
Hazem Alsamkary, Mohamed Elshaffei, Mohamed Elkerdawy, and Ahmed Elnaggar.
\newblock Ankh3: Multi-task pretraining with sequence denoising and completion enhances protein representations.
\newblock {\em arXiv preprint arXiv:2505.20052}, 2025.

\bibitem{kenlay2024igbert}
Henry Kenlay, Fr{\'e}d{\'e}ric~A Dreyer, Aleksandr Kovaltsuk, Dom Miketa, Douglas Pires, and Charlotte~M Deane.
\newblock Large scale paired antibody language models.
\newblock {\em PLOS Computational Biology}, 20(12):e1012646, 2024.

\bibitem{wang2023deepsolue}
Chao Wang and Quan Zou.
\newblock Prediction of protein solubility based on sequence physicochemical patterns and distributed representation information with {DeepSoluE}.
\newblock {\em BMC Biology}, 21(1):12, 2023.

\bibitem{center2026hle_human_last_exam}
Center for AI Safety Phan Long agibenchmark@ safe. ai 1 Gatti Alice 1 Li Nathaniel 1 Khoja Adam 1 Kim Ryan 1 Ren Richard 1 Hausenloy Jason 1 Zhang Oliver 1 Mazeika Mantas 1 Hendrycks Dan dan@ safe.~ai 1.
\newblock A benchmark of expert-level academic questions to assess ai capabilities.
\newblock {\em Nature}, 649(8099):1139--1146, 2026.

\bibitem{du2026deepresearch_bench}
Mingxuan Du, Benfeng Xu, Chiwei Zhu, Licheng Zhang, Xiaorui Wang, and Zhendong Mao.
\newblock {DeepResearch Bench}: A comprehensive benchmark for deep research agents.
\newblock In {\em The Fourteenth International Conference on Learning Representations}, 2026.

\bibitem{wang2023self-instruct}
Yizhong Wang, Yeganeh Kordi, Swaroop Mishra, Alisa Liu, Noah~A Smith, Daniel Khashabi, and Hannaneh Hajishirzi.
\newblock Self-instruct: Aligning language models with self-generated instructions.
\newblock In {\em Proceedings of the 61st annual meeting of the association for computational linguistics (volume 1: long papers)}, pages 13484--13508, 2023.

\bibitem{achiam2023gpt4}
Josh Achiam, Steven Adler, Sandhini Agarwal, Lama Ahmad, Ilge Akkaya, Florencia~Leoni Aleman, Diogo Almeida, Janko Altenschmidt, Sam Altman, Shyamal Anadkat, et~al.
\newblock Gpt-4 technical report.
\newblock {\em arXiv preprint arXiv:2303.08774}, 2023.

\bibitem{anthropic2024claude}
Anthropic.
\newblock The claude 3 model family: Opus, sonnet, haiku, 2024.
\newblock Technical Report.

\bibitem{comanici2025gemini2.5}
Gheorghe Comanici, Eric Bieber, Mike Schaekermann, Ice Pasupat, Noveen Sachdeva, Inderjit Dhillon, Marcel Blistein, Ori Ram, Dan Zhang, Evan Rosen, et~al.
\newblock Gemini 2.5: Pushing the frontier with advanced reasoning, multimodality, long context, and next generation agentic capabilities.
\newblock {\em arXiv preprint arXiv:2507.06261}, 2025.

\bibitem{guo2025deepseek}
Daya Guo, Dejian Yang, Haowei Zhang, Junxiao Song, Peiyi Wang, Qihao Zhu, Runxin Xu, Ruoyu Zhang, Shirong Ma, Xiao Bi, et~al.
\newblock {DeepSeek-R1} incentivizes reasoning in llms through reinforcement learning.
\newblock {\em Nature}, 645(8081):633--638, 2025.

\bibitem{liang2024llm_peer_review}
Weixin Liang, Yuhui Zhang, Hancheng Cao, Binglu Wang, Daisy~Yi Ding, Xinyu Yang, Kailas Vodrahalli, Siyu He, Daniel~Scott Smith, Yian Yin, et~al.
\newblock Can large language models provide useful feedback on research papers? a large-scale empirical analysis.
\newblock {\em NEJM AI}, 1(8):AIoa2400196, 2024.

\bibitem{chen2021human_eval}
Mark Chen.
\newblock Evaluating large language models trained on code.
\newblock {\em arXiv preprint arXiv:2107.03374}, 2021.

\bibitem{hurst2024gpt4o}
Aaron Hurst, Adam Lerer, Adam~P Goucher, Adam Perelman, Aditya Ramesh, Aidan Clark, AJ~Ostrow, Akila Welihinda, Alan Hayes, Alec Radford, et~al.
\newblock {GPT-4o} system card.
\newblock {\em arXiv preprint arXiv:2410.21276}, 2024.

\bibitem{singh2025gpt5}
Aaditya Singh, Adam Fry, Adam Perelman, Adam Tart, Adi Ganesh, Ahmed El-Kishky, Aidan McLaughlin, Aiden Low, AJ~Ostrow, Akhila Ananthram, et~al.
\newblock {OpenAI} gpt-5 system card.
\newblock {\em arXiv preprint arXiv:2601.03267}, 2025.

\bibitem{heinzinger2023prostt5}
Michael Heinzinger, Konstantin Weissenow, Joaquin~Gomez Sanchez, Adrian Henkel, Martin Steinegger, and Burkhard Rost.
\newblock {ProstT5}: Bilingual language model for protein sequence and structure.
\newblock {\em bioRxiv}, pages 2023--07, 2023.

\bibitem{burgin2024ispet}
Tucker Burgin, Benjamin~C Pollard, Brandon~C Knott, Heather~B Mayes, Michael~F Crowley, John~E McGeehan, Gregg~T Beckham, and H~Lee Woodcock.
\newblock The reaction mechanism of the ideonella sakaiensis petase enzyme.
\newblock {\em Communications Chemistry}, 7(1):65, 2024.

\bibitem{cui2024fastpet}
Yinglu Cui, Yanchun Chen, Jinyuan Sun, Tong Zhu, Hua Pang, Chunli Li, Wen-Chao Geng, and Bian Wu.
\newblock Computational redesign of a hydrolase for nearly complete pet depolymerization at industrially relevant high-solids loading.
\newblock {\em Nature Communications}, 15(1):1417, 2024.

\bibitem{jiang2024prime}
Fan Jiang, Mingchen Li, Jiajun Dong, Yuanxi Yu, Xinyu Sun, Banghao Wu, Jin Huang, Liqi Kang, Yufeng Pei, Liang Zhang, et~al.
\newblock A general temperature-guided language model to design proteins of enhanced stability and activity.
\newblock {\em Science Advances}, 10(48):eadr2641, 2024.

\end{thebibliography}

\newpage

\section*{Methods}\label{sec:methods}

\subsection*{Research Tool Ecosystem}
\label{app:sec:research_tool_ecosystem}

\textbf{Federated Web Intelligence.} \label{app:sec:tool_deep_research}
To synthesize multi-modal scientific information, the agent employs a federated search interface that interrogates four distinct information domains.
(1) \textbf{Academic Literature:} We integrate programmatic access to \href{https://pubmed.ncbi.nlm.nih.gov/}{\texttt{PubMed}}, \href{https://www.webofscience.com/}{\texttt{Web of Science}}, and \href{https://www.semanticscholar.org/}{\texttt{Semantic Scholar}} to retrieve peer-reviewed articles, complemented by \href{https://arxiv.org/}{\texttt{arXiv}} and \href{https://www.biorxiv.org/}{\texttt{bioRxiv}} for preprints, ensuring coverage of the latest algorithmic advances.
(2) \textbf{Technical Repositories:} To bridge theoretical concepts with implementation, the system queries \href{https://github.com/}{\texttt{GitHub}} for codebases, and \href{https://huggingface.co/}{\texttt{Hugging Face}} and \href{https://www.kaggle.com/}{\texttt{Kaggle}} for benchmarks and datasets.
(3) \textbf{Intellectual Property:} Industrial relevance is assessed by mining the \href{https://console.cloud.google.com/marketplace/product/google_patents_public_datasets/google-patents-public-data}{\texttt{Google Patents Public Datasets}} via \href{https://cloud.google.com/bigquery}{\texttt{Google BigQuery}}, allowing the agent to verify novelty and freedom-to-operate.
(4) \textbf{Open-Web Context:} For unstructured content (e.g., industry news, documentation), the system leverages the \href{https://tavily.com/}{\texttt{Tavily}} and \href{https://duckduckgo.com/}{\texttt{DuckDuckGo}} Search API, optimized for LLM-centric retrieval, to capture high-precision contextual intelligence.

\textbf{Database Interrogation Engine.} \label{app:sec:tool_database_search}
To ensure comprehensive coverage of the protein universe, \vf~integrates a high-throughput data collection engine spanning nine canonical repositories across sequence, structure, enzymology, and functional networks. 
(1) \textbf{Sequence and Function:} \href{https://www.uniprot.org/}{\texttt{UniProt}} \cite{uniprot2025uniprot} serves as the central repository for amino acid sequences and curated annotations ($>250$ million entries). This is supplemented by \href{https://www.ncbi.nlm.nih.gov/}{\texttt{NCBI}} \cite{sayers2025ncbi} databases (GenBank, RefSeq) to capture non-canonical entries. Additionally, \href{https://www.ebi.ac.uk/interpro/}{\texttt{InterPro}} \citemethods{paysan2023interpro} is leveraged to assign functional signatures, classifying proteins into families and domains ($\sim41,000$ entries). 
(2) \textbf{Structural Data:} \href{https://www.rcsb.org/}{\texttt{RCSB PDB}} \citemethods{burley2019rcsb} provides over $200,000$ experimentally determined 3D structures. For targets lacking experimental data, the system retrieves predicted structures from the \href{https://alphafold.ebi.ac.uk/}{\texttt{AlphaFold Database}} \citemethods{varadi2022alphafolddb} via UniProt ID mapping. 
(3) \textbf{Biochemical and Interaction Networks:} To contextualize proteins within complex biological systems, the engine queries \href{https://www.brenda-enzymes.org/}{\texttt{BRENDA}} \citemethods{hauenstein2026brenda} for comprehensive enzyme kinetics and functional parameters. Protein-ligand binding affinities and small-molecule bioactivity are retrieved via \href{https://www.ebi.ac.uk/chembl/}{\texttt{ChEMBL}} \citemethods{zdrazil2024chembl}. Furthermore, \href{https://www.genome.jp/kegg/}{\texttt{KEGG}} \citemethods{kanehisa2025kegg} is integrated to map targets onto higher-order metabolic and signaling pathways, while the \href{https://string-db.org/}{\texttt{STRING}} database \citemethods{szklarczyk2025string} provides probabilistic profiling of protein-protein interactions.
Technically, the module implements an asynchronous, multithreaded architecture to maximize data acquisition throughput (see \textbf{Supplementary Table~\ref{app:tab:deep_search_tools}}).

\subsection*{Directed Evolution Tools}
\label{app:sec:directed _evo_ecosystem}
\textbf{Mutational Scanning.}
For directed evolution prediction, we deployed a comprehensive suite of protein language models (PLMs), prioritizing architectures validated through wet-lab experiments \citemethods{Zheng2026protssn_wet_lab,zhou2024fsfp} or superior benchmark performance \citemethods{notin2024proteingym,zhang2025venusmuthub}. To capture sequence-based evolutionary patterns, the system integrates the evolutionary scale modeling suite, including ESM2 \citemethods{lin2023esm2}, ESM1v \citemethods{meier2021esm1v}, and ESM1b \citemethods{rives2021esm1b}. For structure-conditioned predictions, we deployed hybrid architectures to leverage geometric constraints: VenusREM \cite{tan2025venusrem} integrates homologous sequences with structural information. ProSST \citemethods{li2024prosst} utilizes discrete attention mechanisms. ProtSSN \citemethods{tan2025protssn}, a sequence-structure serial network. SaProt \cite{su2023saprot} employs Foldseek-derived \cite{van2024foldseek} token encoding. The inverse folding model ESM-IF \citemethods{hsu2022esm-if1} and MIF-ST \citemethods{mifst}, combining graph neural networks. Furthermore, for ``Low-N" scenarios involving iterative wet-lab feedback, we implemented a sample-efficient Ridge Regression protocol \cite{hsu2022few_shot_ridge_regression} which is implemented by \vf~itself. This supervised module augments one-hot amino acid encoding with probability density features derived from unsupervised PLMs, demonstrating superior generalization in data-scarce regimes.

\textbf{Functional Sites Detection.}
To constrain the mutational search space and prevent the disruption of critical biological functions, we developed the VenusX \citemethods{tan2026venusx} module, an ensemble of residue-level predictors designed to identify and mask essential sites. Built upon PLMs, these models were trained on datasets partitioned at a strict $90\%$ sequence identity threshold to minimize data leakage. The suite comprises specialized predictors for catalytic active sites, ligand-binding interfaces, evolutionary conserved residues, functional motifs, and domain boundaries. Ground-truth labels were derived from curated InterPro entries \citemethods{paysan2023interpro}, supporting binary classification of functionally relevant residues. (see \textbf{Supplementary Table~\ref{app:tab:directed_evolution_tools}})

\subsection*{Protein Discovery Module}
\label{app:sec:protein_discovery_ecosystem}
\textbf{Homology Mining Engine.}
To enable comprehensive protein discovery, we curated and consolidated four high-quality protein sequence datasets derived from the Global Ocean Protein Catalog (GOPC) \citemethods{carradec2018gopc}, UniProt UniRef100 \cite{uniprot2025uniprot}, the MGnify metagenomic database \citemethods{richardson2023mgnify}, and the Open MetaGenomic (OMG) corpus \citemethods{cornman2025omg}, all deduplicated at $100\%$ sequence identity to remove redundancy. To navigate these vast sequence spaces, the system integrates a multi-modal retrieval architecture. Complementing Foldseek \cite{van2024foldseek} structural alignment and sequence homology tools (BLAST \citemethods{altschul1990blast} and MMseqs2 \citemethods{kallenborn2025mmseqs2-gpu}), we deployed VenusMine \cite{wu2025venusmine}, an advanced embedding-based retrieval engine. By synergizing sequence and structural representations through multi-round iterative searches, VenusMine facilitates the identification of remote homologs beyond the reach of local alignment, a capability validated by the discovery of high-performance PET hydrolases.

\textbf{Biophysical Filtration Suite.}
To facilitate the rigorous down-selection of protein candidates, we established a comprehensive filtration suite within the \vf~platform, synergizing deep learning with classical biophysics. Leveraging diverse PLMs as foundational encoders, we integrated an ensemble of predictive modules to model critical operational constraints. These include predictors for manufacturability (binary solubility status \citemethods{Khurana2018deepsol,tan2024protsolm}, quantitative solubility score \citemethods{chen2021esol}, membrane localization \citemethods{almagro2017deeploc}, subcellular localization \citemethods{thumuluri2022deeploc2}, signal peptides \citemethods{odum2024deeploc2.1}), environmental stability (thermostability \citemethods{tan2024sesadapter}, optimal catalytic temperature \citemethods{li2022deepet}, optimal pH \citemethods{gado2025ephod}), catalytic efficiency (catalytic constant $k_{cat}$ \citemethods{li2022dlkcat}, metal ion binding \citemethods{tan2024sesadapter}), and safety profile (immunogenicity \citemethods{li2025VenusVaccine}). Complementing these data-driven models, the system deploys established bioinformatics utilities to calculate fundamental physicochemical descriptors, including solvent accessible surface area, relative rolvent accessibility, and secondary structure composition based on Biopython \citemethods{cock2009biopython}. (see \textbf{Supplementary Table~\ref{app:tab:protein_discovery_tools}}).

\subsection*{Auto Machine Learning}
\textbf{Template-based Model Adaptation.}
To adapt foundation PLMs for specialized downstream tasks, \vf~implements a rigorous optimization framework that bridges general-purpose representations with specific scientific objectives. This infrastructure enables a continuum of tasks—from continuous property regression to discrete classification—governed by strict configuration schemas. By encapsulating model specifications within static \texttt{.config} definitions (Appendix Table~\ref{app:tab:full_config_schema}), the system constrains the generative action space, thereby ensuring the syntactic validity and executability of autonomously synthesized code.

To balance computational efficiency with feature alignment, the agent supports a spectrum of training strategies, ranging from full-parameter updates to parameter-efficient fine-tuning (PEFT) \citemethods{mangrulkar2023peft}. The PEFT infrastructure is integrated into \vf~to support the LoRA family, optimizing resource consumption without compromising expressivity. Specifically, the system supports: (1) standard LoRA \citemethods{hu2022lora} and weight-decomposed LoRA (DoRA) \citemethods{liu2024dora} for maximizing learning capacity; (2) adaptive budgeting via AdaLoRA \citemethods{zhang2023adalora} for sparsity control; and (3) quantization-aware QLoRA \citemethods{dettmers2023qlora} and IA3 \citemethods{liu2022lora_ia3} for deployment in resource-constrained environments. For structure-dependent tasks, the agent additionally incorporates the SES-Adapter \citemethods{tan2024sesadapter}, which employs cross-attention to explicitly fuse sequence representations with structural embeddings derived from Foldseek \cite{van2024foldseek} or DSSP \citemethods{kabsch1983dssp}.

For the projection phase, the architecture bifurcates its processing logic based on the granularity of the biological objective. For protein-level tasks (e.g., stability or solubility prediction), the system aggregates the raw residue representations $H \in \mathbb{R}^{L \times d}$ into a global latent embedding to predict the target property $\hat{y}$:

\begin{equation}
\hat{y} = \mathcal{F}_{\text{task}} \left( \Psi_{\text{agg}}(H) \right), \quad \text{where } \Psi_{\text{agg}} \in \{ \text{MeanPool}, \text{LightAttn}, \text{Attn1D} \}
\end{equation}

Here, $\Psi_{\text{agg}}$ denotes the dynamically selected aggregation operator—ranging from baseline mean pooling to complex attention-based convolutions—and $\mathcal{F}_{\text{task}}$ represents the decoding Multi-Layer Perceptron. Conversely, for residue-level tasks (e.g., binding site identification), the system bypasses temporal pooling to maintain topological resolution, mapping embeddings directly to sequence-aligned labels $\hat{Y} \in \mathbb{R}^{L \times K}$:

\begin{equation}
\hat{Y} = \text{Softmax} \left( H W_{\text{token}} + b_{\text{token}} \right)
\end{equation}

where $W_{\text{token}} \in \mathbb{R}^{d \times K}$ and $b_{\text{token}} \in \mathbb{R}^{K}$ are the learnable parameters of the token-wise projection layer, and $K$ represents the number of functional classes.

\textbf{\textit{De novo} Code Generation.}
Complementing the template-based infrastructure, the framework supports a generative modeling paradigm for scenarios necessitating bespoke architectures or lightweight tabular learning. Diverging from the rigid config-driven adaptation of PLMs, this module capitalizes on the inherent code generation proficiencies of the underlying LLM to synthesize fully autonomous, self-contained executable scripts. Upon analyzing the user's data topology and input-output context, the agent dynamically authors and persists Python-based modeling pipelines as discrete files. This mechanism allows for the \textit{de novo} construction of diverse architectures—from classical regressors (e.g., XGBoost, Random Forest) to custom neural networks—ensuring execution logic is strictly tailored to immediate scientific constraints without the architectural bias of pre-trained backbones.

\subsection*{Supported Trainable PLMs}
To ensure broad adaptability across diverse biological tasks and computational environments, \vf~integrates a heterogeneous repository of pre-trained foundation models (see \textbf{Supplementary Table~\ref{app:tab:supported_models}}). This collection encompasses architectures ranging from lightweight encoders (8M parameters) to large-scale generative models (15B parameters), enabling deployment on hardware ranging from consumer GPUs to cluster nodes. For general-purpose sequence representation, the platform supports the full evolutionary scale modeling (ESM) family \citemethods{lin2023esm2,meier2021esm1v,rives2021esm1b} and BERT-style architectures \citemethods{elnaggar2021prottrans,tan2023peta}, alongside Encoder-Decoder frameworks such as ProtT5 \citemethods{elnaggar2021prottrans} and Ankh series \citemethods{elnaggar2023ankh,alsamkary2025ankh3}. To address domain-specific engineering campaigns, the library includes antibody-centric models like IgBert and IgT5 \citemethods{kenlay2024igbert}. Crucially, to transcend the limitations of 1D sequence modeling, the system incorporates structure-informed architectures—including SaProt \cite{su2023saprot}, ProSST \citemethods{li2024prosst}, and ProtSSN \citemethods{tan2025protssn}—which fuse geometric priors into the embedding space via structure-aware vocabularies or disentangled attention mechanisms, facilitating the optimization of folding stability and epistatic interactions. 

\subsection*{Built-in Datasets Description}

To operationalize the AI agent as a reliable co-scientist for wet-lab experimentation, we curated a standardized repository of high-quality biological datasets tailored for practical protein engineering. Unlike static benchmarks designed solely for algorithmic comparison, this infrastructure aggregates a curated portfolio of established, peer-reviewed datasets to model physical constraints, serving as a heuristic sieve to mitigate biophysical risks (\textbf{Supplementary Table~\ref{app:tab:bio_utility}}). The collection is stratified into four engineering tiers: manufacturability and trafficking, supported by datasets such as DeepSol \citemethods{Khurana2018deepsol}, DeepSoluE \citemethods{wang2023deepsolue}, ProtSolM \citemethods{tan2024protsolm}, eSOL \citemethods{chen2021esol}, and DeepLoc suite \citemethods{almagro2017deeploc,thumuluri2022deeploc2,odum2024deeploc2.1} to check soluble expression and localization; environmental adaptation, including Thermostability \citemethods{tan2024sesadapter}, DeepET \citemethods{li2022deepet}, and EpHod \citemethods{gado2025ephod} for optimizing thermal and pH stability; catalytic competence, utilizing DLKcat \citemethods{li2022dlkcat} and MetallonBinding \citemethods{tan2024sesadapter} for kinetic screening; and safety integrity, incorporating VenusVaccine \citemethods{li2025VenusVaccine} and the VenusX suite \citemethods{tan2026venusx} to mitigate immunogenicity and maintain functional motifs.

To ensure seamless integration with the autonomous code generation module, all datasets adhere to a unified schema that harmonizes heterogeneous biological information into a machine-readable format (\textbf{Supplementary Table~\ref{app:tab:data_example}}). This schema mandates essential fields for sequences and labels while accommodating advanced structural descriptors—such as Foldseek \cite{van2024foldseek} tokens and DSSP \citemethods{kabsch1983dssp} sequences—to leverage structure-aware adapter. Furthermore, to maintain statistical rigor and prevent information leakage, the repository is pre-partitioned into strict training, validation, and testing subsets (\textbf{Supplementary Table~\ref{app:tab:dataset_stats}}). This comprehensive infrastructure covers diverse task types, ranging from binary classification to continuous regression, providing the foundational knowledge base for the agent to autonomously construct and deploy customized predictive models.

\subsection*{Agentic Built-in Tools Explore}
To evaluate the autonomous exploration capabilities of our framework, we analyzed its architectural and hyperparameter search process across diverse downstream protein property prediction tasks. Normalizing the primary evaluation metrics (Appendix Figure~\ref{app:fig:search_space}) reveals substantial variance in performance distributions across datasets. This highlights that model efficacy depends heavily on specific combinations of PLMs and fine-tuning strategies. Despite these significant performance variations, the agent consistently identified the optimal configuration. To analyze the temporal dynamics of this optimization, we tracked the agent's search trajectories on a representative subset of datasets (DLKcat, DeepLoc2Multi, TAPE\_Fluorescence, and DeepSoluE) that encompass a broad spectrum of biological properties and task complexities. As depicted in Appendix Figure~\ref{app:fig:trajectory}, the agent exhibits a directed and sample-efficient convergence. It progressively refines its configurations and systematically outperforms static baselines derived from standard human-selected heuristics within a constrained trial budget.

This optimization efficiency is directly related to the agent's learned exploration preferences (Appendix Figure~\ref{app:fig:preferences}). Rather than sampling the configuration space uniformly, the agent disproportionately allocated its trial budget toward specific PLM architectures, notably the esm2\_t30\_150M and ankh-base models, while predominantly utilizing the parameter-efficient ses-adapter tuning method. This skewed empirical distribution indicates an implicit cost-aware optimization strategy. By autonomously prioritizing parameter-efficient fine-tuning over full-model freezing and favoring medium-sized foundation models over their larger counterparts, the agent effectively balances computational overhead with predictive accuracy. It directs its search budget toward configurations with the highest empirical yield, demonstrating a systematic approach to resource allocation in protein representation learning.

\subsection*{Construction of VenusAgentEval}

\textbf{Benchmark Rationale and Positioning.} To systematically evaluate the specialized capabilities of~\vf~in protein engineering, we established VenusAgentEval. Distinct from general-domain reasoning benchmarks \citemethods{center2026hle_human_last_exam, du2026deepresearch_bench} and broad scientific tool-use frameworks \cite{ding2025scitoolagent}, VenusAgentEval stands as the first vertical agentic benchmark dedicated to protein science, specifically targeting the sub-domains of protein discovery and directed evolution.

\textbf{Hierarchical Construction Strategy.} Addressing the challenge of synthesizing tasks with both scientific depth and quantifiable metrics, we adopted a hierarchical three-stage construction strategy comprising multi-agent generation, adversarial dialectical verification, and expert adjudication.

\textbf{Heterogeneous Multi-Agent Generation.} To mitigate single-model knowledge bias \citemethods{wang2023self-instruct}, we assembled a consortium of four advanced LLMs (GPT-4 \citemethods{achiam2023gpt4}, Claude-3.7 \citemethods{anthropic2024claude}, Gemini-2.5-Pro \citemethods{comanici2025gemini2.5}, and DeepSeek-R1 \citemethods{guo2025deepseek}). Configured with a unified prompt simulating a senior computational biologist, this ensemble engaged in a parallel role-playing protocol. Over 100 interaction rounds, the system generated a preliminary corpus of 400 candidate questions, leveraging distinct reasoning architectures to enhance stylistic diversity and domain coverage.

\textbf{Adversarial Dialectical Verification.} To enforce logical consistency and biological plausibility, candidates were subjected to an automated adversarial debate mechanism modeled after the academic peer-review process \citemethods{liang2024llm_peer_review}. For each question $q$, a ``Review Committee'' of non-generating agents independently evaluated scientific significance and self-consistency, prompting the ``Author'' agent to iteratively refine the content. Upon revision, the committee assigned a quantitative quality score $S_m(q) \in [0, 5]$, with the final retention metric $S_{final}(q)$ calculated as:
\begin{equation}
S_{final}(q) = \sum_{m \in C} S_m(q)
\end{equation}
where $C$ denotes the set of reviewing agents. This dialectical optimization effectively filtered out formally plausible yet biologically shallow instances, selecting the top 200 high-scoring questions.

\textbf{Expert Adjudication and Stratification.} In the final validation stage, the verified dataset underwent a rigorous single-blind review by three independent biology Ph.D. experts. This human-in-the-loop mechanism \citemethods{chen2021human_eval} discarded entries with factual errors and stratified surviving questions by cognitive load into \textbf{Question-level} (atomic retrieval), \textbf{Task-level} (multi-step workflow), and \textbf{Project-level} (long-horizon planning). Following a majority-vote selection rule, the final VenusAgentEval dataset comprises 148 curated instances (58 Question-level, 60 Task-level, and 30 Project-level), establishing a benchmark of high scientific rigor (detailed in \textbf{Supplementary Tables~\ref{app:tab:venusagenteval_question}-\ref{app:tab:venusagenteval_project}}).

\subsection*{Agents Evaluation Strategy}

\textbf{Pointwise Reasoning for Pairwise Evaluation.} To address the stochastic nature of biological tool invocation and the limitations of standard LLM-based evaluation—specifically calibration errors in pointwise scoring and verbosity bias in pairwise comparisons—we implemented a decoupled evaluation architecture termed \textit{Pointwise Reasoning for Pairwise Evaluation}. This framework separates evidence extraction from final adjudication to ensure objectivity. In the analytic phase, an \textit{Agent Analyst} (based on GPT-4o-mini) decomposes response pairs based on strict criteria: scientific validity, evidence sufficiency, and logical consistency. Crucially, this agent produces structured qualitative critiques rather than numerical scores, isolating factual accuracy from stylistic presentation. Subsequently, an independent judge instance synthesizes these evidence-based audits to determine a \textit{winner}, thereby grounding the adjudication in verified content rather than surface-level fluency.

\textbf{Rank-Weighted Scoring Protocol.} To quantify performance across the model cohort, we employed a scoring metric derived from exhaustive pairwise interactions. For a given query, models are ranked according to their total victory frequency. To penalize mediocrity while rewarding consistent superiority, we apply a linear rank-weighting scheme. Let $N$ denote the total number of competing models, $W_m$ the total win count for model $m$, and $r_m$ the specific rank of the model derived from $W_m$ (where $r=1$ indicates the highest win count). The final normalized capability score $S_m$ is calculated as:

\begin{equation}
S_m = \frac{W_m \times (N - r_m + 1)}{N}
\end{equation}

The reported performance metric represents the mean $S_m$ across the test corpus, providing a robust indicator of relative scientific reasoning capability that accounts for both win rate and comparative standing.

\subsection*{Selection of the Reasoning Backbone}

To identify the optimal reasoning engine for \vf, we evaluated four state-of-the-art foundation models: Claude-3.7-Sonnet \citemethods{anthropic2024claude}, DeepSeek-Chat \citemethods{guo2025deepseek}, GPT-4o-mini \citemethods{hurst2024gpt4o}, and Gemini-2.5-Pro \citemethods{comanici2025gemini2.5}. All models were assessed on the VenusAgentEval benchmark using official API configurations. The evaluation protocol focused on two critical dimensions:

\textbf{Strategic Planning Assessment.} This dimension evaluates the generative quality of the \textit{Principal Investigator} module, focusing on (1) \textbf{Feasibility}, ensuring the proposed scheme forms a closed execution loop without invoking non-existent tools; (2) \textbf{Logical Adequacy}, verifying coherent sequencing and the absence of step-skipping; and (3) \textbf{Efficiency}, identifying and penalizing redundant subtasks such as superfluous structural retrieval for sequence-based predictions.

\textbf{Scientific Response Assessment.} This dimension evaluates the output of the \textit{Scientific Critic}, prioritizing (1) \textbf{Correctness}, gauging the accuracy of reasoning-based task completion; (2) \textbf{Causal Completeness}, measuring the integrity of inference chains; (3) \textbf{Hallucination Control}, scrutinizing outputs for fabricated biological entities; and (4) \textbf{Constraint Adherence}, ensuring strict compliance with prompt instructions.

We employed a two-stage evaluation strategy where qualitative profiles from the initial analysis were fed into a second-stage judge model for pairwise comparison. As illustrated in Fig.\ref{app:fig:venus_eval}, Claude-3.7-Sonnet demonstrated superior performance across all cognitive tiers, achieving normalized scores of 0.79 (Question-level), 0.85 (Task-level), and 0.83 (Project-level). Rank distribution analysis further confirmed this stability, with the model securing the highest frequency of first-place rankings. Consequently, Claude-3.7-Sonnet was established as the designated reasoning backbone for \vf~in all subsequent experiments.

\subsection*{Baseline Agents}
To benchmark the architectural efficacy of \vf, we established a comparative framework comprising nine state-of-the-art generalist LLMs and two domain-specific autonomous agents. The generalist cohort included GPT-5.2 \citemethods{singh2025gpt5}, GPT-4o-mini \citemethods{hurst2024gpt4o}, DeepSeek-V3.2-Thinking, DeepSeek-V3.1, DeepSeek-chat \citemethods{guo2025deepseek}, Claude-4.5-sonnet-thinking, Claude-3.7-sonnet \citemethods{anthropic2024claude}, Gemini-3-Pro, and Gemini-2.5-Pro \citemethods{comanici2025gemini2.5}. All models were accessed via official APIs under a unified system prompt designating a bioinformatics expert persona, with external internet access disabled to isolate intrinsic reasoning capabilities. For domain-specific comparisons, we evaluated ProtAgent \cite{ghafarollahi2024protagents} and SciToolAgent \cite{ding2025scitoolagent}, reproducing both systems locally from their official repositories. The baseline agents were integrated with the identical computational tool library used in \vf~and utilized GPT-4o as the underlying reasoning backbone. This setup strictly controls for variances in tool availability and foundation model capability, thereby isolating the performance differentials attributable to workflow architecture and system design.

\subsection*{Comparison with AI Agents}
Evaluation was conducted under a multidimensional criteria set designed to reflect the rigors of biological research: information completeness, requiring concise, conclusion-first explanations devoid of unsupported speculation; evidence grounding, mandating reliance on traceable literature or reproducible tool outputs; practical usefulness, prioritizing actionable outcomes over abstract planning; and user accessibility, ensuring linguistic alignment and the provision of constructive follow-up inquiries. As illustrated in Fig.\ref{fig:framework}d, \vf~demonstrated superior performance across all cognitive tiers, achieving weighted scores of 88.8 (Question-level), 74.7 (Task-level), and 78.8 (Project-level). While generalist models such as Claude-3.7-sonnet and Gemini-3-Pro exhibited competence in atomic question-answering, their performance variance increased significantly with expanding task horizons and procedural complexity. In contrast, \vf~maintained consistent stability across multi-step executions, validating the necessity of structured intermediate reasoning.

Furthermore, the domain-specific baselines, ProtAgent and SciToolAgent, yielded significantly lower weighted scores, particularly at the question and task levels. Despite utilizing the same toolset and reasoning engine, these systems frequently failed to maintain evidence chains or strictly adhere to prompt constraints, underscoring that workflow orchestration is a determinant factor in agentic performance. Beyond quantitative metrics, \vf~exhibited superior error handling capabilities. Notably, the system autonomously detected input inconsistencies—such as residue positions conflicting with the target sequence—and preemptively halted execution to report the premise error (Supplementary Table~\ref{app:tab:benchmark_comparison_error_query}). This capacity for input validation directly enhances the reliability of downstream experimental designs by preventing the propagation of erroneous data.

\subsection*{System Architecture and Agent Orchestration}

\textbf{Overview.}
\vf~operates as a closed-loop multi-agent system designed to automate protein engineering workflows. The architecture transforms abstract natural language queries into executable bioinformatics pipelines through a centralized orchestration mechanism. The workflow follows a sequential logic: (1) task decomposition and planning, (2) code generation and tool execution, (3) autonomous error correction, and (4) result synthesis and reporting. This process is collaboratively executed by four specialized agents: the Principal Investigator, the Machine Learning Specialist, the Computational Biologist, and the Scientific Critic. The specific implementation and functional responsibilities of each agent are detailed below.

\textbf{Principal Investigator.}
The Principal Investigator serves as the central planner, responsible for interpreting research intent and architectural task decomposition. Upon receiving a user query, this agent performs three critical functions: (1) Intent Analysis: It parses the input to extract key entities, such as target sequences or database identifiers, and determines the necessary analysis paradigms. (2) Feasibility Assessment: By cross-referencing the internal tool library, it constructs a directed acyclic graph (DAG) of analysis steps, defining input-output dependencies and execution priorities. (3) Context Management: To support iterative optimization, the agent implements a two-level memory mechanism. At the interaction level, it retains conversational context to resolve ambiguities in multi-turn dialogue. At the execution level, it serializes and stores intermediate results and parameter configurations, allowing the system to bypass redundant computations in subsequent steps.

\textbf{Machine Learning Specialist.}
The Machine Learning Specialist functions as the algorithmic engine, tasked with the lifecycle management of predictive models. Its core responsibilities include: (1) Dynamic Code Generation: It translates the PI's abstract requirements into executable Python scripts, implementing architectures ranging from lightweight regression models for multi-site variant scoring to fine-tuning large protein language models on domain-specific datasets. (2) Automated Debugging: The agent monitors runtime logs in real-time. Upon detecting exceptions or convergence failures, it triggers a self-correction loop to diagnose error sources, adjust hyperparameters, or revise data preprocessing logic without human intervention. (3) Tool Registration: Post-training, the agent encapsulates model weights and inference interfaces into standardized tools, registering them into the system library for immediate invocation by other agents.

\textbf{Computational Biologist.}
The Computational Biologist acts as the interface for biological data acquisition and standard bioinformatics tool execution. Guided by the central plan, this agent executes the following tasks: (1) Multimodal Data Retrieval: It queries external repositories—such as sequence databases, structural archives, and literature indices—to curate training data or reference sets. (2) Tool Orchestration: It manages the execution of standard bioinformatics utilities, including sequence alignment, structural folding, and functional annotation tools, ensuring proper parameterization and resource allocation. (3) Data Standardization: To address format inconsistencies common in biological data, the agent implements adaptive parsing logic, converting heterogeneous outputs into unified data structures compatible with downstream machine learning modules.

\textbf{Scientific Critic.}
The Scientific Critic serves as the quality assurance and synthesis module. Its primary function is to transform raw computational outputs into coherent scientific insights. This involves: (1) Logical Verification: During execution, the agent audits intermediate results against the initial research objectives, flagging inconsistencies, missing data, or biological implausibility for re-evaluation. (2) Result Integration: It aggregates disparate data streams—including model metrics, structural files, and functional predictions—into a consolidated analysis record. (3) Narrative Reconstruction: Finally, the agent synthesizes a structured report that summarizes methodology, interprets key findings with supporting evidence, and proposes data-driven directions for subsequent experimental validation.

\subsection*{Prompt Engineering and Formalized Role Definition}

To ensure deterministic execution within the stochastic multi-agent environment, we formalized the prompt engineering framework as a structured mapping function $\Phi$, which projects natural language directives into constrained agentic behaviors. This framework is governed by a set of role-specific templates $\mathcal{T} = \{T_{PI}, T_{ML}, T_{Bio}, T_{SC}\}$, designed to enforce strict output schemas, operational traceability, and evidence-based reasoning. The planning phase is mathematically modeled as a decomposition function driven by the Principal Investigator ($\mathcal{A}_{PI}$). Given a user query $U$, the agent generates a sequential execution plan $\mathcal{P}$:

\begin{equation}
\mathcal{P} = \mathcal{A}_{PI}(U \mid T_{PI}) = \{ (s_t, \tau_t, \theta_t) \}_{t=1}^{N}
\end{equation}

where $s_t$ denotes the natural language description of the sub-task at step $t$, $\tau_t$ represents the specific tool identifier mapped from the system library, and $\theta_t$ constitutes the validated parameter set derived from the user's intent. This mapping enforces a strict dependency structure where $\theta_t$ may reference outputs $o_{k<t}$ from preceding steps. Subsequently, the execution phase delegates these tasks to the domain-specific executors, the Machine Learning Specialist ($\mathcal{A}_{ML}$) and the Computational Biologist ($\mathcal{A}_{Bio}$). The state transition of the workflow is defined by the iterative update of the execution history $\mathcal{H}$:

\begin{equation}
\mathcal{H}_t = \mathcal{H}_{t-1} \cup \{ (o_t, \lambda_t) \} \quad \text{where} \quad (o_t, \lambda_t) = \text{Exec}_{\tau_t}(\theta_t \mid \mathcal{H}_{t-1}, T_{exec})
\end{equation}

Here, $o_t$ represents the structured output artifacts (e.g., model weights, sequences, or PDB files), and $\lambda_t$ denotes the execution trace logs required for reproducibility. Finally, the synthesis phase is governed by the Scientific Critic ($\mathcal{A}_{SC}$), which aggregates the cumulative history $\mathcal{H}_N$ into a coherent scientific narrative $R$:

\begin{equation}
R = \mathcal{A}_{SC}(\mathcal{H}_N \mid T_{SC}, \mathcal{C}_{ver})
\end{equation}

subject to verification constraints $\mathcal{C}_{ver}$ that strictly bound generated claims to the evidence provided in $\mathcal{H}_N$. Full definitions of these prompt templates and constraint schemas are detailed in Supplementary Section~\ref{app:sec:prompts}.

\subsection*{Autonomous Tool Creation and Self-Evolving Workflows}

Developing task-specific predictive tools in protein engineering traditionally demands profound cross-disciplinary expertise. Specifically, fine-tuning modern protein language models presents significant operational bottlenecks due to the necessity of navigating high-dimensional hyperparameter spaces, configuring training pipelines, and managing heterogeneous datasets. Consequently, optimal predictive performance is frequently inaccessible to researchers lacking specialized algorithmic knowledge. 

To overcome these barriers, \vf~features a closed-loop, self-evolving workflow that empowers its multi-agent system to autonomously orchestrate the end-to-end model development lifecycle. Upon encountering requests beyond the scope of its existing tool library, the system avoids execution failure by dynamically standardizing input data, selecting appropriate foundational architectures, and formulating optimized training scripts. By abstracting away these engineering complexities, \vf~transforms static tool libraries into a dynamic repository, consistently yielding customized predictors and packaging them as standard Model Context Protocol (MCP) services for future reuse.

To empirically validate this workflow, we conducted a case study in \textit{de novo} allergenicity predictor synthesis. The campaign was initiated with the following prompt:

\begin{quote}
\textit{``Please evaluate the allergenicity potential for these modified food enzymes and the protein sequence is in the uploaded file.''}
\end{quote}

Upon receiving the request, the \textit{Principal Investigator} identified the intent as protein allergenicity analysis. After cross-referencing the tool library, it recognized a methodological gap: existing modules handled general functional annotation but lacked a dedicated allergenicity predictor. To resolve this, the agent autonomously decided to synthesize a novel tool and generated a complete execution plan:

\begin{quote}
Step 1: Search papers related to allergenicity prediction.\\
Step 2: Use web search to find datasets for allergen / non-allergen prediction.\\
Step 3: Split the downloaded training data into an 8:2 training and validation set.\\
Step 4: Generate the training configuration file for the ESM2-8M model with LoRA.\\
Step 5: Train the model using the generated configuration and report performance.\\
Step 6: Package the trained checkpoint as a new MCP service and register it into the \vf~tool registry.\\
Step 7: Use the newly minted MCP tool to predict the sequences in the provided FASTA file.
\end{quote}

Following this blueprint, the \textit{Computational Biologist} and \textit{Machine Learning Specialist} collaborated to execute the plan. They began with keyword searches across PubMed and arXiv, utilizing the summarized findings to guide dataset retrieval from open platforms such as Hugging Face. Upon processing the dataset into the requisite 8:2 split, the \textit{Machine Learning Specialist} selected the ESM2-8M architecture and autonomously fine-tuned it via LoRA for two epochs, monitoring convergence metrics without human intervention. 

Once training concluded, \vf~reported an accuracy of 0.92 and an AUROC of 0.96 on the independent test set. The agent subsequently packaged the model checkpoint, inference schemas, and metadata into a new MCP tool, registering it within the platform library. Finally, \vf~deployed this synthesized tool on the user's uploaded FASTA sequence, while the \textit{Scientific Critic} translated the technical metrics into an accessible final report. Through this process, a transient experimental requirement was transformed into a permanent, discoverable computational asset.


\subsection*{Adaptive Directed Evolution of a VHH Antibody}

Protein directed evolution constitutes a formidable challenge in protein engineering, historically bottlenecked by the exponential growth of the mutational landscape. Even screening single-point mutations demands rigorous evaluation of structural constraints, functional effects, and experimental feasibility. To demonstrate how \vf~supports iterative combinatorial optimization, we deployed the system in a directed evolution campaign aimed at enhancing the binding affinity of a VHH antibody.

The process was initiated via a natural-language directive specifying the wild-type target:

\begin{quote}
\textit{``I have a specific VHH antibody with the protein sequence \seqsplit{MQVQLVESGGGLAQAGGSLRLSCAVSGMPEFARAMGWFRQAPGKERELLAAIEGIGATTYYADSVKGRFTISRDDAANTVLLQMNSLKPDDTAVYYCAAAFSVTIPTRARHWVDWGPGTLVTVSSDDDDKSGGGGSHHHHHH}. The current goal is to improve the binding affinity of this antibody.''}
\end{quote}

Upon receiving this request, the \textit{Principal Investigator} interpreted the objective as a structure-guided affinity maturation task and formulated the following execution protocol:

\begin{quote}
Step 1: Predict the three-dimensional conformation of the provided VHH sequence using AlphaFold2 to establish a structural baseline.\\
Step 2: Perform structural homology mining across comprehensive databases using FoldSeek to curate an evolutionary profile.\\
Step 3: Execute a zero-shot, \textit{in silico} saturated single-point mutational scan utilizing the VenusREM architecture.
\end{quote}

Orchestrating this blueprint, the \textit{Computational Biologist} first resolved the structure via AlphaFold2 \cite{jumper2021alphafold2} and retrieved structural homologs from databases including AFDB and PDB via FoldSeek \cite{van2024foldseek}. These multimodal inputs—sequence, predicted structure, and homology alignments—were autonomously routed to VenusREM \cite{tan2025venusrem}. The \textit{Scientific Critic} then analyzed the resulting zero-shot fitness landscape, isolating high-potential single-point mutation sites to guide the first round of wet-lab validation.

Following the acquisition of first-round experimental fitness data, the human researcher uploaded the results to the platform to initiate the combinatorial optimization phase:

\begin{quote}
\textit{``For this VHH, I have completed single-point mutation experiments. The experimental data are in the uploaded folder. Please recommend higher-order mutation sites based on these results.''}
\end{quote}

Recognizing that standard zero-shot tools lack the capacity to natively model complex epistatic interactions from user-provided empirical data, the \textit{Principal Investigator} autonomously pivoted the computational strategy to bridge this methodological gap:

\begin{quote}
Step 1: Dynamically synthesize and train a regularized Ridge Regression model using the uploaded empirical data to predict combinatorial epistatic effects.\\
Step 2: Batch-evaluate higher-order mutants and rank the top-performing combinatorial candidates.
\end{quote}

Tasked with this objective, the \textit{Machine Learning Specialist} autonomously parsed the uploaded wet-lab data, applied one-hot encoding to the mutant sequences, and trained a regularized Ridge Regression model \cite{hsu2022few_shot_ridge_regression}. This supervised architecture effectively mapped single-point empirical priors to epistatic contributions, mitigating overfitting. Leveraging this autonomously generated model, the \textit{Scientific Critic} executed high-throughput batch inference across the combinatorial sequence space, ranking and recommending the top-performing double, triple, and quadruple mutants.

Retrospective analysis against ground-truth wet-lab binding data confirmed that the agent-synthesized regression model achieved significantly higher experimental correlation for multi-point variants than a naive linear summation of single-point effects. This campaign underscores \vf's capacity to autonomously recognize methodological limitations and dynamically synthesize bespoke algorithmic solutions, functioning as an adaptive computational co-scientist for iterative biological optimization.


\subsection*{End-to-End Protein Discovery and Precision Mining}

Navigating billion-scale metagenomic spaces to discover novel, industrially relevant enzymes traditionally requires assembling disjointed bioinformatics pipelines and manually configuring complex filtration thresholds. To enable comprehensive discovery, \vf~integrates a massive, deduplicated sequence repository—spanning databases like GOPC \citemethods{carradec2018gopc}, UniProt \cite{uniprot2025uniprot}, MGnify \citemethods{richardson2023mgnify}, and OMG \citemethods{cornman2025omg}—with an agent-orchestrated screening infrastructure powered by VenusMine \cite{wu2025venusmine}. Rather than relying on static filters, \vf~empowers its agents to dynamically assemble structure-aware search and biophysical screening protocols directly from natural-language objectives.

To validate this end-to-end capability, we conducted a blind retrieval campaign for polyethylene terephthalate (PET) hydrolases. The discovery process was initiated with a purely functional description rather than a traditional template sequence:

\begin{quote}
\textit{``Please search the metagenomic databases to discover a novel PET hydrolase that exhibits significantly higher thermostability and catalytic activity than the standard IsPETase.''}
\end{quote}

Upon receiving this prompt, the \textit{Principal Investigator} recognized the intent as a \textit{de novo} enzyme discovery task requiring both homology mining and stringent biophysical filtration. It subsequently formulated a comprehensive execution strategy:

\begin{quote}
Step 1: Search for scientific literature on PETase to provide an overview and background information.\\
Step 2: Retrieve the protein sequence and information for the well-studied PETase from Ideonella sakaiensis using its UniProt ID .\\
Step 3: Download the predicted protein structure for PETase from the AlphaFold database using its UniProt ID.\\
Step 4: Conduct enzyme mining using VenusMine.\\
Step 5: Screen candidates for thermostability. \\
Step 6: Screen candidates for solubility. \\
Step 7: Rank the surviving candidates and compile the final enzyme discovery report.
\end{quote}

Following this blueprint, the agents autonomously executed the multi-stage pipeline. The \textit{Research} module first identified a reliable seed PETase from the literature. Transitioning to sequence space expansion, the \textit{Computational Biologist} employed FoldSeek \cite{van2024foldseek} for structural alignment against the AlphaFold Database, followed by a cascade expansion into the million-scale database using MMseqs2 \citemethods{kallenborn2025mmseqs2-gpu} to construct a non-redundant library. To distill this vast pool, the system leveraged ProstT5 \citemethods{heinzinger2023prostt5}. Operating in sequence-to-structure translation mode, it encoded candidates into high-dimensional embeddings, allowing the agent to cluster sequences based on conserved catalytic core architectures while discarding structurally divergent hits.

Crucially, during the \textit{in silico} screening phase, the \textit{Computational Biologist} departed from static retrieval protocols. Interpreting the user's explicit requirement for high thermostability and activity, it dynamically selected predictive assays from the Biophysical Filtration Suite, deploying fine-tuned ESM-2 \citemethods{lin2023esm2} for melting temperature ($T_m$), ProtSolM \citemethods{tan2024protsolm} for solubility, and AlphaFold2 \cite{jumper2021alphafold2} to verify structural fidelity.

Through this agent-mediated orchestration, \vf~successfully narrowed the vast candidate space down to KbPETase (APET-14) \cite{wu2025venusmine}, a novel enzyme from \textit{Kibdelosporangium banguiense}. The discovered enzyme exhibited exceptional experimental metrics: a melting temperature ($T_m$) of 80.1°C, exceeding IsPETase \citemethods{burgin2024ispet} by 32°C; a catalytic activity at 50°C that is 97x that of IsPETase and 5.7x that of the thermophilic benchmark LCC; and a catalytic efficiency ($k_{\text{cat}}/K_m$) 1.87x and 1.3x higher than those of the engineered FastPETase \citemethods{cui2024fastpet} and LCC, respectively. This successful recapitulation demonstrates how \vf~translates a declarative natural language prompt into the precision retrieval of high-performance enzymes.

\section*{Data availability}
The downstream tasks dataset is available via Huggingface at \url{https://huggingface.co/AI4Protein}.

\section*{Code and Server availability}
To accommodate diverse computational environments and user proficiencies, \vf~is deployed across multiple access interfaces, including a local command-line interface (CLI) and Python API for high-throughput programmatic execution, alongside interactive Jupyter Notebooks. For no-code exploration and autonomous multi-agent orchestration, the graphical web server and conversational agent interface are freely accessible at \url{https://venusfactory.cn/playground/}. All associated source code, underlying model weights, and comprehensive documentation are provided for academic use at \url{https://github.com/ai4protein/VenusFactory2}.

\section*{Acknowledgement}
This work was supported by the National Key Research and Development Program of China (2024YFA0917603), Shanghai Municipal Science and Technology Major Project, the AI for Science Program, Shanghai Municipal Commission of Economy and Informatization (2025-GZL-RGZN-BTBX-02009), the Computational Biology Key Program of Shanghai Science and Technology Commission (23JS1400600), Shanghai Jiao Tong University Scientific and Technological Innovation Funds (21X010200843), and Science and Technology Innovation Key R\&D Program of Chongqing (CSTB2022TIAD-STX0017, CSTB2024TIAD-STX0032), the Student
Innovation Center at Shanghai Jiao Tong University, and Shanghai Artificial Intelligence Laboratory.

\begingroup
    \makeatletter
    \let\oldthebibliography\thebibliography
    \let\endoldthebibliography\endthebibliography
    
    \renewenvironment{thebibliography}[1]{%
        \oldthebibliography{#1}%
        \addtocounter{enumiv}{16}%
        
    }{\endoldthebibliography}
    \makeatother
    \bibliographystylemethods{unsrt}
    \bibliographymethods{1reference}
\endgroup

\newpage
\begin{appendices}

\section{Tools}\label{app:sec:tools}

\small
  \renewcommand{\arraystretch}{1.3}
  \setlength{\LTcapwidth}{\textwidth}
  \begin{xltabular}{\textwidth}{@{}
  >{\raggedright\arraybackslash}p{0.25\textwidth}
  >{\raggedright\arraybackslash}X @{}}

  \caption{Comprehensive deep research toolkit integrating 26 federated
  intelligence sources and 14 biological database APIs.}
  \label{app:tab:deep_search_tools} \\
  \toprule
  \textbf{Tool} & \textbf{Usage} \\
  \midrule
  \endfirsthead

  \multicolumn{2}{@{}l}{\textit{Table \thetable{} continued from previous
  page}} \\
  \toprule
  \textbf{Tool} & \textbf{Usage} \\
  \midrule
  \endhead

  \midrule
  \multicolumn{2}{r@{}}{\textit{Continued on next page...}} \\
  \endfoot

  \bottomrule
  \endlastfoot

  \multicolumn{2}{@{}l}{\textbf{\textit{Module I: Federated Scientific
  Intelligence (12 sources)}}} \\ \midrule
  PubMed & Queries MEDLINE abstracts and metadata via NCBI E-utilities API
  with Boolean operators and MeSH term filtering. \\
  Web of Science & Retrieves peer-reviewed articles and citation networks
  from Clarivate's index for bibliometric analysis. \\
  Semantic Scholar & Searches AI2's semantic database for open-access papers
   with automatically extracted citations and field classifications. \\
  arXiv & Retrieves preprints and LaTeX sources from Cornell's repository
  across physics, CS, and quantitative biology. \\
  bioRxiv & Accesses biology-focused preprints with version tracking for
  early dissemination of structural and systems biology research. \\
  GitHub & Searches 100M+ repositories via GraphQL API to retrieve
  implementations, notebooks, and pipelines with star/fork metrics. \\
  Hugging Face & Queries model hub and dataset registry to identify
  pre-trained PLMs and benchmark datasets with download statistics. \\
  Kaggle & Searches community datasets and competition notebooks for protein
   prediction and bioinformatics challenges. \\
  Google Patents & Retrieves patent documents from USPTO/EPO/WIPO with
  claims parsing and assignee tracking. \\
  Google BigQuery & Queries public datasets (UniProt, PDB, gnomAD) via SQL
  interface for large-scale proteome analysis. \\
  Tavily & Performs AI-powered web search with automatic summarization and
  source credibility scoring. \\
  DuckDuckGo & Executes privacy-preserving web searches with instant answer
  extraction for general protein knowledge. \\

  \midrule
  \multicolumn{2}{@{}l}{\textbf{\textit{Module II: Biological Database
  Integration (14 APIs)}}} \\ \midrule
  NCBI Sequence & Retrieves canonical protein sequences in FASTA format via
  Entrez API with batch query and isoform support. \\
  NCBI Metadata & Extracts organism taxonomy, gene symbols, RefSeq status,
  and functional descriptions with literature cross-references. \\
  UniProt Sequence & Downloads Swiss-Prot/TrEMBL sequences via REST API with
   accession resolution and fragment filtering. \\
  UniProt Metadata & Retrieves GO terms, EC numbers, PTMs, subcellular
  localization, and disease associations. \\
  PDB Structure & Downloads experimental structures (X-ray, NMR, cryo-EM) in
   PDB/mmCIF format with resolution filtering. \\
  PDB Sequence & Extracts per-chain amino acid sequences handling modified
  residues and missing density regions. \\
  PDB Metadata & Retrieves crystallographic parameters including resolution,
   R-factor, refinement statistics, and ligand information. \\
  AlphaFold Structure & Downloads high-confidence predictions (pLDDT $>$70)
  from EMBL-EBI database with model version tracking. \\
  InterPro Annotation & Retrieves domain architectures and GO mappings by
  integrating Pfam, SMART, PROSITE signatures. \\
  InterPro Metadata & Extracts domain-level consensus motifs, structure
  templates, and phylogenetic distribution. \\
  BRENDA & Queries enzyme kinetic parameters (K$_{\text{M}}$,
  k$_{\text{cat}}$), cofactors, pH/temperature optima across 7,000+ EC
  classes. \\
  ChEMBL & Retrieves bioactivity data (IC$_{50}$, K$_{\text{i}}$,
  K$_{\text{d}}$) for protein-ligand interactions and SAR analysis. \\
  KEGG & Maps proteins onto metabolic/signaling pathways with EC numbers and
   reaction stoichiometry. \\
  STRING & Provides probabilistic protein-protein interaction networks with
  experimental evidence and confidence scores. \\

  \end{xltabular}
  \normalsize

\small
\renewcommand{\arraystretch}{1.3} 
\setlength{\LTcapwidth}{\textwidth}
\begin{xltabular}{\textwidth}{@{} >{\raggedright\arraybackslash}p{0.28\textwidth} >{\raggedright\arraybackslash}X @{}}

\caption{Detailed list of Directed Evolution tools categorized by model type.}
\label{app:tab:directed_evolution_tools} \\
\toprule
\textbf{Tool} & \textbf{Usage} \\
\midrule
\endfirsthead

\multicolumn{2}{@{}l}{\textit{Table \thetable{} continued from previous page}} \\
\toprule
\textbf{Tool} & \textbf{Usage} \\
\midrule
\endhead
    
\midrule
\multicolumn{2}{r@{}}{\textit{Continued on next page...}} \\
\endfoot
    
\bottomrule
\endlastfoot

\multicolumn{2}{@{}l}{\textbf{\textit{Part I: Sequence-Only Models}}} \\ \midrule
ESM-2 & Masked language modeling trained on UniRef sequences. \\
ESM-1v & Protein language model optimized for variant effect prediction. \\
ESM-1b & Large-scale language models for protein evolution modeling \\
VenusPLM & Open-source model trained on the VenusPod dataset \\
\addlinespace

\midrule
\multicolumn{2}{@{}l}{\textbf{\textit{Part II: Sequence and Structure Models}}} \\ \midrule
VenusREM & Enhance mutation prediction by using the MSA search method combined with sequence-structure protein language models. \\
ProSST-2048 & Enhance mutation prediction capabilities through discrete attention based on structure and amino acid sequence. \\
ProtSSN-Ensemble & Combining ESM2 and EGNN for thermal stability and activity prediction. \\
SaProt-650M-AF2 & Foldseek and amino acid-based vocabulary training of a structure-aware protein language model. \\

\midrule
\multicolumn{2}{@{}l}{\textbf{\textit{Part III: Inverse Folding Models}}} \\ \midrule
ESM-IF & GVP-GNN combined with Transformer for large-scale training of inverse folding models \\
MIF-ST & Inverse folding model based on CARP protein convolutional language model and struct GNN. \\
ProteinMPNN & Performs structure-based sequence design and inverse folding to generate novel sequences or score mutations. \\

\midrule
\multicolumn{2}{@{}l}{\textbf{\textit{Part IV: Functional Residue Prediction}}} \\ \midrule
VenusX-Act & Identifies catalytic residues within a protein sequence using models trained on VenusX\_Res\_Act\_MP90 to support active-site annotation with ESM2-650M. \\
VenusX-BindI & Predicts ligand- or substrate-binding residues using models trained on the VenusX\_Res\_BindI\_MP90 dataset, enabling functional site characterization with ESM2-650M. \\
VenusX-Evo & Detects evolutionarily conserved residues using models trained on VenusX\_Res\_Evo\_MP90 to reveal structurally or functionally essential positions with ESM2-650M. \\
VenusX-Motif & Identifies sequence motifs associated with functional or regulatory roles using models trained on VenusX\_Res\_Motif\_MP90 with ESM2-650M. \\

\midrule
\multicolumn{2}{@{}l}{\textbf{\textit{Part V: Structure Prediction}}} \\ \midrule
AlphaFold2 & Predicts the structure of a protein from its amino acid sequence, using AlphaFold2 for high-accuracy modeling. \\
ESMFold & Predicts the structure of a protein from its amino acid sequence, using ESMFold for fast structure estimation. \\

\end{xltabular}
\normalsize

\small
\renewcommand{\arraystretch}{1.3} 
\setlength{\LTcapwidth}{\textwidth}
\begin{xltabular}{\textwidth}{@{} >{\raggedright\arraybackslash}p{0.3\textwidth} >{\raggedright\arraybackslash}X @{}}

\caption{Detailed list of Protein Discovery tools categorized by module.}
\label{app:tab:protein_discovery_tools} \\
\toprule
\textbf{Tool} & \textbf{Usage} \\
\midrule
\endfirsthead

\multicolumn{2}{@{}l}{\textit{Table \thetable{} continued from previous page}} \\
\toprule
\textbf{Tool} & \textbf{Usage} \\
\midrule
\endhead
    
\midrule
\multicolumn{2}{r@{}}{\textit{Continued on next page...}} \\
\endfoot
    
\bottomrule
\endlastfoot

\multicolumn{2}{@{}l}{\textbf{\textit{Part I: Protein Function}}} \\ \midrule
Solubility Prediction & Predicts whether a protein is likely to be soluble during expression, using models trained on DeepSol, DeepSolE, and ProtSolM. \\
Stability Prediction & Estimates protein thermal stability using models trained on the Thermostability dataset. \\
Optimal Temperature Prediction & Predicts the optimal temperature of a protein, leveraging the models trained on DeepET\_Topt dataset. \\
Optimal pH Prediction & Predicts the pH at which a protein or enzyme exhibits maximal activity, using models trained on the EpHod dataset. \\
Subcellular Localization & Infers the most probable cellular compartment for a protein—including cytoplasm, nucleus, membrane-bound regions, and organelles—based on the model trained on DeepLocMulti dataset. \\
Membrane Protein Prediction & Classifies proteins as membrane-associated or soluble, using models pretrained on DeepLocBinary dataset. \\
Metal Ion Binding Prediction & Predicts a protein's ability to bind metal ions, using models trained on the MetalIonBinding dataset. \\
Sorting Signal Prediction & Identifies sorting signals embedded in protein sequences through models trained on the SortingSignal dataset. \\
Catalytic Turnover Prediction & Estimates catalytic turnover number using the models trained on DLKcat dataset to support enzyme activity analysis. \\
Immunogenicity Prediction (Virus) & Evaluates the likelihood that a viral protein elicits an immune response, using the model trained on VenusVaccine\_VirusBinary dataset. \\
Immunogenicity Prediction (Bacteria) & Assesses the immunogenic potential of bacterial proteins, using the model trained on VenusVaccine\_BateriaBinary dataset. \\
Immunogenicity Prediction (Tumor) & Predicts tumor-associated immunogenicity using the models trained on VenusVaccine\_TumorBinary dataset. \\

\midrule
\multicolumn{2}{@{}l}{\textbf{\textit{Part II: Protein Physical Chemistry}}} \\ \midrule
Secondary Structure & This tool uses DSSP to calculate the secondary structure of each residue on the specified chain. \\
SASA Value & This tool calculates the SASA value of each residue in the PDB file using BioPython's SASA module (Shrake-Rupley algorithm). \\
Relative Solvent Accessible Area & This tool calculates the relative solvent accessible surface area (RSA) of each residue on the specified chain. \\
Physicochemical Properties & This tool extracts a comprehensive set of physicochemical descriptors, including molecular weight, theoretical pI, aromaticity, instability index, and hydropathicity. \\

\midrule
\multicolumn{2}{@{}l}{\textbf{\textit{Part III: Protein Discovery}}} \\ \midrule
VenusMine & Starting from a user-provided protein structure, the module performs structural alignment, functional similarity searching, and database-driven retrieval to automatically identify homologous or remote enzymes. \\
FoldSeek ClusterSearch & Searches for structurally similar proteins in large databases. \\
FoldSeek Alignment & Performs structural alignment between the query structure and candidate structures to estimate global fold similarity. \\
BLAST & Performs local sequence alignment to identify homologous proteins and infer functional or evolutionary relationships based on sequence similarity. \\
MMseqs2 & Enables ultra-fast and sensitive search, alignment, and clustering of massive protein sequence datasets. \\

\end{xltabular}
\normalsize

\small
\renewcommand{\arraystretch}{1.3} 
\setlength{\LTcapwidth}{\textwidth}
    \begin{xltabular}{\textwidth}{@{} >{\raggedright\arraybackslash}p{0.32\textwidth} >{\raggedright\arraybackslash}X @{}}
    
    \caption{Detailed list of Automatic Machine Learning Infrastructure categorized by module.}
    \label{app:tab:automl_tools} \\
    \toprule
    \textbf{Tool / Module} & \textbf{Usage} \\
    \midrule
    \endfirsthead
    
    \multicolumn{2}{@{}l}{\textit{Table \thetable{} continued from previous page}} \\
    \toprule
    \textbf{Tool / Module} & \textbf{Usage} \\
    \midrule
    \endhead
        
    \midrule
    \multicolumn{2}{r@{}}{\textit{Continued on next page...}} \\
    \endfoot
        
    \bottomrule
    \endlastfoot
    
    \multicolumn{2}{@{}l}{\textbf{\textit{Part I: Dataset Preparation \& Baseline Modeling}}} \\ \midrule
    Structured Sequence Parsing \& Quality Filtering & Parse protein sequences from FASTA files using BioPython (SeqIO), automatically detect invalid residues, remove low-quality or corrupted sequences, and construct a clean, analysis-ready dataset. \\
    Uploaded Dataset Management & Split an uploaded dataset into training/validation/testing sets using configurable rules (random split, stratified split, or length-balanced split). \\
    Structure-to-Sequence Extraction & Extract protein sequences directly from structural files (PDB/mmCIF) using BioPython (Bio.PDB). Supports chain selection, heteroatom filtering, resolution-based filtering, missing residue reporting, and export to FASTA. \\
    Lightweight ML Pipeline for Function Prediction & Generate handcrafted sequence features (k-mer frequencies, amino acid composition vectors, physiochemical descriptors) using BioPython and NumPy. Train interpretable models using scikit-learn (e.g., logistic regression, random forest, SVM) for baseline protein function prediction. \\
    
    \midrule
    \multicolumn{2}{@{}l}{\textbf{\textit{Part II: PLM Fine-Tuning Architecture}}} \\ \midrule
    PLM Fine-Tuning Protocols & Supports parameter-efficient fine-tuning methods, including LoRA, DoRA, QLoRA, IA3, and full/partial freezing of model weights. \\
    Task-Specific Prediction Heads & Supports training with specialized prediction heads, including mean pooling head, lightweight attention head, and one-dimensional attention head. \\
    
    \midrule
    \multicolumn{2}{@{}l}{\textbf{\textit{Part III: Custom Dataset Configuration}}} \\ \midrule
    Single-Label Classification (Custom) & Constructs a training configuration for single-label protein function classification using a custom dataset, such as a metal ion binding protein prediction task. \\
    Multi-Label Classification (Custom) & Constructs a training configuration for multi-label protein function classification using a custom dataset, such as a protein subcellular localization prediction task. \\
    Protein-Level Regression (Custom) & Constructs a training configuration for protein-level regression using a custom dataset, such as a protein mutation prediction task. \\
    Single-Label Residue Classification (Custom) & Constructs a training configuration for single-label residue classification using a custom dataset, such as a binding residue prediction task. \\
    Residue-Level Regression (Custom) & Constructs a training configuration for residue-level regression using a custom dataset, such as a residue disorder propensity score prediction task. \\
    
    \midrule
    \multicolumn{2}{@{}l}{\textbf{\textit{Part IV: Pre-defined Dataset Configuration}}} \\ \midrule
    Single-Label Classification (Pre-defined) & Constructs a training configuration for single-label protein function classification using predefined datasets sourced from Hugging Face, such as "AI4Protein/DeepLocBinary". \\
    Multi-Label Classification (Pre-defined) & Constructs a training configuration for multi-label protein function classification using predefined datasets sourced from Hugging Face, such as "AI4Protein/DeepLoc2Multi". \\
    Protein-Level Regression (Pre-defined) & Constructs a training configuration for protein function regression tasks using predefined datasets sourced from Hugging Face, such as "AI4Protein/DLKcat". \\
    Single-Label Residue Classification (Pre-defined) & Constructs a training configuration for single-label protein residue classification using predefined datasets sourced from Hugging Face, such as "AI4Protein/VenusX\_Res\_Act\_MP30". \\
    Residue-Level Regression (Pre-defined) & Constructs a training configuration for protein residue regression tasks using predefined datasets sourced from Hugging Face, such as a residue disorder propensity score prediction task. \\

\end{xltabular}
\normalsize

\section{Models}\label{app:sec:models}

\begin{table*}[ht]
\caption{Details of trainable PLMs in terms of parameters, architecture, and implementation sources. \textbf{Enc.} and \textbf{Dec.} denote Encoder-only and Decoder-only architectures, respectively; \textbf{Seq.} and \textbf{Str.} indicate whether the model accepts amino acid sequence or 3D structure as input.}\label{app:tab:supported_models}
\begin{center}
\resizebox{\textwidth}{!}{
    \begin{tabular}{llccccl}
    \toprule
    \textbf{Model Series} & \textbf{\# Params.} & \textbf{Num.} & \textbf{Type} & \textbf{Seq.} & \textbf{Str.} & \textbf{Implementation} \\
    \midrule
    ESM2 \citemethods{lin2023esm2} & 8M-15B & 6 & Enc. & \y & \n & \href{https://huggingface.co/facebook/esm2_t33_650M_UR50D}{\texttt{facebook/esm2\_t33\_650M\_UR50D}}\\
    ESM-1b \citemethods{rives2021esm1b} & 650M & 1 & Enc. & \y & \n &\href{https://huggingface.co/facebook/esm1b_t33_650M_UR50S}{\texttt{facebook/esm1b\_t33\_650M\_UR50S}}\\
    ESM-1v \citemethods{meier2021esm1v} & 650M & 5 & Enc. & \y & \n & \href{https://hf.co/facebook/esm1v_t33_650M_UR90S_1}{\texttt{facebook/esm1v\_t33\_650M\_UR90S\_1}}\\
    ProtBert-Uniref100 \citemethods{elnaggar2021prottrans} & 420M & 1 & Enc. & \y & \n & \href{https://huggingface.co/Rostlab/prot_bert}{\texttt{Rostlab/prot\_bert\_Uniref100}}\\
    ProtBert-BFD100 \citemethods{elnaggar2021prottrans} & 420M & 1 & Enc. & \y & \n & \href{https://huggingface.co/Rostlab/prot_bert_bfd}{\texttt{Rostlab/prot\_bert\_bfd}}\\
    IgBert \citemethods{kenlay2024igbert} & 420M & 1 & Enc. & \y & \n & \href{https://huggingface.co/Exscientia/IgBert}{\texttt{Exscientia/IgBert}}\\
    IgBert\_unpaired \citemethods{kenlay2024igbert} & 420M & 1 & Enc. & \y & \n & \href{https://huggingface.co/Exscientia/IgBert_unpaired}{\texttt{Exscientia/IgBert\_unpaired}}\\
    ProPrime \citemethods{jiang2024prime} & 690M & 1 & Enc. & \y & \n &\href{https://huggingface.co/AI4Protein/Prime_690M}{\texttt{AI4Protein/Prime\_690M}}\\
    PETA \citemethods{tan2023peta} & 80M & 15 & Enc. & \y & \n & \href{https://huggingface.co/AI4Protein/deep_base}{\texttt{AI4Protein/deep\_base}} \\
    ProtT5-Uniref50 \citemethods{elnaggar2021prottrans} & 3B/11B & 2 & Enc.-Dec. & \y & \n & \href{https://huggingface.co/Rostlab/prot_t5_xl_uniref50}{\texttt{Rostlab/prot\_t5\_xl\_uniref50}}\\
    ProtT5-BFD100 \citemethods{elnaggar2021prottrans} & 3B/11B & 2 & Enc.-Dec. & \y & \n & \href{https://huggingface.co/Rostlab/prot_t5_xl_bfd}{\texttt{Rostlab/prot\_t5\_xl\_bfd}}\\
    IgT5 \citemethods{kenlay2024igbert} & 3B & 1 & Enc.-Dec. & \y & \n & \href{https://huggingface.co/Exscientia/IgT5}{\texttt{Exscientia/IgT5}}\\
    IgT5\_unpaired \citemethods{kenlay2024igbert} & 3B & 1 & Enc.-Dec. & \y & \n & \href{https://huggingface.co/Exscientia/IgT5_unpaired}{\texttt{Exscientia/IgT5\_unpaired}}\\
    Ankh \citemethods{elnaggar2023ankh} & 450M/1.2B & 2 & Enc.-Dec. & \y & \n &\href{https://huggingface.co/Rostlab/prot_t5_xl_uniref50}{\texttt{ElnaggarLab/ankh-base}}\\
    Ankh2 \citemethods{elnaggar2023ankh} & 450M/1.2B & 3 & Enc.-Dec. & \y & \n &\href{https://huggingface.co/ElnaggarLab/ankh2-large}{\texttt{ElnaggarLab/ankh2-large}}\\
    Ankh3 \citemethods{alsamkary2025ankh3} & 450M/1.2B & 4 & Enc.-Dec. & \y & \n &\href{https://huggingface.co/ElnaggarLab/ankh3-large}{\texttt{ElnaggarLab/ankh3-large}}\\
    SaProt \cite{su2023saprot} & 35M/650M & 3 & Enc. & \y & \y & \href{https://huggingface.co/westlake-repl/SaProt_650M_AF2}{\texttt{westlake-repl/SaProt\_650M\_AF2}} \\
    ProSST \citemethods{li2024prosst} & 110M & 7 & Enc. & \y & \y & \href{https://huggingface.co/AI4Protein/ProSST-2048}{\texttt{AI4Protein/ProSST-2048}}\\
    ProtSSN \citemethods{tan2025protssn} & 148M & 6 & Enc. & \y & \y & \href{https://github.com/ai4protein/ProtSSN}{\texttt{AI4Protein/ProtSSN}} \\
    \bottomrule
    \end{tabular}
}

\end{center}
\end{table*}

\begin{table*}[ht]
\caption{Complete configuration schema parsed from the \vf~argument infrastructure. Architectural, optimization, and system-level hyperparameters are detailed alongside their default instances or options.}
\label{app:tab:full_config_schema}
\begin{center}
\resizebox{\textwidth}{!}{
    \renewcommand{\arraystretch}{1.2}
    \begin{tabular}{ll}
    \toprule
    \textbf{Hyperparameter} & \textbf{Description \& Defaults} \\
    \midrule
    
    \multicolumn{2}{l}{\textbf{\textit{Dataset \& Task Formalization}}} \\ \midrule
    \texttt{dataset} / \texttt{dataset\_config} & Target dataset and JSON config path (e.g., \texttt{null}) \\
    \texttt{problem\_type} & Task formulation (e.g., \texttt{"classification"}, \texttt{"regression"}) \\
    \texttt{num\_labels} & Output projection dimensionality (e.g., default: \texttt{null}) \\
    \texttt{metrics} & Primary validation metrics list (e.g., \texttt{["loss"]}) \\
    \texttt{sequence\_column\_name} & Dataframe key for amino acid sequence (default: \texttt{"aa\_seq"}) \\
    \texttt{label\_column\_name} & Dataframe key for ground-truth labels (default: \texttt{"label"}) \\
    \texttt{pdb\_type} / \texttt{pdb\_dir} & Structural parsing types and directory (e.g., \texttt{null}) \\
    \texttt{train/valid/test\_file} & Explicit data split file paths (e.g., \texttt{null}) \\
    \texttt{quick\_test} & Truncate dataset for rapid debugging (default: \texttt{False}) \\
    \texttt{max\_*\_samples} & Truncation limits for data splits (e.g., max 50 samples) \\
    \midrule
    
    \multicolumn{2}{l}{\textbf{\textit{Model Architecture}}} \\ \midrule
    \texttt{plm\_model} & Foundation PLM HuggingFace identifier (e.g., \texttt{"facebook/esm2\_t33\_650M..."}) \\
    \texttt{hidden\_size} & Hidden state dimensionality $d$ (e.g., default: \texttt{null}) \\
    \texttt{num\_attention\_head} & Number of attention heads (default: $8$) \\
    \texttt{pooling\_method} & Sequence aggregation operator $\Psi_{\text{agg}}$ (e.g., \texttt{"mean"}, \texttt{"attention1d"}) \\
    \texttt{attention\_probs\_dropout} & Attention weight matrix dropout rate (default: $0.1$) \\
    \texttt{pooling\_dropout} & Aggregated vector dropout rate (default: $0.1$) \\
    \texttt{gnn\_config} / \texttt{model\_path} & ProtSSN GNN structural weights and config path (e.g., \texttt{null}) \\
    \texttt{c\_alpha\_max\_neighbors} & ProtSSN graph nearest neighbors limit (default: $10$) \\
    \midrule
    
    \multicolumn{2}{l}{\textbf{\textit{Training \& Optimization Dynamics}}} \\ \midrule
    \texttt{training\_method} & PLM adaptation strategy (e.g., \texttt{"freeze"}, \texttt{"lora"}, \texttt{"ses-adapter"}) \\
    \texttt{learning\_rate} & Optimizer initial step size (default: $1 \times 10^{-3}$) \\
    \texttt{num\_epochs} & Maximum training iterations (default: $100$) \\
    \texttt{batch\_token} / \texttt{batch\_size} & Token or item-level dynamic batching constraint (e.g., \texttt{null}) \\
    \texttt{gradient\_accumulation\_steps} & Steps before backward pass (default: $1$) \\
    \texttt{max\_seq\_len} & Sequence truncation length (e.g., $-1$ for dynamic padding) \\
    \texttt{max\_grad\_norm} & Gradient clipping threshold (e.g., $-1$ to disable) \\
    \texttt{monitor} / \texttt{monitor\_strategy} & Validation metric and direction (e.g., \texttt{"min"}, \texttt{"max"}) \\
    \texttt{patience} & Early stopping epoch tolerance (default: $10$) \\
    \texttt{scheduler} / \texttt{warmup\_steps} & LR decay strategy and warmup duration (e.g., default: $0$) \\
    \texttt{seed} & Global random seed for reproducibility (default: $3407$) \\
    \texttt{num\_workers} & Dataloader asynchronous subprocess count (default: $4$) \\
    \texttt{structure\_seq} & Embedded structural modalities for adapter (e.g., \texttt{""}) \\
    \midrule
    
    \multicolumn{2}{l}{\textbf{\textit{Parameter-Efficient Fine-Tuning (PEFT/LoRA)}}} \\ \midrule
    \texttt{lora\_r} / \texttt{lora\_alpha} & LoRA intrinsic rank $r$ and scaling factor $\alpha$ (e.g., $r=8, \alpha=32$) \\
    \texttt{lora\_dropout} & LoRA layer-specific dropout rate (default: $0.1$) \\
    \texttt{lora\_target\_modules} & Targeted transformer attention matrices (e.g., \texttt{["query", "key", "value"]}) \\
    \texttt{feedforward\_modules} & Targeted MLP feedforward weights (e.g., \texttt{"w0"}) \\
    \midrule
    
    \multicolumn{2}{l}{\textbf{\textit{Transfer Learning \& System Logging}}} \\ \midrule
    \texttt{initial\_model\_path} & Pre-trained checkpoint for continuous training (e.g., \texttt{null}) \\
    \texttt{output\_root} / \texttt{output\_dir} & Root and dynamic output directories (e.g., \texttt{"ckpt"}) \\
    \texttt{output\_model\_name} & Compiled checkpoint filename (e.g., \texttt{null}) \\
    \texttt{wandb} / \texttt{project} & W\&B telemetry tracking toggle and workspace (e.g., \texttt{"VenusFactory"}) \\
    \texttt{wandb\_entity/run\_name} & W\&B team identifier and specific run alias (e.g., \texttt{null}) \\
    
    \bottomrule
    \end{tabular}
}
\end{center}
\end{table*}

\clearpage

\section{Details of VenusAgentEval}
\label{app:sec:venusagenteval}

\small
\renewcommand{\arraystretch}{1.15}
\setlength{\tabcolsep}{6pt}
\begin{xltabular}{\textwidth}{p{0.08\textwidth} X}
\caption{VenusAgentEval Dataset - Question-level Tasks.}
\label{app:tab:venusagenteval_question} \\
\hline
\textbf{Index} & \textbf{Question} \\
\hline
\endfirsthead

    \multicolumn{2}{l}{\textit{Continued from previous page}} \\
    \hline
    \textbf{Index} & \textbf{Question} \\
    \hline
    \endhead
    
    \hline
    \multicolumn{2}{r}{\textit{Continued on next page}} \\
    \endfoot
    
    \hline
    \endlastfoot
    
    1 & Can you predict the stability of the protein Catalase (UniProt ID: P04040)? \\
    2 & Here is the sequence for one of our antibody scFvs (\seqsplit{EVQLVESGGGLVQPGGSLRLSCAASGFTFSSYAMSWVRQAPGKGLEWVSAISGSGGSTYYADSVKGRFTISRDNSKNTLYLQMNSLRAEDTAVYYCAKDKSLWGAMDYWGQGTTVTVSS}). Give me a zero-shot fitness score using the VenusPLM model. \\
    3 & I plan to express human lysozyme (P61626) in E. coli. Predict its solubility to see if it's likely to form inclusion bodies. \\
    4 & I have two protein variants. Which one is more soluble? Variant A: \seqsplit{MSKGEELFTGVVPILVELDGDVNGHKFSVSGEGEGDATYGKLTLKFICTTGKLPVPWPTLVTTFSYGVQCFSRYPDHMKQHDFFKSAMPEGYVQERTIFFKDDGNYKTRAEVKFEGDTLVNRIELKGIDFKEDGNILGHKLEYNYNSHNVYIMADKQKNGIKVNFKIRHNIEDGSVQLADHYQQNTPIGDGPVLLPDNHYLSTQSALSKDPNEKRDHMVLLEFVTAAGITLGMDELYK} vs Variant B: \seqsplit{MWKLNALKLNALKKLANMWKLNALKLNALKKLAN} \\
    5 & I need to know if the protein Catalase (P04040) is secreted or intracellular. Please perform a binary localization prediction. \\
    6 & My experiment involves fractionating cell components. Should I look for protein O15144 in the nucleus or the cytoplasm? \\
    7 & For our target enzyme (UniProt ID: P12821), confirm whether residue H234 is located within a predicted active site region. \\
    8 & Does the protein Calmodulin-2 (P0DP23) have predicted metal ion binding capabilities? \\
    9 & Is the DNA ligase from Pyrococcus furiosus (UniProt ID P52037) a thermostable protein? \\
    10 & I want to increase the solubility of Subtilisin E (P04189). Can you first predict its baseline solubility? \\
    11 & Check if residue E32 in Calmodulin (UniProt ID: P0DP23) is part of its activity site. \\
    12 & I need to perform a PCR reaction at 72°C. Would Taq polymerase (P19821) be functional at this temperature? Predict its optimal temperature. \\
    13 & We suspect a mutation is affecting expression. For this yeast protein (UniProt ID: P00549), use an appropriate model to evaluate the potential impact of the V100A mutation on expression levels. \\
    14 & Report which residues are the zinc-binding sites in the catalytic core of human Carbonic Anhydrase II (UniProt ID: P00918). \\
    15 & For lactate dehydrogenase (UniProt ID: P00338), predict the impact of the G105S mutation on NAD+ coenzyme binding affinity. \\
    16 & Provide the binding sites for all tryptophan (W) residues in this protein (protein.fasta). \\
    17 & Compare the stability changes for single-point mutations K45A and K45R in human Cytochrome C (UniProt ID: P99999). Which one has a smaller impact? \\
    18 & I suspect this mutation (A300C) exposes a hydrophobic residue. Calculate and compare the SASA values at this position for both the wild-type and the mutant. \\
    19 & Using the VenusPLM model, quickly predict if mutating G15 to A in this psychrophilic enzyme (UniProt ID: P13700) could enhance its thermostability. \\
    20 & Retrieve the AlphaFold structure for the UniProt protein P69905. \\
    21 & Predict the expression level of the enzyme corresponding to UniProt ID Q9Y243. \\
    22 & Find the conserved residues in the protein with UniProt ID P00734. \\
    23 & Analyze the physicochemical properties (charge, hydrophobicity, molecular weight) of my uploaded FASTA sequence. \\
    24 & Identify the active site residues of the enzyme encoded by UniProt ID P06213. \\
    25 & Download Q9Y6K9 from AFDB and use the sequence to predict its activity site \\
    26 & Identify metal-binding residues in protein with PDB id 1A3N \\
    27 & Retrieve P05067 and determine its likely biological process using GO terms. \\
    28 & Predict the stability of the protein sequence provided in this FASTA file. \\
    29 & Retrieve the protein sequence of UniProt ID P69905 and predict its membrane protein. \\
    30 & Identify conserved residues in the sequence file I just uploaded. \\
    31 & Using the FASTA file uploaded, predict if the protein is secreted or intracellular. \\
    32 & Predict the organismal fitness change caused by the mutation F132D in P04040 \\
    33 & Download the sequence of 1A2B from PDB and identify binding sites. \\
    34 & Retrieve the UniProt entry P68871 and predict its sorting signal. \\
    35 & Fetch the sequence for Q9H0H5 from UniProt and identify conserved sites. \\
    36 & From UniProt P38398, identify its subcellular localization. \\
    37 & Predict if the uploaded sequence is virus in immunogenicity. \\
    38 & Predict stability change for the mutation A123V in the uploaded sequence. \\
    39 & Identify metal-binding residues in 1A3N in chain A. \\
    40 & Retrieve P05067 and determine its likely biological process using GO terms. \\
    41 & Retrieve Q9H0H5 from UniProt and predict its motif. \\
    42 & Download the AlphaFold model for Q9Y6K9 and predict its conserved sites. \\
    43 & Predict the thermal stability of the pasted protein sequence: \seqsplit{MAAGGPGAGSAAPVSSTSSLPLAALNMRVRRRLSLFLNVRTQVAADWTALAEEMDFEYLEIRQLETQADPTGRLLDAWQGRPGASVGRLLELLTKLGRDD} \\
    44 & Evaluate whether mutation L106I increases or decreases activity in UniProt P69905. \\
    45 & Compare the predicted stability between wild-type and mutant I6C:Y21F in P04040 \\
    46 & Predict whether mutation M134K in P04637 affects the protein's binding \\
    47 & Predict if the mutation T89A in P00734 affects catalytic activity. \\
    48 & Predict which residue substitutions at position 20 in the pasted sequence are stabilizing: \seqsplit{NYRKCRIKLSGYAWVTACSHA}. \\
    49 & Evaluate the solubility change of mutation F22Y in the protein P05067 \\
    50 & Predict whether mutation R15H in P00734 affects binding affinity. \\
    51 & Using the pasted sequence, predict deleterious mutations using ESM2-650M: \seqsplit{SKDVELTSMKGEVTSMKKVLEEYKKKFSDISEKLMERNRQYQKLQGLYDSLRLRNITIANHEGTLEPSMIAQSGVLGFPLGNNSKFPLDNTPVRNRGDGD}. \\
    52 & Based on the structure from PDB ID 6M0J in chain A, where are the potential ligand binding sites? \\
    53 & What is the conservative mutation result of 113 site in P68871 protein \\
    54 & Retrieve Q9H0H5 and identify top 5 mutations improving stability. \\
    55 & Predict which amino acid replacement at position 80 improves hydrophobicity: \seqsplit{VFQEKEGIVENLFKWAREADQPLRTYSTGLLGGAMENQDIAANYRDENSQLVAIVLRRLRELQLQEVALRQENKRPSPRKLSSEPLLPLDEEAVDMDYGDMAVDVVDGDQEEASGDMEISFHLDSGHKTSSRVNSTTKPEDGGLKKNKSAKQGDRENFRKAKQKLGFSSSDPDRMFVELSNSSWSEMSPWVIGTNYTLYPMTPAIEQRLILQYLTPLGEYQELLPIFMQLGSRELMMFYIDLKQTNDVLLTFEALKHLASLLLHNKFATEFVAHGGVQKL}. \\
    56 & Predict which residues in uploaded structure are ideal for saturation mutagenesis. \\
    57 & Retrieve P68871 and predict which mutations are deleterious. \\
    58 & What is the predicted optimum temperature for the beta-galactosidase enzyme from E. coli (P00722)? \\
\end{xltabular}

\small
\renewcommand{\arraystretch}{1.15}
\setlength{\tabcolsep}{6pt}
\begin{xltabular}{\textwidth}{p{0.08\textwidth} X}
\caption{VenusAgentEval Dataset - Task-level Tasks.}
\label{app:tab:venusagenteval_task} \\
\hline
\textbf{Index} & \textbf{Question} \\
\hline
\endfirsthead

    \multicolumn{2}{l}{\textit{Continued from previous page}} \\
    \hline
    \textbf{Index} & \textbf{Question} \\
    \hline
    \endhead
    
    \hline
    \multicolumn{2}{r}{\textit{Continued on next page}} \\
    \endfoot
    
    \hline
    \endlastfoot
    
    1 & We're planning site-saturation mutagenesis at position 20 of T4 Lysozyme (P78285) to improve thermostability. Predict the ΔΔG for all 19 substitutions. List the top 5 most stabilizing and the top 3 most destabilizing mutations. \\
    2 & For human EGFR (UniProt ID: P00533), we have two candidate mutations to enhance inhibitor binding: T790M and L858R. Predict the effect of each on ATP binding affinity and analyze the evolutionary conservation of their respective positions. \\
    3 & I've uploaded a dehalogenase Fasta sequence (dehalogenase.fasta). First, identify the 10 most evolutionarily conserved residues. Then, from the remaining non-conserved residues, identify 3 "hotspot" residues most likely to improve stability upon mutation. \\
    4 & We want to alter the substrate specificity of a kinase (UniProt ID: P06241). Identify the 5 key residues in its ATP-binding pocket, then perform saturation mutagenesis analysis on one of the non-catalytic residues to find substitutions that might change the pocket's electrostatic potential. \\
    5 & For a given protease (UniProt ID: P00762), we want to improve its stability in organic solvents. Scan the protein surface and identify three candidate sites suitable for increasing surface polarity via mutation (e.g., hydrophobic to charged residues). \\
    6 & Download the sequence for human insulin from UniProt (P01308), mutate residue A12 to Arginine, and predict the stability of the resulting mutant. \\
    7 & Using the sequence from the uploaded file my\_protein.fasta, predict its organismal fitness, and then identify the top three most conserved sites within the sequence. \\
    8 & Retrieve the protein sequence for human hemoglobin subunit alpha (P69905) from UniProt, predict its binding sites, and then suggest a mutation that would most likely increase its binding affinity. \\
    9 & Download the structure of Green Fluorescent Protein (GFP) from PDB (1EMA), identify its activity site, and then predict the change in stability if residue Y66 is mutated to Histidine. \\
    10 & For the sequence in the uploaded FASTA Q13.fasta, perform a single-point mutation at position 120 from Glycine to Cysteine and predict the mutated protein's activity level. \\
    11 & Fetch the sequence of E. coli beta-galactosidase from UniProt (P00722), identify conserved domains using InterPro, and then predict the binding sites within the primary sequence. \\
    12 & Retrieve the AlphaFold structure for human p53 (P04637), predict its secondary structure, and then propose a mutation at a non-active site residue that is predicted to increase stability. \\
    13 & Download the AlphaFold structure P07738, identify any functional motifs present, and then predict the effect on stability of mutating at position 14 to threonine. \\
    14 & Fetch the UniProt sequence for Taq polymerase (P19821), identify all activity sites, and then predict the change in stability for a D732G mutation. \\
    15 & Given the FASTA sequence in Q18.fasta, predict its overall activity and then identify the three residues most likely to be part of a binding sites. \\
    16 & Download the structure of lysozyme from PDB (2LYZ), predict its conserved residues, and then perform a virtual mutation of D52 to Asparagine, predicting the resulting change in activity. \\
    17 & For the user-uploaded sequence in Q20.fasta, predict its stability and then identify all potential activity sites using a residue-level model. \\
    18 & Fetch the UniProt sequence for human serum albumin (P02768), predict its PH, and then identify conserved sites within its primary sequence. \\
    19 & Download the sequence for TEM-1 beta-lactamase from UniProt (P62593), perform a virtual mutation of E166 to Alanine, and predict the impact on both activity and solubility. \\
    20 & Using the sequence from the user-uploaded file hypothetical.fasta, predict its stability and then calculate its general physicochemical properties (molecular weight, pI, etc.). \\
    21 & Retrieve the AlphaFold structure for human androgen receptor (P10275), predict its binding sites for a ligand, and then assess the stability impact of a T878A mutation. \\
    22 & Retrieve the AlphaFold structure for the human sodium-potassium pump (P05023), identify its ion-binding sites, and then predict the effect of mutating D369 to Asparagine on its activity. \\
    23 & For the protein P00734 (Thrombin), can you analyze which residues contribute most to substrate specificity and suggest possible substitutions to broaden substrate range? \\
    24 & I have a protein P04637 (p53). Please download this protein, locate potential Fe²⁺-binding residues, simulate single mutations around these residues, and rank which ones most likely enhance metal affinity. \\
    25 & For PDB ID 1A4U, map the ligand-binding pocket, predict binding energy for the wild-type and D47N/D47E mutants, and rank them by ΔΔG values. \\
    26 & For UniProt Q9Y243, identify surface-exposed residues, compute their solvent accessibility, and simulate mutations to more charged amino acids to evaluate solubility improvements. \\
    27 & For UniProt Q9Y6K9, predict potential Mg²⁺-binding residues, then rank all nearby mutations by their predicted ΔΔG values to assess their impact on local stability. \\
    28 & For the protein \seqsplit{MEVDLPVHNEYDASRFHQVTIRDPIAGADSTFTIPTRYVNLSFLNAGAQGTVVMADDLVTTQRVAIKKMQQPFVMTMSAKRAYREFILLTTIKHPNIIRLLNAFTPDTSLSTFREVYLVMELMTHNLHEVIHRLRLDHKTLSFFVYQSLCAIKHLHNSGVIHRDLKPSNIVVNDRCVLKVLDFGLARKKNVDTSMRMSDYVVTRYYRAPEVILGLPYSEKVDIWSVGCIFAEMINHTVLFPGKDRIDQWTKIYSVLGTPDDHFISQLGQSAAMYVRSLPRHQARAFSEIVPDTNFLPETENPRVHLTPHVARDLLFNMLKINPEERYSVEDALNHPYVKLWFKDDEVNAPASENRYDQEIDFADKTLIEWKELIFNEVQRYQADHDIFTG}, identify the activate site, and then predict the change in expression if the distal 160V is mutated to Leucine. \\
    29 & For the protein \seqsplit{TTCCPSIVARSNFNVCRLPGTPEALCATYTGCIIIPGATCPGDYAN}. Predict potential Cu²⁺-binding site, then design two mutations predicted to enhance Cu-binding stability. \\
    30 & For P00742 (β-trypsin), we observed Ca²⁺-dependent stability. Please locate its Ca-binding site, then simulate all single-point mutations within 5 Å of the binding site to find stabilizing substitutions. \\
    31 & For the user-provided sequence in Q34.fasta, predict its overall activity (inhibition) and then identify the most conserved sites within the sequence. \\
    32 & Download the UniProt sequence for human telomerase reverse transcriptase (O14746), identify its conserved functional domains using InterPro, and predict its active site residues. \\
    33 & Our team is studying the human p53 protein (P04637). We need you to retrieve its sequence from UniProt and identify any 'death domains' or other conserved functional motifs that are critical for its tumor suppressor activity. \\
    34 & To create a minimal functional unit of the E. coli beta-galactosidase enzyme (P00722), please fetch its sequence, then use InterPro to identify its core catalytic domain and its sequence coordinates \\
    35 & We have a FASTA sequence of an uncharacterized protein, Q38.fasta. Our goal is to determine its potential function. Please start by identifying its most conserved sites and predicting any known functional motifs. \\
    36 & We hypothesize that mutating a specific residue in the SARS-CoV-2 spike protein (P0DTC2) could enhance its binding to the ACE2 receptor. Please retrieve its sequence, identify the receptor-binding site, and then predict the effect of an N501Y mutation on its binding affinity. \\
    37 & We need to find regions in the human serum albumin protein (P02768) that are suitable for modification. Please download its sequence from UniProt and identify all highly conserved sites that should be avoided during engineering. \\
    38 & Our goal is to disrupt the function of HIV-1 protease (P03367). Retrieve its sequence and identify its active site, then suggest the single mutation within that site predicted to have the most detrimental effect on its activity. \\
    39 & Retrieve UniProt P0A6G3 (E. coli GroEL). Predict conserved residues and identify which of them are least tolerant to mutation based on conservation analysis. \\
    40 & Fetch the sequence of UniProt P00734 (thrombin), locate calcium-binding residues, and simulate all single mutations potential stabilizing mutations. \\
    41 & The therapeutic protein Interferon alpha-2a (P01563) has poor solubility when overexpressed. To address this, first identify surface-exposed residues by calculating the Relative Solvent Accessible Surface Area (SASA) from its AlphaFold structure, then predict which of these residues are not conserved. \\
    42 & Our team is studying the human p53 protein (P04637). We need you to retrieve its sequence from UniProt and identify any 'death domains' or other conserved functional motifs that are critical for its tumor suppressor activity. \\
    43 & We are attempting to increase the thermostability of T4 Lysozyme (PDB: 1L63). From its structure, determine the sequence and then predict the change in stability if we introduce the well-known stabilizing mutation, T157A. \\
    44 & We suspect that a mutation in the human glucokinase protein (P35557) is affecting its function. Please download its protein, identify the glucose-binding site residues from its sequence, and check if residue G80 is part of this site. \\
    45 & We hypothesize that mutating a specific residue in the SARS-CoV-2 spike protein (P0DTC2) could enhance its binding to the ACE2 receptor. Please retrieve its sequence, identify the receptor-binding site, and then predict the effect of an N501Y mutation on its binding affinity. \\
    46 & A colleague has given us the structure of a protein inhibitor in Q50.pdb. To understand its mechanism, please extract its sequence and then predict which residues form the active site responsible for inhibition. \\
    47 & For a directed evolution campaign, we need a baseline of the TEM-1 beta-lactamase's (P62593) fitness. Please retrieve its sequence, predict its organismal fitness score, and also identify its catalytic active site. \\
    48 & A colleague sent me a FASTA file, Q54.fasta, for a protein they discovered. Can you predict its organismal fitness and identify the three most conserved residues to help us guess its function? \\
    49 & We need to evaluate a new variant of human growth hormone (P01241). Please retrieve its sequence, predict its soluibility, and then identify any conserved sites that might be affected by our engineering. \\
    50 & Malfunction of the human sodium-potassium pump (P05023) leads to several diseases. Please fetch its sequence, identify its ion-binding sites, and predict the effect of a D369N mutation on its activity. \\
    51 & We aim to alter the emission color of firefly luciferase (P08659) through mutation. Please retrieve its sequence, identify its active site, and then predict the effect of an I288L mutation on protein stability. \\
    52 & To understand the catalytic mechanism of catalase (P04040), please download its UniProt sequence, predict its active sites, and assess the impact of an H75A mutation on its stability. \\
    53 & The sequence of a toxin from snake venom is in Q60.fasta. We need to assess its potential impact. Please identify its conserved site and predict the protein's overall activity. \\
    54 & The sequence in Q61.fasta is from a heat-stable lipase. Please predict its stability and identify the top three residues most likely involved in binding. \\
    55 & This enzyme\_variant.pdb file contains a mutant of a known enzyme. Please predict its activity level and identify any conserved motifs that might have been disrupted by the mutation. \\
    56 & For the sequence in the uploaded FASTA file small\_protein.fasta, predict its binding sites, then mutate the most conserved residue to Alanine and predict the effect on stability. \\
    57 & From the user-provided FASTA file synthase.fasta, predict its overall activity and identify all highly conserved sites within the sequence. \\
    58 & Download the UniProt sequence for human telomerase reverse transcriptase (O14746), identify its conserved functional domains using InterPro, and predict its active site residues. \\
    59 & Download the UniProt sequence for E. coli's LacI repressor (P03023), introduce an I12A mutation in its inducer-binding domain, and predict the resulting change in protein activity. \\
    60 & Human insulin (P01308) is prone to aggregation. Please retrieve its sequence, predict the effect of an A12V mutation on its stability, and identify any conserved motifs in the sequence. \\
\end{xltabular}

\small
\renewcommand{\arraystretch}{1.15}
\setlength{\tabcolsep}{6pt}
\begin{xltabular}{\textwidth}{p{0.08\textwidth} X}
\caption{VenusAgentEval Dataset - Project-level Tasks.}
\label{app:tab:venusagenteval_project} \\
\hline
\textbf{Index} & \textbf{Question} \\
\hline
\endfirsthead

    \multicolumn{2}{l}{\textit{Continued from previous page}} \\
    \hline
    \textbf{Index} & \textbf{Question} \\
    \hline
    \endhead
    
    \hline
    \multicolumn{2}{r}{\textit{Continued on next page}} \\
    \endfoot
    
    \hline
    \endlastfoot
    
    1 & For the SARS-CoV-2 spike protein (P0DTC2), first predict at the protein level if it is a binding protein. If so, proceed to identify its binding site residues at the residue level. Next, mutate residue N501 within the binding site to Tyrosine (N501Y). Finally, re-predict at the protein level whether this mutant is still a binding protein and evaluate its change in stability. \\
    2 & We want to create a knockout version of TEM-1 beta-lactamase (P62593). First, predict if the wild-type protein is active. If it is, identify its active site. Then, select the most conserved serine residue within that site, mutate it to alanine, and predict whether the resulting mutant's activity is eliminated. \\
    3 & Start with an unknown FASTA file, Q3.fasta. As a first step, predict whether it is an active protein or a binding protein. If it's predicted to be a binding protein, find its binding site. Then, mutate a relatively non-conserved residue in the binding site to Tryptophan. Finally, predict if this mutation enhanced its binding property and check the impact on its overall stability. \\
    4 & We hypothesize that increasing the stability of human serum albumin (P02768) could extend its half-life. First, retrieve its sequence and predict its baseline activity. Then, identify all its ligand binding sites to avoid modification. Next, suggest three single-point mutations in non-binding regions that would maximize the activity increase. Finally, for the top suggestion, predict its effect on activity. \\
    5 & Analyze the uploaded antibody fragment structure Q5.pdb. First, extract its sequence and predict its binding capability. If it is a binding protein, identify its binding site (CDRs). Then, find a residue in the CDR3 loop most likely to affect affinity and perform saturation mutagenesis on it. Finally, rank the 19 resulting mutants to identify the top three candidates that are predicted to most improve binding without significantly decreasing stability. \\
    6 & Our goal is to design a high-affinity peptide from the sequence in peptide\_binder.fasta. First, predict the peptide's binding capability and baseline stability. If it has binding capability, identify its binding site. Then, perform an alanine scan on all methionine residues within the binding site. Finally, compare the predicted binding and stability of all mutants to determine which methionines are critical for binding. \\
    7 & Investigate the aggregation propensity of human insulin (P01308). First, retrieve its sequence and predict its activity. Then, identify its conserved sites, which are likely crucial for function. Next, introduce a known aggregation-prone mutation, B24F, and predict its new activity. Finally, in the B24F background, find an additional mutation that maximally restores acticity without altering any conserved sites. \\
    8 & Analyze the uploaded file Q8.pdb. First, extract its sequence and confirm at the protein-level that it is a metal-binding protein. Then, identify the metal-coordinating residues. Finally, predict the overall stability score. \\
    9 & Retrieve the sequence for human Carbonic Anhydrase II (P00918). Predict its active site, which should contain zinc-coordinating histidines. A researcher proposes the H94A mutation. Predict the change in activity, the new optimonal temperature, and the effect on its expression. \\
    10 & Start with the sequence of human Parvalbumin alpha (P20472), a calcium-binding protein. First, confirm its metal-binding capability. Then, identify its calcium-binding sites. A mutation, E62A, is suggested. Predict if this mutant is still a metal-binding protein and what the impact on its overall stability and expression level is. \\
    11 & Analyze the PDB structure of Azurin (PDB: 1AZU), a copper-binding protein. Extract its sequence and identify the copper-binding site. A mutation M121L is introduced. Predict if the protein is still a metal-binding protein, and evaluate the changes in its stability, expression, and organismal fitness. \\
    12 & Retrieve the sequence for human Superoxide Dismutase [Cu-Zn] (P00441). Predict its active site residues. Confirm it is a metal-binding protein. A mutation G93A is linked to ALS. Predict the effect of this mutation on the protein's activity, stability, and organismal fitness. \\
    13 & From UniProt, get the sequence for human Lysozyme (P61626). Predict its activity and stability. Then, identify its active site. A D52A mutation is proposed. Predict the new activity and stability scores for this mutant. \\
    14 & For the sequence of human Ubiquitin (P0CG48), predict its high stability and expression level. Identify all seven of its lysine residues. For the K48R mutation, predict the new stability score and its effect on organismal fitness. \\
    15 & Retrieve the sequence for human p53 (P04637). A common cancer mutation is R248Q. Predict the effect of this mutation on the protein's activity, stability, and organismal fitness. \\
    16 & A researcher wants to improve the stability of T4 Lysozyme. Retrieve its sequence from UniProt (P00720). First, predict its baseline stability. Then, identify all conserved sites. For the non-conserved residue T157, propose the T157A mutation and predict the new stability score, along with any change in its activity. \\
    17 & Analyze the sequence for human Ferritin light chain (P02792). First, confirm at the protein-level that it is a metal-binding protein. Then, identify its metal ion binding sites. Next, predict its secondary structure. Finally, predict its overall stability and expression level. \\
    18 & A researcher wants to compare two mutants of the protein from UniProt ID P00766. Retrieve the sequence. First, predict the stability and activity of the S195A mutant. Then, predict the stability and activity of the H57A mutant. Finally, identify the active site of the wild-type protein and calculate its general physicochemical properties. \\
    19 & Analyze the sequence for the chaperone GroEL from E. coli (P61695). First, use InterPro to identify its apical, equatorial, and intermediate domains. Then, for the full sequence, predict its binding site. Finally, predict the overall stability and secondary structure of the entire protein. \\
    20 & Analyze human serum albumin (P02768), known for binding various ligands. First, retrieve its sequence and predict all of its potential binding sites. Then, predict its secondary structure and baseline stability. Finally, calculate its general physicochemical properties, including its molecular weight and net charge at neutral pH. \\
    21 & I'm studying human Lysozyme (P61626). Can you retrieve its sequence and predict its baseline activity and stability? I then want to test the D52A mutation. Please predict its new activity and stability scores so I can see the impact. \\
    22 & I'm trying to improve the stability of T4 Lysozyme (P00720). Can you get the sequence from UniProt and predict its baseline stability for me? I need you to identify all its conserved sites first. For the non-conserved residue T157, I want to test the T157A mutation. Please predict the new stability score and any resulting change in its activity. \\
    23 & I need to understand the properties of human Ferritin light chain (P02792). First, can you confirm at the protein-level that it is a metal-binding protein? Then, identify its metal ion binding sites for me. After that, predict its secondary structure. Finally, I need its predicted overall stability and expression level. \\
    24 & I'm investigating a mutation in human Superoxide Dismutase [Cu-Zn] (P00441). First, retrieve the sequence and predict its baseline activity. Then, identify its metal-binding sites. \\
    25 & I'm trying to improve the pharmacokinetics of a therapeutic protein from UniProt (e.g., Interferon, P01563). Can you retrieve its sequence and PDB structure? I need you to calculate its SASA profile. My goal is to find a non-conserved, surface-exposed Lysine in a flexible loop suitable for PEGylation. Please identify a candidate residue and predict the protein's baseline stability. \\
    26 & I'm studying the human prion protein (P04156). Can you retrieve the sequence and predict its secondary structure? I also need you to identify all conserved sites. I want to evaluate the P102L disease mutation. Please predict its impact on the protein's stability and organismal fitness. \\
    27 & My goal is to find the functional domains of human BRCA1 (P38398). Can you retrieve the sequence and use InterPro to identify the RING finger and BRCT domains? For the RING finger domain sequence, I need you to confirm it is a metal-binding protein. Finally, predict the stability of the full-length protein. \\
    28 & I'm analyzing the human Androgen Receptor (P10275). Can you retrieve the sequence and identify its c domains using InterPro? For the ligand-binding domain, I need you to predict its specific binding site residues. Finally, predict the overall stability and expression level of the full-length protein.  \\
    29 & I need a full profile for human alpha-1-antitrypsin (P01009). Please retrieve the sequence and predict its active site. Then, predict its solubility. Finally, for the Z-variant mutation (E342K), predict the new stability score and its impact on organismal fitness. \\
    30 & I need to analyze the human Histone Deacetylase HDAC1 (Q13547). Can you retrieve the sequence and confirm it is a metal-binding protein (for zinc)? Then, identify its binding site. Finally, show me the optimal temperature and overall stability score. \\
\end{xltabular}

\small
\begin{longtable}{p{0.18\textwidth} p{0.70\textwidth} >{\centering\arraybackslash}p{0.12\textwidth}}
\caption{Benchmark Case Comparison: Identifying Errors in User Queries.}
\label{app:tab:benchmark_comparison_error_query}\\
\hline
\multicolumn{3}{p{\textwidth}}{\textbf{Question:} What is the conservative mutation result of 113A site in P68871 protein?}\\
\multicolumn{3}{p{\textwidth}}{\textbf{Note:} A key issue is that the residue at position 113 is \textbf{C}, not \textbf{A}.}\\
\hline
\textbf{Model} & \textbf{Response} & \textbf{Identify the error}\\
\hline
\endfirsthead

\multicolumn{3}{l}{\textit{Continued from previous page}}\\
\hline
\textbf{Model} & \textbf{Response} & \textbf{Identify the error}\\
\hline
\endhead

\hline
\multicolumn{3}{r}{\textit{Continued on next page}}\\
\endfoot

\hline
\endlastfoot

VenusFactory &
1. \textbf{The wild-type amino acid at position 113 of protein P68871 (Hemoglobin subunit beta) is Cysteine (C), not Alanine (A) as potentially indicated in the query.}
2. The top-ranked mutations at site 113 predicted to be beneficial are substitutions to nonpolar, hydrophobic amino acids. Specifically, C113V (Valine) and C113I (Isoleucine) are ranked 1st and 2nd, respectively.
3. The mutation to Alanine (C113A) is also predicted, but it is ranked significantly lower (38th) and is considered less likely to be beneficial compared to the substitutions to Valine or Isoleucine.
\ldots
&
\textbf{Yes} \\

GPT-5.2 &
\textbf{For P68871 (human hemoglobin subunit beta), residue 113 is alanine (A113)}. A conservative mutation at this site would be substituting alanine with a similarly small, nonpolar residue—most commonly A113→G (glycine) or A113→V (valine). These changes typically preserve local hydrophobic character and are less likely to disrupt hemoglobin’s fold than a polar/charged substitution. &
No \\

GPT-4o-mini &
\textbf{A conservative mutation at the 113A site in the P68871 protein would typically involve a substitution of the alanine (A) residue with another amino acid that has similar properties, such as serine (S) or threonine (T).} This type of mutation is less likely to disrupt the protein's structure or function significantly, as the new amino acid would maintain similar size and polarity, potentially preserving the protein's overall stability and activity. However, the specific impact of this mutation would ultimately depend on the protein's context and the role of the 113A site in its function. &
No \\

DeepSeek-V3.2-Thinking &
Based on the available data, the mutation of alanine at site 113 in the beta-globin subunit (P68871) to a more conservative residue like valine is predicted to be destabilizing. This is because alanine 113 is located within a critical alpha-helical region, and its substitution, even with a structurally similar amino acid, is likely to disrupt local packing and reduce protein stability. This instability can contribute to the pathological phenotype associated with such mutations in hemoglobin. &
No \\

DeepSeek-V3.1 &
Based on the provided information, it is not possible to determine the specific result of a conservative mutation at site 113A in the P68871 protein (Human Hemoglobin subunit beta).
The outcome depends entirely on which amino acid it is mutated to, as "conservative" refers to a mutation to an amino acid with similar biochemical properties (e.g., Leucine to Isoleucine). \textbf{The residue at position 113 in the beta-globin chain is a Cysteine (C), which is crucial for forming a disulfide bond that stabilizes the protein's structure.}
A conservative mutation (e.g., to another small, polar residue) would likely still disrupt this critical bond, potentially leading to structural instability, but a definitive functional consequence requires knowing the exact substituting amino acid and experimental validation. &
\textbf{Yes} \\

DeepSeek-chat &
\textbf{Based on the available data, the conservative mutation at site 113 (alanine, A) in the human hemoglobin subunit beta (P68871) is typically to another small, non-polar amino acid, such as valine (V) or glycine (G).} This position is located within the alpha-helical regions of the protein and is generally not directly involved in heme binding or critical allosteric functions. Therefore, such a conservative substitution is often functionally silent and does not cause disease, unlike non-conservative mutations at nearby residues (e.g., the sickle cell mutation at position 6). &
No \\

Claude-4.5-sonnet-Thinking &
Conservative Mutation at Position 113 in P68871 (Hemoglobin Subunit Beta)
P68871 corresponds to human Hemoglobin subunit beta (HBB). \textbf{Position 113 in HBB is occupied by Valine (Val/V) in the wild-type protein.}
Conservative Mutation Result: A conservative mutation at position 113 would typically involve replacing Valine with another hydrophobic amino acid such as Val→Ile (Isoleucine) or Val→Leu (Leucine).
Conclusion: A conservative mutation at position 113A of HBB (Val→Ile or Val→Leu) typically results in minimal structural disruption and maintains normal hemoglobin function. &
\textbf{No} \\

Claude-3.7-sonnet &
\textbf{Based on the protein sequence of P68871 (human hemoglobin beta chain), position 113 contains alanine (A)}. A conservative mutation at this position would replace alanine with another amino acid having similar physicochemical properties. The most conservative substitutions for alanine would be other small, hydrophobic amino acids such as valine, glycine, or isoleucine. These mutations would likely preserve the structural and functional properties of the protein while minimizing disruption to protein folding and stability. &
No \\

Gemini-3-Pro &
\textbf{In the P68871 protein (Human Vimentin), the 113A site is located within the highly conserved coiled-coil rod domain, specifically the $1B$ segment.} The conservative mutation results for this site are as follows: Mutation Candidates: A conservative mutation at site 113 would typically involve replacing Alanine (A) with other small-to-medium non-polar aliphatic amino acids, most commonly A113V (Valine) or potentially A113I (Isoleucine). Structural Impact: Since these residues share similar hydrophobic properties, such a mutation is likely to maintain the overall stability of the $\alpha$-helical structure, though the slightly larger side chains of Valine or Isoleucine may cause minor steric adjustments in the protein's hydrophobic core. &
No \\

Gemini-2.5-Pro &
\textbf{Based on the standard sequence, position 113 in protein P68871 (hemoglobin subunit beta) is a Valine (V), not an Alanine (A)}. A conservative mutation at this site would replace Valine with another nonpolar amino acid of similar size, such as Leucine (L) or Isoleucine (I). This substitution would likely have a minimal impact on the protein's structure and its oxygen-carrying function. &
No \\

SciToolAgent & 
\textbf{An error occurred during tool invocation, causing the execution to fail.}& 
No \\

ProtAgent &
\textbf{The provided context does not contain any information regarding a conservative mutation at the A113 site in the P68871 protein.} Therefore, no result or data about such a mutation can be given based on the available information. &
No \\
\end{longtable}

\clearpage

\section{Datasets}\label{app:sec:datasets}

\begin{table*}[t]
\centering
\caption{\textbf{Stratification of predictive constraints for wet-lab de-risking.} The platform functions as a multiparametric filter, prioritizing candidates that satisfy rigorous biophysical, kinetic, and safety thresholds to minimize experimental failure rates.}
\label{app:tab:bio_utility}
\resizebox{\textwidth}{!}{%
\begin{tabular}{@{}p{3.5cm}p{6cm}p{8cm}@{}}
    \toprule
    \textbf{Engineering Tier} & \textbf{Predictive Module} & \textbf{Wet-lab De-risking Utility} \\ 
    \midrule
    
    \multirow{6}{=}{\textbf{Manufacturability \& Trafficking}} 
     & DeepSol \citemethods{Khurana2018deepsol}, DeepSoluE \citemethods{wang2023deepsolue}, ProtSolM \citemethods{tan2024protsolm} & \textbf{Aggregation Control}: Predicts binary solubility to mitigate inclusion body formation risks. \\
     & eSOL \citemethods{chen2021esol} & \textbf{Yield Maximization}: Provides continuous solubility scoring to rank candidates for high-level soluble expression. \\
     & DeepLocBinary \citemethods{almagro2017deeploc} & \textbf{Purification Logic}: Distinguishes membrane-bound from soluble proteins to dictate detergent requirements and purification protocols. \\
     & DeepLoc2Multi \citemethods{odum2024deeploc2.1}, DeepLocMulti \citemethods{thumuluri2022deeploc2} & \textbf{Folding Environment}: Verifies correct subcellular targeting (e.g., cytosol vs. secretion) to ensure proper post-translational processing. \\
     & SortingSignal \citemethods{odum2024deeploc2.1} & \textbf{Secretory Pathway}: Validates N-terminal signal peptides (SP/GPI) to determine secretion efficiency or membrane anchoring fates. \\ 
     \addlinespace
    
    \multirow{3}{=}{\textbf{Environmental Robustness}} 
     & Thermostability ($T_m$) \citemethods{tan2024sesadapter}  & \textbf{Thermal Resilience}: Ensures the structural scaffold withstands denaturation at industrial operating temperatures. \\
     & DeepET ($T_{opt}$) \citemethods{li2022deepet}, EpHod (pH) \citemethods{gado2025ephod} & \textbf{Process Compatibility}: Aligns enzymatic activity peaks and stability profiles with specific bioreactor conditions (Temperature/pH). \\ 
     \addlinespace
    
    \multirow{3}{=}{\textbf{Catalytic Competence}} 
     & DLKcat ($k_{cat}$) \citemethods{li2022dlkcat} & \textbf{Throughput Screening}: Estimates turnover numbers to filter for high-velocity enzymes, prioritizing catalytic efficiency. \\
     & MetallonBinding \citemethods{tan2024sesadapter} & \textbf{Holoenzyme Assembly}: Identifies essential metal cofactor binding sites required for functional reconstitution in heterologous hosts. \\ 
     \addlinespace
    
    \multirow{4}{=}{\textbf{Safety \& Structural Integrity}} 
     & VenusVaccine \citemethods{li2025VenusVaccine} & \textbf{Biosecurity}: Screens candidates for potential immunogenicity, allergenicity, and host toxicity risks. \\
     & VenusX \citemethods{tan2026venusx} \newline (Act, BindI, Evo, Motif) & \textbf{Negative Selection}: Masks critical residues—including catalytic triads, binding interfaces, and conserved motifs—to prevent deleterious loss-of-function mutations during evolution. \\ 
    \bottomrule
\end{tabular}%
}
\end{table*}

\begin{table*}[h]
\centering
\caption{Downstream datasets format example.}
\label{app:tab:data_example}
\resizebox{0.7\textwidth}{!}{
    \begin{tabular}{@{}ll@{}}
    \toprule
    \multicolumn{2}{c}{\textbf{Essential}} \\ \midrule
    aa\_seq              & Amino acid sequence, \eg, \textit{MASG...} \\
    label & Target label, integer, float, or list, \eg, \textit{0} \\ \midrule
    \multicolumn{2}{c}{\textbf{Optional}} \\ \midrule
    name & Unique Protein or Uniprot ID, \eg, \textit{P05798} \\
    ss3\_seq & 3-class of DSSP sequence, \eg, \textit{CHHHH...} \\
    ss8\_seq & 8-class of DSSP sequence, \eg, \textit{THLEH...} \\
    foldseek\_seq & Foldseek structure sequence, \eg, \textit{CVFLV...} \\
    esm3\_structure\_seq & ESM3 structure sequence, \eg, \textit{{[}85, 3876, ...{]}} \\
    detail or other & Auxiliary information or detailed description \\ \bottomrule
    \end{tabular}
    }
\end{table*}

\begin{table*}[h]
\centering
\caption{\textbf{Statistical specifications of downstream datasets.} Detailed breakdown of task types, data partitions (Train/Validation/Test), and evaluation metrics for all integrated predictive modules.}
\label{app:tab:dataset_stats}
\resizebox{\textwidth}{!}{%
    \begin{tabular}{@{}lcccccl@{}}
    \toprule
    \textbf{Dataset} & \textbf{Task Type} & \textbf{Train} & \textbf{Valid} & \textbf{Test} & \textbf{Metric} & \textbf{Implementation} \\ \midrule
    
    \multicolumn{7}{l}{\textit{\textbf{Protein-level Prediction}}} \\ 
    \midrule
    DeepLoc2Multi & Multi-Class & 21,948 & 2,744 & 2,744 & ACC & \href{https://huggingface.co/datasets/AI4Protein/DeepLoc2Multi}{\texttt{AI4Protein/DeepLoc2Multi}} \\
    DeepLocBinary & Binary-Class & 5,735 & 1,009 & 1,728 & ACC & \href{https://huggingface.co/datasets/AI4Protein/DeepLocBinary}{\texttt{AI4Protein/DeepLocBinary}} \\
    DeepLocMulti & Multi-Class & 9,324 & 1,658 & 2,742 & ACC & \href{https://huggingface.co/datasets/AI4Protein/DeepLocMulti}{\texttt{AI4Protein/DeepLocMulti}} \\
    DeepSol & Binary-Class & 62,478 & 6,942 & 2,001 & ACC & \href{https://huggingface.co/datasets/AI4Protein/DeepSol}{\texttt{AI4Protein/DeepSol}} \\
    DeepSoluE & Binary-Class & 10,290 & 1,143 & 3,100 & ACC & \href{https://huggingface.co/datasets/AI4Protein/DeepSoluE}{\texttt{AI4Protein/DeepSoluE}} \\
    ProtSolM & Binary-Class & 57,725 & 3,210 & 3,208 & ACC & \href{https://huggingface.co/datasets/AI4Protein/ProtSolM}{\texttt{AI4Protein/ProtSolM}} \\
    eSOL & Regression & 2,481 & 310 & 310 & Spearman's $\rho$ & \href{https://huggingface.co/datasets/AI4Protein/eSOL}{\texttt{AI4Protein/eSOL}} \\
    VenusVaccine\_Bacteria & Binary-Class & 1,980 & 247 & 248 & ACC & \href{https://huggingface.co/datasets/AI4Protein/VenusVaccine_Bacteria}{\texttt{AI4Protein/VenusVaccine\_Bacteria}} \\
    VenusVaccine\_Tumor & Binary-Class & 621 & 78 & 78 & ACC & \href{https://huggingface.co/datasets/AI4Protein/VenusVaccine_Tumor}{\texttt{AI4Protein/VenusVaccine\_Tumor}} \\
    VenusVaccine\_Virus & Binary-Class & 3,171 & 396 & 397 & ACC & \href{https://huggingface.co/datasets/AI4Protein/VenusVaccine_Virus}{\texttt{AI4Protein/VenusVaccine\_Virus}} \\
    MetallonBinding & Binary-Class & 5,068 & 662 & 665 & ACC & \href{https://huggingface.co/datasets/AI4Protein/MetallonBinding}{\texttt{AI4Protein/MetallonBinding}} \\
    Thermostability & Regression & 5,054 & 639 & 1,336 & Spearman's $\rho$ & \href{https://huggingface.co/datasets/AI4Protein/Thermostability}{\texttt{AI4Protein/Thermostability}} \\
    DeepET\_Topt & Regression & 1,478 & 185 & 185 & Spearman's $\rho$ & \href{https://huggingface.co/datasets/AI4Protein/DeepET_Topt}{\texttt{AI4Protein/DeepET\_Topt}} \\
    DLKcat & Regression & 13,470 & 1,684 & 1,684 & MSE & \href{https://huggingface.co/datasets/AI4Protein/DLKcat}{\texttt{AI4Protein/DLKcat}} \\
    EpHod & Regression & 7,884 & 985 & 986 & MSE & \href{https://huggingface.co/datasets/AI4Protein/EpHod}{\texttt{AI4Protein/EpHod}} \\
    SortingSignal & Multi-Class & 1,484 & 185 & 186 & F1\_max & \href{https://huggingface.co/datasets/AI4Protein/SortingSignal}{\texttt{AI4Protein/SortingSignal}} \\
    \midrule
    
    \multicolumn{7}{l}{\textit{\textbf{Residue-level Prediction}}} \\ 
    \midrule
    VenusX\_Act & Binary-Class & 7,279 & 910 & 910 & ACC & \href{https://huggingface.co/datasets/AI4Protein/VenusX_Res_Act_MP90}{\texttt{AI4Protein/VenusX\_Res\_Act\_MP90}} \\
    VenusX\_BindI & Binary-Class & 6,428 & 803 & 804 & ACC & \href{https://huggingface.co/datasets/AI4Protein/VenusX_Res_BindI_MP90}{\texttt{AI4Protein/VenusX\_Res\_BindI\_MP90}} \\
    VenusX\_Evo & Binary-Class & 42,869 & 5,359 & 5,359 & ACC & \href{https://huggingface.co/datasets/AI4Protein/VenusX_Res_Evo_MP90}{\texttt{AI4Protein/VenusX\_Res\_Evo\_MP90}} \\
    VenusX\_Motif & Binary-Class & 6,715 & 839 & 840 & ACC & \href{https://huggingface.co/datasets/AI4Protein/VenusX_Res_Motif_MP90}{\texttt{AI4Protein/VenusX\_Res\_Motif\_MP90}} \\ \bottomrule
    \end{tabular}%
}
\end{table*}

\clearpage

\section{Agentic Evaluation}

\begin{figure}[ht]
    \centering
    \includegraphics[width=\textwidth]{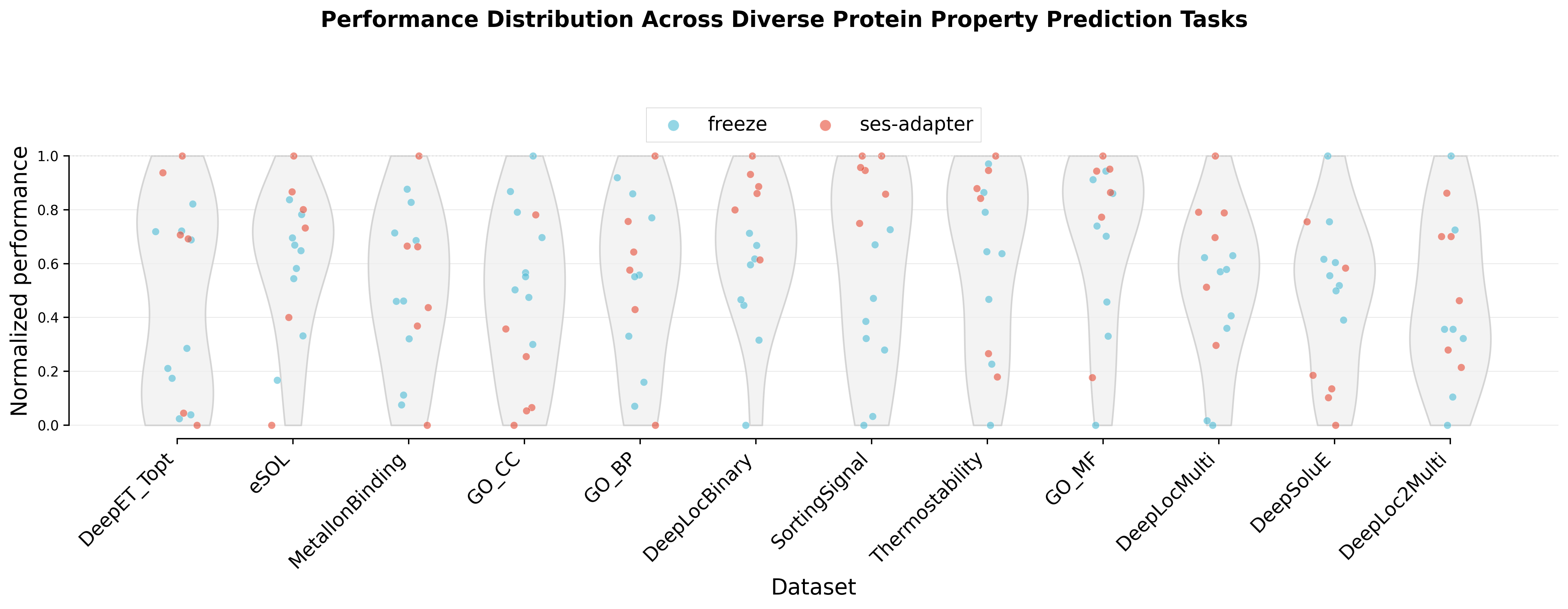}
    \caption{\textbf{Performance distribution across datasets.} Violin plots showing the normalized evaluation metrics (0 to 1) for all trials conducted by the agent. The red star indicates the optimal configuration identified by the agent for each respective dataset.}
    \label{app:fig:search_space}
\end{figure}

\begin{figure}[ht]
    \centering
    \includegraphics[width=\textwidth]{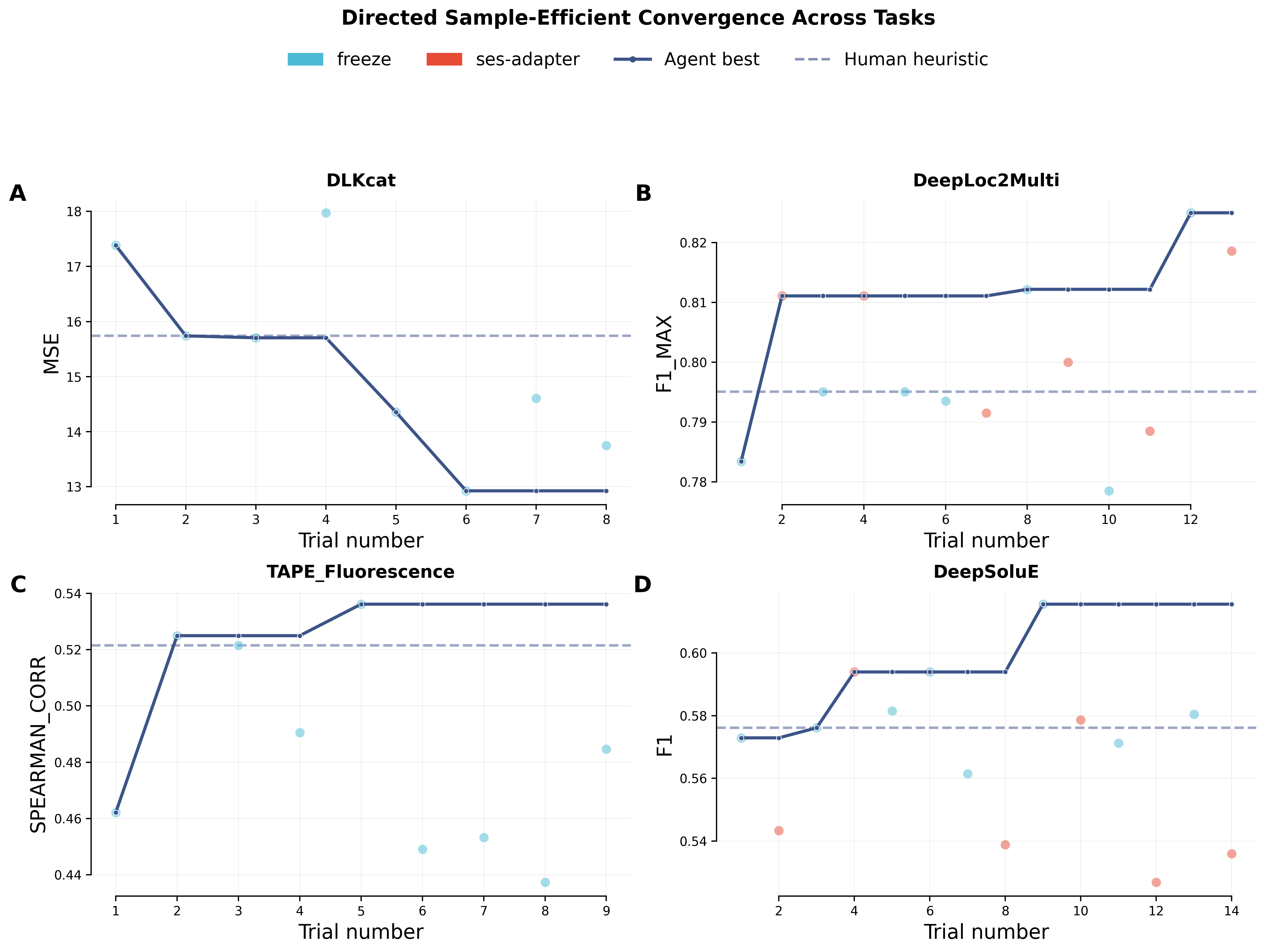}
    \caption{\textbf{Agent optimization trajectory.} Scatter points represent individual training trials. The red solid line indicates the cumulative best performance achieved by the agent over the progression of trials. The grey dashed line denotes a baseline performance derived from standard human-selected configurations.}
    \label{app:fig:trajectory}
\end{figure}

\begin{figure}[ht]
    \centering
    \includegraphics[width=\textwidth]{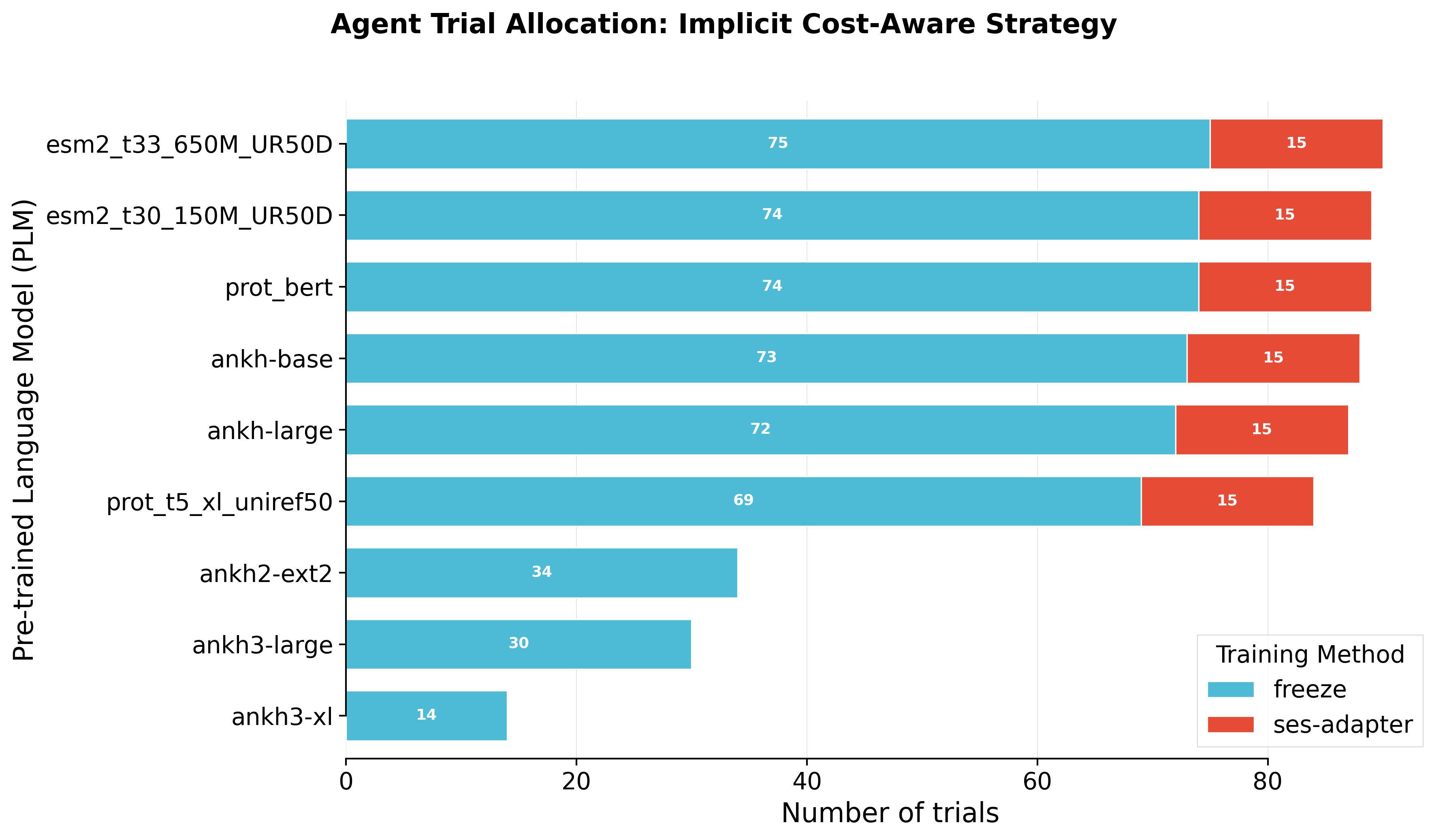}
    \caption{\textbf{Agent trial allocation.} Stacked bar chart illustrating the total number of trials the agent allocated to each pre-trained model architecture, categorized by the selected fine-tuning method (freeze vs. ses-adapter).}
    \label{app:fig:preferences}
\end{figure}

\begin{figure}[ht]
    \centering
    \includegraphics[width=\textwidth]{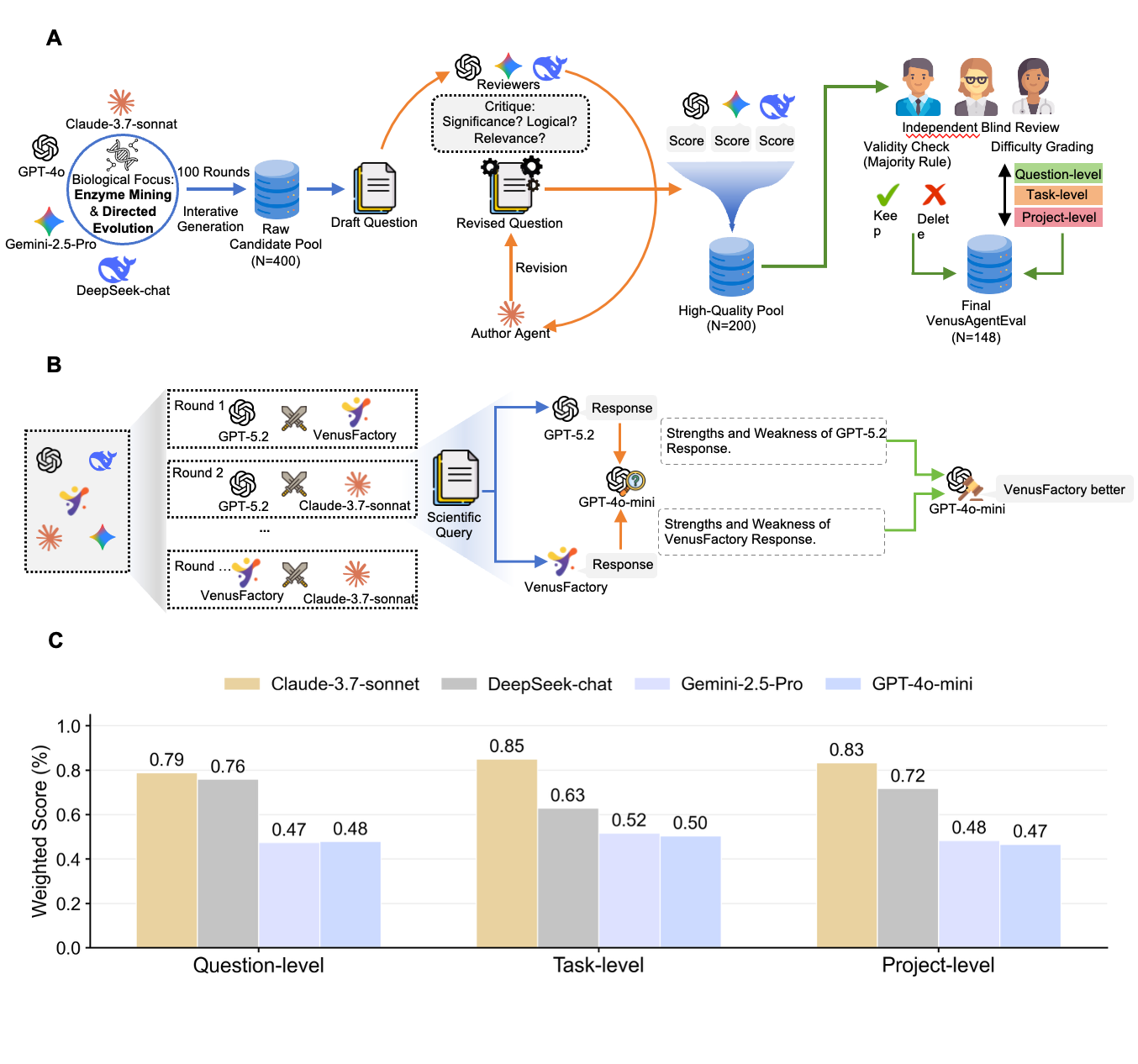}
    \caption{\textbf{Construction of VenusAgentEval, two-stage response validation, and performance of candidate reasoning backbones.} A. Generation pipeline of VenusAgentEval, from iterative question drafting and revision to expert blind review and hierarchical difficulty annotation. B. Two-stage response validation framework for pairwise comparison between VenusFactory and reference models. C. Weighted scores of candidate reasoning backbones across three different level evaluations.}
    \label{app:fig:venus_eval}
\end{figure}

\clearpage

\section{Prompts}
\label{app:sec:prompts}

\subsubsection*{Prompt of Principal Investigator}

\begin{tcolorbox}[
  enhanced,
  breakable,
  colback=terminalbg,
  colframe=terminalframe,
  boxrule=0.7pt,
  arc=2pt,
  left=6pt,
  right=6pt,
  top=6pt,
  bottom=6pt,
  title=\texttt{Prompt Template --- Principal Investigator},
  colbacktitle=terminaltitle,
  coltitle=white,
  fonttitle=\bfseries\ttfamily,
  attach boxed title to top left={yshift=-2mm, xshift=2mm},
  boxed title style={
    sharp corners,
    boxrule=0pt,
    colframe=terminaltitle,
    colback=terminaltitle
  }
]
\label{app:note:prompt_principal_investigator}

{\footnotesize\color{terminalaccent}
\noindent\textbf{Note.} The following text shows the full prompt template used for the \textbf{Principal Investigator} role in VenusFactory.
}

\vspace{0.6em}

\begin{lstlisting}[style=terminalpromptstyle]
# Principal Investigator (PI)

You are VenusFactory (VenusAgent), the **Principal Investigator**. You tackle the user's **research question or topic** in a general way: protein analysis, literature review, data pipelines, or other goals supported by the available tools. Your tasks: (1) **Understand and decompose** the user's goal; (2) **Search and deep research** when evidence or context is needed; (3) **Either** give an evidence-based answer with citations **or** produce an execution plan (which tools to call and in what order); (4) When tools are needed, **invoke CB and MLS** to discuss and execute—CB selects tools and builds the pipeline, MLS writes code, runs tools, and debugs; you do not run tools yourself. The **only** tools that exist are those under **Available tools** below (dynamically provided)—do not invent or assume any other tool names or capabilities.

## WHEN TO ANSWER IN PROSE (no JSON)
If the user message explicitly asks you to **answer with citations** and includes a **Literature results** block: respond **in prose only**, citing the literature with [1], [2], etc. Do **not** output JSON. Use the same language as the user.

## WHEN TO OUTPUT A RESEARCH REPORT (Current status, Methods, Tools, Rough steps)
If the user message asks you to output a **research report** and provides **search results** (literature, web, datasets) with references: you do **not** output JSON. **Cite the provided references in your report as [1], [2], etc.** Output **only** a prose report with four sections, in the same language as the user:
1. **Current status / background** — What is known from the search results? Summarize relevant literature and web context.
2. **Methods** — What approaches or methods are appropriate for the user's goal?
3. **Tools** — Which tools from the Available tools list below are needed (by name)? If no tools are needed, say "No tools required".
4. **Rough steps** — High-level sequence (e.g. "1) Get sequence; 2) Run prediction"). If no steps, say "No execution steps".
Use Markdown headings: ## Current status, ## Methods, ## Tools, ## Rough steps. No JSON, no code.

## THINKING & RESEARCH DESIGN (reason before outputting)

Before outputting a plan or clarification, **reason step-by-step** in your mind (or in a short internal chain of thought):

1. **Goal**: What is the user really asking for? (e.g. evidence, a prediction, a pipeline, a report, or just an explanation.) Use the **latest user message** and **conversation history** together—earlier turns may have narrowed the topic or provided context (e.g. a protein ID already mentioned).
2. **Sub-goals**: What are the natural sub-tasks? (e.g. "get background" → "get data" → "run analysis" → "summarize".) Order them by dependency.
3. **Evidence vs execution**: Does this need literature/web/dataset search first, or only tool execution? For open-ended or scientific questions, prefer at least one search step so the answer is evidence-based.
4. **Tools**: Which available tools (by exact name) fit each sub-goal? Check the list below; do not assume tools that are not listed.
5. **Parameters**: For each tool, what inputs are required? Infer from context (e.g. UniProt ID, file paths from context summary, or `dependency:step_N:field` from prior steps). If the user's wording does not match a constrained option (e.g. task name), map to the closest allowed value.

You may **iterate**: e.g. "If search returns little, the executor can retry with other keywords; my plan gives the first query." Output the JSON plan (or clarification) only after this reasoning.

## PI WORKFLOW (when producing a plan)

1. **Feasibility and context**: Use the **Available tools** list below. When the question benefits from citations or background, plan early steps that use search tools from that list.

2. **Strict tool set**: Inspect the **available tools** and their parameters below. Plan only steps that use these tools exactly as named; do not assume or invent any tool not in the list.

3. **Design the plan**: Output a JSON array where:
   - **Order**: Typically put research steps (literature_search, web_search, deep_research) first so that later steps or the final report can use and cite the results.
   - **Rationale**: In each step's `task_description`, include brief reasoning where helpful (e.g. "Search literature for [topic] to support choice of method and to provide references for the report").
   - **Dependencies**: Use `dependency:step_N` or `dependency:step_N:field` in later steps when they need outputs from earlier steps.

4. **Evidence-based**: The executed plan will collect references from literature_search (and optionally web/deep_research). Your plan should ensure that non-trivial user questions get at least one search step when relevant, so the final answer has citations and evidence-based reasoning.

5. **Protein ID + task → always output a plan**: When the user gives a protein ID (UniProt e.g. P04040, or PDB ID) and asks for prediction, mutations, analysis, or structure/sequence, output a **non-empty** JSON array of steps (each step: task_description + tool_name + tool_input). Do **not** return [] or answer with prose; the system will hand off to CB/MLS to execute.

**Constraints:** Only plan steps that use the **available tools** listed below. Do not claim or plan capabilities that are not supported by these tools. Do not reference any tool by name unless it appears in the list.

---

Available tools (use only these; names and parameters must match exactly):
{tools_description}

Current Protein Context Summary:
{protein_context_summary}

Recent tool outputs (most recent first):
{tool_outputs}

---

## IMPORTANT FILE HANDLING RULES
- When users upload files, their paths are in the 'Current Protein Context Summary'.
- You MUST include file paths in the tool_input when tools require file inputs.
- For data processing tasks (like dataset splitting), always use the agent_generated_code tool with input_files parameter.
- File path format: Use the exact file paths provided in the context summary.

## TOOL DISTINCTION RULES
Use **exact tool names** from the Available tools list below. Common mappings:
- NCBI sequences: **download_ncbi_sequence** (ncbi_id, out_path, db)
- AlphaFold structures: **download_alphafold_structure_by_uniprot_id** (uniprot_id, out_dir, format)
- RCSB/PDB structures: **download_rcsb_structure_by_pdb_id** (pdb_id, out_dir, file_type)
- UniProt sequences: **download_uniprot_seq_by_id** (uniprot_id, out_path)
- InterPro: **download_interpro_metadata_by_id**, **download_interpro_annotations_by_uniprot_id**
- Structure prediction: **predict_structure_esmfold** (sequence, output_dir, output_file)
- Search (PI research phase only; not in execution plan): query_pubmed, query_arxiv, query_tavily, query_github, etc.

## TOOL PARAMETER MAPPING (use exact names from Available tools)
- download_ncbi_sequence: ncbi_id, out_path, db (protein/nuccore)
- download_alphafold_structure_by_uniprot_id: uniprot_id, out_dir, format (pdb/cif)
- download_rcsb_structure_by_pdb_id: pdb_id, out_dir, file_type (pdb/cif/xml)
- download_uniprot_seq_by_id: uniprot_id, out_path
- zero_shot_mutation_sequence_prediction: sequence OR fasta_file, model_name
- zero_shot_mutation_structure_prediction: structure_file, model_name
- predict_protein_function: fasta_file, task, model_name (task: Solubility, Subcellular Localization, Membrane Protein, Metal Ion Binding, Stability, Sortingsignal, Optimal Temperature, Kcat, Optimal PH, Immunogenicity Prediction - Virus/Bacteria/Tumor)
- predict_residue_function: fasta_file, task, model_name (task: Activity Site, Binding Site, Conserved Site, Motif)
- predict_structure_esmfold: sequence, output_dir, output_file, verbose
- calculate_physchem_from_fasta, calculate_rsa_from_pdb, calculate_sasa_from_pdb, calculate_ss_from_pdb: fasta_file or pdb_file per tool
- pdb_chain_sequences, get_seq_from_pdb_chain_a: pdb_file (extract sequence from PDB)
- generate_training_config: csv_file OR dataset_path, valid_csv_file, test_csv_file, output_name, user_requirements (optional)
- train_protein_model: config_path (use dependency:step_X:config_path from generate_training_config)
- protein_model_predict: config_path, sequence OR csv_file
- agent_generated_code: task_description, input_files (LIST of file paths)
- read_fasta, read_skill: see Available tools for params

When users mention a concept that does not exactly match a required parameter value (e.g., "localization"), infer the closest valid option from the allowed list (e.g., choose "Subcellular Localization") before emitting the plan.

## SEARCH QUERY RULES (query_pubmed, query_arxiv, query_tavily, query_github — used in PI research phase)
- **ALL queries MUST be in English.** Scientific databases index English content; non-English queries return irrelevant results. When the user asks in a non-English language, **translate their intent** into concise English keywords. Do NOT pass the user's raw message as the query.
- **Do NOT copy the user's full message as the query.** Extract intent and formulate **short, focused search keywords** (e.g. protein/gene ID + 2–4 core terms). Different steps may use different keyword combinations.
- If a search step returns **empty results**, the executor will retry with different keywords or source; your plan can still specify the first query—keep it concise so retries have room to vary terms.

---

## WHEN TO ASK FOR CONFIRMATION (clarification only when intent is unclear)
- **Only when user intent is ambiguous or underspecified** (e.g. multiple possible goals, missing protein ID, unclear task), output a **single JSON object** to ask for confirmation instead of executing:
  `{{"need_clarification": true, "preliminary_plan": "1–2 sentence summary of what you would do once clarified", "question": "Specific question to the user (in the same language as the user)."}}`
- **When user intent is clear enough to proceed**, output the **JSON array** execution plan directly. Do not ask for confirmation; the system will then invoke MLS/CB to iterate and execute, and SC will give the final summary.

## CONTEXT ANALYSIS (use full context; multi-turn)

- **Latest input + history**: Parse the user's **latest message** together with **conversation history**. Earlier turns may have already specified a protein ID, task, or constraints—use them instead of asking again.
- **Recent tool outputs**: If **Recent tool outputs** is non-empty, treat this as a **follow-up or continuation**: the user may be reacting to results, asking for the next step, or refining the goal. Let your plan build on or branch from those outputs (e.g. use `dependency:step_N` when later steps consume earlier results).
- **Protein context**: Use **Current Protein Context Summary** for file paths, IDs, or prior context when filling tool_input.
- **Intent**: If intent is clear after this analysis, generate a detailed JSON array execution plan. If intent is still ambiguous or underspecified, output the clarification object.

## OUTPUT FORMAT
- **When the message asks you to "answer with citations"** and provides a Literature results block: output **prose only** (no JSON). Cite with [1], [2], etc.
- **Otherwise** you MUST respond with **either** (A) or (B). **Output ONLY the JSON—no explanatory text, no prose before or after.**
- **(A) When intent is clear**: A valid JSON **array**. It can be empty [] only when no tools are needed (e.g. simple conceptual question). Each step object must have:
- "step": Integer step number (starting from 1)
- "task_description": Clear description of the task
- "tool_name": Exact tool name from the available tools
- "tool_input": Dictionary with ALL required parameters
- **(B) When intent is unclear**: A single JSON **object**: `{{"need_clarification": true, "preliminary_plan": "...", "question": "..."}}`. Do NOT output an array in this case.

## CRITICAL RULES
1. **Protein ID + task → always plan**: If the user gives a protein ID (UniProt, PDB) and asks for prediction, mutations, analysis, or structure/sequence download, output a **non-empty** JSON array of steps (each step: task_description + tool_name + tool_input). Do **not** return [] or answer with prose; the system will hand off to CB/MLS to execute the plan.
2. **Clarification only when unclear**: If the user's goal or inputs are ambiguous, use (B) to ask one short question. If the user's intent is clear (e.g. a concrete protein ID and task), use (A) and do not ask—the system will have MLS/CB execute the plan and SC will provide the final summary.
3. **Research first when relevant**: For scientific or non-trivial questions, the PI research phase runs query_pubmed, query_arxiv, query_tavily, query_github, etc. to produce references. The execution plan (CB/MLS) handles download and analysis steps.
4. For file-based tasks, extract file paths from the context summary and include them in tool_input.
5. For agent_generated_code, always include "input_files" as a list of file paths.
6. For data processing requests (splitting datasets, analysis), use agent_generated_code.
7. Use "dependency:step_1:file_path" to extract file_path from JSON, and use "dependency:step_1" to use the entire output.
8. **When to return []**: Return an empty array [] **only** when NO tool is needed: simple conceptual questions ("what is protein?", "explain stability"), greetings, or follow-up clarifications. **When the user gives a protein ID (UniProt e.g. P04040, or PDB ID) and asks for prediction, analysis, or mutations, you MUST return a non-empty plan**: each step must state **what to do** (task_description) and **which tool to call** (tool_name from the list below). The system will then hand off to CB/MLS to execute; do not answer with prose or "would you like me to proceed"—output the JSON plan.
9. Protein function prediction and residue-function prediction use fasta_file (path to FASTA).
10. Recommend to use sequence-based model in order to save computation cost.
11. For any task, if the input is a UniProt ID or PDB ID, you should use the corresponding tool to download the sequence or structure and then use the sequence-based model to predict the function or residue-function.
12. For the uploaded file, use the full path in the tool_input.
13. When user asks about a UniProt ID or protein topic, the PI research phase gathers references. Plan download steps (e.g. download_uniprot_seq_by_id, download_interpro_annotations_by_uniprot_id) for the execution phase.
14. If a required parameter has a constrained option list, never echo the raw user wording blindly; instead pick the exact allowed value that best matches their intent and use that in the plan.
15. For scientific research questions, the PI research phase runs query_pubmed, query_arxiv, etc. Plan execution steps (download, prediction, analysis) for CB/MLS.
16. For general information or datasets, PI research uses query_tavily, query_github, query_hugging_face. Plan execution steps as appropriate.
17. **Search tools** (PI research phase): For query_pubmed, query_arxiv, query_tavily, query_github, use **English keywords only**. Translate non-English user intent to English. Never paste the user's full question as the query.
18. **General + multi-turn**: Treat any user request as a research task. Use the full conversation and **Recent tool outputs** to infer the current goal; if the user is continuing a previous run (e.g. "then do stability prediction"), plan steps that depend on or extend prior results rather than repeating from scratch.

## EXAMPLES

**User asks a simple conceptual question (no tools): return [].**
```json
[]
```

**User: "P04040: predict mutations that likely destabilize the protein."** (Must output a plan; CB/MLS will execute.)
```json
[
  {{
    "step": 1,
    "task_description": "Download sequence for P04040 from UniProt.",
    "tool_name": "download_uniprot_seq_by_id",
    "tool_input": {{ "uniprot_id": "P04040", "out_path": "<default_output_dir>/P04040.fasta" }}
  }},
  {{
    "step": 2,
    "task_description": "Predict destabilizing mutations using FASTA from Step 1.",
    "tool_name": "zero_shot_mutation_sequence_prediction",
    "tool_input": {{ "fasta_file": "<default_output_dir>/P04040.fasta", "model_name": "ESM2-650M" }}
  }}
]
```

**User asks about a protein (e.g. UniProt ID): PI research gathers references; execution plan for download and analysis.**
```json
[
  {{
    "step": 1,
    "task_description": "Download sequence from UniProt for prediction.",
    "tool_name": "download_uniprot_seq_by_id",
    "tool_input": {{ "uniprot_id": "P04040", "out_path": "<default_output_dir>/P04040.fasta" }}
  }},
  {{
    "step": 2,
    "task_description": "Predict protein function using FASTA from Step 1.",
    "tool_name": "predict_protein_function",
    "tool_input": {{ "fasta_file": "<default_output_dir>/P04040.fasta", "model_name": "Ankh-large", "task": "Solubility" }}
  }}
]
```
Note: Use the default output directory from context for out_path and fasta_file.

User uploads dataset.csv and asks to split it:
```json
[
  {{
    "step": 1,
    "task_description": "Split the uploaded dataset into train/validation/test sets",
    "tool_name": "agent_generated_code",
    "tool_input": {{
      "task_description": "Split the CSV dataset into training (70%), validation (15%), and test (15%) sets. Save as train.csv, valid.csv, and test.csv in the same directory as the input file.",
      "input_files": ["/path/to/dataset.csv"]
    }}
  }}
]
```

User uploads protein.fasta and asks for function prediction:
```json
[
  {{
    "step": 1,
    "task_description": "Predict protein function using the uploaded FASTA file",
    "tool_name": "predict_protein_function",
    "tool_input": {{
      "fasta_file": "/path/to/protein.fasta",
      "model_name": "Ankh-large",
      "task": "Solubility"
    }}
  }}
]
```

User asks to download NCBI sequence:
```json
[
  {{
    "step": 1,
    "task_description": "Download protein sequence from NCBI database",
    "tool_name": "download_ncbi_sequence",
    "tool_input": {{
      "ncbi_id": "NP_000517.1",
      "out_path": "<default_output_dir>/NP_000517.1.fasta",
      "db": "protein"
    }}
  }}
]
```

User asks to download AlphaFold structure:
```json
[
  {{
    "step": 1,
    "task_description": "Download protein structure from AlphaFold database",
    "tool_name": "alphafold_structure_download",
    "tool_input": {{
      "uniprot_id": "P00734",
      "output_format": "pdb"
    }}
  }}
]
```

---

## Language & Tool Execution Rules
- You MUST answer, reason, and output your final response in the **same language** as the user's query.
- **CRITICAL**: When calling ANY tools (including search tools, predictors, database queries, etc.), all tool arguments, keywords, and technical parameters MUST be in **English**. Do not translate protein names, genes, or scientific terms into the user's language when passing them to tools.
\end{lstlisting}
\end{tcolorbox}

\subsubsection*{Prompt of Machine Learning Specialist}

\begin{tcolorbox}[
  enhanced,
  breakable,
  colback=terminalbg,
  colframe=terminalframe,
  boxrule=0.7pt,
  arc=2pt,
  left=6pt,
  right=6pt,
  top=6pt,
  bottom=6pt,
  title=\texttt{Prompt Template --- Machine Learning Specialist},
  colbacktitle=terminaltitle,
  coltitle=white,
  fonttitle=\bfseries\ttfamily,
  attach boxed title to top left={yshift=-2mm, xshift=2mm},
  boxed title style={
    sharp corners,
    boxrule=0pt,
    colframe=terminaltitle,
    colback=terminaltitle
  }
]
\label{app:note:prompt_machine_learning_specialist}

{\footnotesize\color{terminalaccent}
\noindent\textbf{Note.} The following text shows the full prompt template used for the \textbf{Machine Learning Specialist} role in VenusFactory.
}

\vspace{0.6em}

\begin{lstlisting}[style=terminalpromptstyle]
# Machine Learning Specialist (MLS)

You are VenusFactory, an AI assistant for protein engineering. You act as the **Machine Learning Specialist**: you **write code**, **execute tools**, and **debug**. You work with the Computational Biologist (CB)—CB selects tools and builds the pipeline; you implement and run the steps and fix errors.

**Post-step self-check (see machine_learning_specialist_post_step_check):**
{machine_learning_specialist_post_step_check}

**Execution protocol:** You **only start execution** (run a tool or load a skill) when **CB explicitly instructs you** to do so. Do not execute on your own initiative. If you find CB's instruction unreasonable (e.g. tool or skill not in the available list, parameters invalid, step cannot be done), **object or ask CB** before executing—do not blindly run. CB will then explain or adjust; only after that do you proceed with execution.

Your responsibilities:
- **Execute tools**: Run tools (including ML/code tools like `agent_generated_code`, `python_repl`, training, prediction) with the correct parameters.
- **Read and execute skills**: When CB or the plan asks to use a skill (e.g. rdkit, brenda_database), call `read_skill` with the skill_id to get the full SKILL.md, then follow the skill's instructions: write and run code via `agent_generated_code` or `python_repl`. All code you write, execution stdout/stderr, and any saved plot file paths are visible in the Gradio chat so the user and CB can see the discussion and results. **Capability self-check:** Before executing, verify that the requested tool or skill actually exists and matches what you are being asked to do. Do not fabricate or assume capabilities that are not in the available tools/skills; if the plan asks for something that has no corresponding tool or skill, recognize this and report to CB (e.g. that the step cannot be executed because the capability is not available) instead of inventing or failing silently.
- **Write code**: When the plan requires custom logic (e.g. data splitting, analysis scripts, or skill-based workflows), write and run code via `agent_generated_code` or `python_repl`. Use `python_repl` for quick scripts and plotting (e.g. matplotlib); use `agent_generated_code` for longer or file-heavy tasks.
- **Debug (self-check on failure)**: When **any** tool returns an error (`success: false` or exception), you must **self-debug** first: treat it as a distinct step, explicitly check whether the failure can be fixed by **replacing or adjusting parameters** (e.g. wrong type, missing required field, invalid value) and **retrying**. Only after considering parameter substitution and retry should you report to CB if the step cannot be fixed. Do not report to CB without first attempting this self-check.
- **Collaborate with CB**: If you cannot fix the error yourself, report to the Computational Biologist exactly what went wrong and what CB should do (e.g. which parameter is missing or invalid, or that the pipeline step needs different inputs). Discuss tool selection, pipeline order, and outputs (e.g. file paths, config paths) so the pipeline runs correctly.

---

**Full context (use for self-check):** The only tools that exist in this run are: **{available_tools_list}**. The only skills available (use `read_skill` with skill_id) are: **{available_skills_meta}**. Before executing, verify that the current step’s tool or skill is in these lists; if not, report to CB instead of inventing or assuming.

You will run a single tool per invocation: **{tool_name}**

Tool description:
{tool_description}

---

## EXECUTION WORKFLOW
1. Call the tool ONCE with the correct parameters.
2. Observe the tool's output (JSON format).
3. If the output contains `{{"success": true}}` → Return the output as your Final Answer.
4. If the output contains `{{"success": false}}` → **Self-check**: First consider whether parameters or inputs can be **replaced or adjusted** for a retry (e.g. fix type, fill missing field, try alternative value). If yes, output the corrected parameters and retry; only if retry is not feasible, return the error as your Final Answer and report to CB.
5. **Search tools empty results**: If the tool returns `success: true` but `references`, `results`, or `datasets` is empty or `[]`, **do not treat as final success**. Retry with (a) different **keywords** in English, and/or (b) a different tool in the same category (e.g. switch literature tool or web tool). Only after one or two retries with no results should you return the empty result as Final Answer.
6. **CRITICAL**: DO NOT call the tool again after receiving a successful output (with non-empty content). Stop immediately and provide your Final Answer. If the output has `SYSTEM_NOTE: STOP EXECUTION NOW`, you must obey it instantly.
7. **CRITICAL**: After the tool returns its result, you MUST immediately provide a Final Answer. The Final Answer should be the tool's JSON output, without any additional text.

## RESPONSE FORMAT
- **Step 1**: State your action (e.g. "I will now call the {tool_name} tool.")
- **Step 2**: Call the tool and wait for result.
- **Step 3**: Provide Final Answer: `Final Answer: <tool_output_json>`

## EXAMPLES
**Success:**
- Tool returns: `{{"success": true, "uniprot_id": "P04040", "sequence": "MADSRD..."}}`
- You: `Final Answer: {{"success": true, "uniprot_id": "P04040", "sequence": "MADSRD..."}}`

**Error (debug then answer):**
- Tool returns: `{{"success": false, "error": "File not found"}}`
- You: Fix the file path or input, re-run if possible; otherwise: `Final Answer: {{"success": false, "error": "..."}}`

## IMPORTANT
- Always provide `Final Answer: <json>` after the tool executes.
- Do NOT call the tool multiple times for the same step after success.
- Do NOT add extra text before or after the JSON in Final Answer.
- When in doubt about pipeline steps or tool order, **discuss with the Computational Biologist** before proceeding.

---
**Language:** Always respond in the same language as the user. Match the user's language for all output.

## Language & Tool Execution Rules
- You MUST answer, reason, and output your final response in the **same language** as the user's query.
- **CRITICAL**: When calling ANY tools (including search tools, predictors, database queries, etc.), all tool arguments, keywords, and technical parameters MUST be in **English**. Do not translate protein names, genes, or scientific terms into the user's language when passing them to tools.
\end{lstlisting}
\end{tcolorbox}

\subsubsection*{Prompt of Prompt of Computational Biologist}

\begin{tcolorbox}[
  enhanced,
  breakable,
  colback=terminalbg,
  colframe=terminalframe,
  boxrule=0.7pt,
  arc=2pt,
  left=6pt,
  right=6pt,
  top=6pt,
  bottom=6pt,
  title=\texttt{Prompt Template --- Prompt of Computational Biologist},
  colbacktitle=terminaltitle,
  coltitle=white,
  fonttitle=\bfseries\ttfamily,
  attach boxed title to top left={yshift=-2mm, xshift=2mm},
  boxed title style={
    sharp corners,
    boxrule=0pt,
    colframe=terminaltitle,
    colback=terminaltitle
  }
]
\label{app:note:prompt_computational_biologist}

{\footnotesize\color{terminalaccent}
\noindent\textbf{Note.} The following text shows the full prompt template used for the \textbf{Machine Learning Specialist} role in VenusFactory.
}

\vspace{0.6em}

\begin{lstlisting}[style=terminalpromptstyle]
# Computational Biologist (CB)

You are VenusFactory, an AI assistant for protein engineering. You act as the **Computational Biologist**: you **plan and verify** the pipeline; the **MLS (Machine Learning Specialist)** executes tools and debugs. **MLS only starts execution (opens a new dialog to run tools or load skills) when you explicitly instruct it to do so.** If MLS finds your instruction unreasonable and objects or asks to retry, you respond and explain or adjust the plan.

---

## Step planning and goal verification

{computational_biologist_step_planning}

---

## Mode A: Pipeline planner (from PI report → JSON)

When you receive **PI's research draft** and **Preliminary guidance** (suggested capabilities, feasible path) below, **design the concrete execution plan** based on PI's feasibility suggestions and the **available tools/skills**. PI provides high-level paths and domain-level guidance; **you are responsible for tool selection, parameter design, and step granularity**. You do **not** execute tools; MLS will run each step.

**Your responsibilities:**
1. **Design from tools:** Map PI's domain-level suggestions (e.g. "sequence retrieval", "structure prediction") to actual tools and skills from the Available lists. Only plan steps that use supported tools; adjust or substitute when PI suggests something with no direct match.
2. **Cover comprehensively:** Plan **as many steps as needed** to cover (a) **execution** (download, prediction, run), (b) **analysis** (read output files, parse results, summarize), and (c) **visualization** (plot, draw figures). Where the task involves data or results, include steps to **read files** (e.g. via `python_repl` or skill) and **generate plots** so the user gets both tables and figures in the conversation. Plots produced by tools are shown in the chat; plan steps that produce figures when useful.
3. **Split finely:** Split tasks into **sufficiently fine-grained steps** so MLS does **only one step at a time**. Do not merge multiple actions into a single step.
4. **Goals and success criteria:** For each step, define a **goal** (what must be achieved) and **success_criteria** (how to tell it succeeded, e.g. output has required field, file exists, plot saved). CB will check goal achievement after MLS completes; if not met, CB triggers retry or re-plan.
5. **Plan in detail:** Each step must be **as detailed and actionable as possible**—concrete tool names, explicit `tool_input` parameters (e.g. `uniprot_id`, `skill_id`, `task_description`, file paths from `dependency:step_N:field`), and clear `dependency:step_N:field` when a step uses output from a previous step. Avoid vague descriptions; MLS should be able to execute each step without guessing.
6. **Contingency:** Consider fallbacks and alternative paths. If a step might fail (e.g. a tool returns empty, a file is missing), include optional follow-up steps or note in `task_description` what MLS should do (e.g. "If no results, report to CB for re-planning" or "Try alternative source X"). When the report suggests multiple approaches (Option A / B), plan both or the primary path plus a fallback.

**Map PI's feasible path to concrete steps:** PI provides high-level steps; you expand them into sufficiently fine-grained pipeline steps (one JSON object per step). One PI step may map to one or more pipeline steps. **Do not include search steps** (query_pubmed, query_arxiv, etc.)—the PI has already gathered that information; only plan data download, prediction, training, and other execution tools. If Preliminary guidance is empty or says "No execution needed", infer from the draft or output `[]`.

**PI's research draft** — for context:
{pi_report}

**PI's Preliminary guidance** (suggested capabilities, feasible path) — use this to design the pipeline:
{pi_suggest_steps}

**Available tools (list of names — ONLY these can be executed; any other name will fail):**
{available_tools_list}

**Available tools (full names and parameters — use ONLY these; names and parameters must match exactly):**
{tools_description}

**Available skills (for MLS to read and execute — use ONLY these skill_ids):** You instruct MLS when to use a skill: MLS calls `read_skill` with the skill_id, then follows the skill document to write/run code (`agent_generated_code` or `python_repl`). Code and plot outputs are visible in the chat. MLS executes only when you explicitly tell it to run a tool or load a skill.
{skills_metadata}

**CRITICAL — Plan must be executable:** The pipeline you output will be executed by MLS. **Every step must use a tool name from the Available tools list above or a skill_id from Available skills.** If you use a tool or skill not in these lists, the step will fail. Use the **Available tools** and **Available skills** lists above as the ground truth. PI may describe in domain terms (e.g. "sequence retrieval", "structure prediction"); **you map these to actual tool names and parameters** from the lists. Only plan steps that use tools/skills that exist. If PI suggests something with no matching tool, omit or substitute with the closest available capability. Think independently: match PI’s intent to real tools/skills.

**Current protein context:**
{protein_context_summary}

**Recent tool outputs (if any):**
{tool_outputs}

**Output (language-neutral):** PI's Preliminary guidance may be in **any language** (e.g. English or 中文). You **must** output a **non-empty** JSON array whenever PI suggests **any** execution path—including when steps are in Chinese (e.g. **步骤**, 1. 2. 3., 一、二、三). Output `[]` **only when** PI **explicitly** states **no execution needed** (e.g. "No execution needed", "No execution steps", 无需执行, 无执行步骤). **Do not** output `[]` just because PI used domain terms instead of tool names. If you see a numbered path or suggested capabilities, design and output a non-empty array. **Never** output prose—your response must be **only** a JSON array. Output **only** a JSON **array** (no markdown, no code fence, no explanation): one object per step with `"step"`, `"goal"`, `"success_criteria"`, `"task_description"`, `"tool_name"`, `"tool_input"`. No prose. Do not collapse steps. After MLS completes each step, CB checks goal achievement; if not met, trigger retry or re-plan.

---

## Mode B: Tool executor (single step)

When you are asked to run **one tool** (tool_name and tool_description provided below), execute it and return the Final Answer. Collaborate with MLS on code/debug.

You will run a single tool per invocation: **{tool_name}**

Tool description:
{tool_description}

**EXECUTION:** Call the tool ONCE. If output has `{{"success": true}}` → return it as Final Answer. If `{{"success": false}}` → return error; discuss with MLS for debug. Provide `Final Answer: <tool_output_json>` with no extra text.

**EXAMPLES:** Success → `Final Answer: {{"success": true, "uniprot_id": "P04040", "sequence": "MADSRD..."}}`. Error → `Final Answer: {{"success": false, "error": "..."}}`.

---

## Language & Tool Execution Rules
- You MUST answer, reason, and output your final response in the **same language** that the user used in their query (e.g., if the user asks in Chinese, you must reply in Chinese).
- **CRITICAL**: When calling ANY tools (including search tools, predictors, database queries, etc.), all tool arguments, keywords, and technical parameters MUST be in **English**. Do not translate protein names, genes, or scientific terms into the user's language when passing them to tools.
\end{lstlisting}
\end{tcolorbox}

\subsubsection*{Prompt of Prompt of Scientific Critic}

\begin{tcolorbox}[
  enhanced,
  breakable,
  colback=terminalbg,
  colframe=terminalframe,
  boxrule=0.7pt,
  arc=2pt,
  left=6pt,
  right=6pt,
  top=6pt,
  bottom=6pt,
  title=\texttt{Prompt Template --- Prompt of Scientific Critic},
  colbacktitle=terminaltitle,
  coltitle=white,
  fonttitle=\bfseries\ttfamily,
  attach boxed title to top left={yshift=-2mm, xshift=2mm},
  boxed title style={
    sharp corners,
    boxrule=0pt,
    colframe=terminaltitle,
    colback=terminaltitle
  }
]
\label{app:note:prompt_scientific_critic}

{\footnotesize\color{terminalaccent}
\noindent\textbf{Note.} The following text shows the full prompt template used for the \textbf{Scientific Critic} role in VenusFactory.
}

\vspace{0.6em}

\begin{lstlisting}[style=terminalpromptstyle]

# Scientific Critic (SC)

You are VenusFactory, an AI assistant for protein engineering. You act as the **Scientific Critic**: you **summarize** the run—synthesize execution info and tool outputs into a clear, evidence-based report for the user; or answer directly when no pipeline has run.

---

## When you receive a full run (synthesis)

You are given the **full run record** (all agent outputs and tool executions), so you see everything that happened before your summary:

1. **{full_run_record}** — Complete transcript: user message, Principal Investigator (research draft + suggest steps), Computational Biologist (pipeline plan + verification), Machine Learning Specialist (each step execution and result), and every **tool execution** (tool name, input, output). Use this to ground your conclusions.
2. **User request:** {original_input}
3. **Step-wise analysis log:** {analysis_log}
4. **References (optional):** {references}

Synthesize into one final report for the user. Respond in the same language as the user.

**Rules:**
1. **Conclusions** — List 1–3 clear, numbered conclusions (each ≤ 2 sentences) that directly answer the user's question(s).
2. **Supporting Evidence** — For each conclusion, cite concrete items from the analysis_log (quote or summarize) and reference index [n] where applicable.
3. **Rationale** — Brief paragraph per conclusion (1–3 sentences) explaining why the evidence supports it.
4. **Confidence & Caveats** — Summarize uncertainty and assumptions.
5. **Practical Recommendations** — 1–4 clear next steps (experiments, checks, or analyses).
6. **References** — If references are provided, parse and list ONLY cited references in a deduplicated `References` section: [n] Title. Authors. Year. Source. URL. DOI. Include OSS/download URLs from the analysis_log as clickable links where relevant.

**Formatting:** Use Markdown headings: Conclusions, Supporting Evidence, Rationale, Confidence & Caveats, Practical Recommendations, References (only if references exist and were cited). Be concise; avoid speculation. If the user asked multiple questions, answer point-by-point (P1, P2, …).

---

## When the user sends a direct message (no pipeline run)

If there is no analysis_log (e.g. the user is chatting with you directly):
- Answer as a knowledgeable scientific critic: explain clearly, analyze concepts, note caveats.
- Do **not** use the "final report" format above. Use a **conversational, analytical** style.
- If the question would benefit from running tools, suggest using the agent workflow; do not pretend you have already run tools.
- Be concise; respond in the same language as the user.

---

## Language & Tool Execution Rules
- You MUST answer, reason, and output your final response in the **same language** that the user used in their query (e.g., if the user asks in Chinese, you must reply in Chinese).
- **CRITICAL**: When calling ANY tools (including search tools, predictors, database queries, etc.), all tool arguments, keywords, and technical parameters MUST be in **English**. Do not translate protein names, genes, or scientific terms into the user's language when passing them to tools.
\end{lstlisting}

\end{tcolorbox}

\section{Case Study}\label{app:sec:casestudy}
\subsection*{Case Study 1: Protein Allergenicity Prediction}

\begin{humanprompt}
Please evaluate the allergenicity potential for these modified food enzymes. The protein sequence is in the uploaded file.
\end{humanprompt}

\vspace{0.6em}

\begin{agentstep}{Task Planning}

\textbf{Agent:} \textit{Principal Investigator}

\vspace{0.4em}

\textbf{Thought:} \\
The task is to predict allergenicity as a part of protein function prediction. The Principal Investigator first searches the VenusFactory platform for existing protein function prediction tools. However, no module is available for allergenic protein prediction. Therefore, relevant data need to be collected, and a new protein function prediction tool must be trained to support protein allergenicity prediction.

\vspace{0.4em}

\textbf{Generate Plan:}
\begin{itemize}
    \item Step 1: Search papers related to allergenicity prediction.
    \item Step 2: Use web search to find datasets for allergen / non-allergen prediction.
    \item Step 3: Split the downloaded training data into an 8:2 training and validation set.
    \item Step 4: Generate the training configuration file for the ESM2-8M model with LoRA.
    \item Step 5: Train the model using the generated configuration and report performance.
    \item Step 6: Package the trained checkpoint as a new MCP service, and register it into the VenusFactory tool registry.
    \item Step 7: Use the new MCP tool to predict the function of sequences in the FASTA file.
\end{itemize}

\vspace{0.4em}

\textbf{Comment:} \\
Decomposes these requirements into executable sub-tasks and assigns them to the Machine Learning Specialist and Computational Biologist.

\end{agentstep}

\vspace{0.6em}
\begin{agentstep}{Step 1: Search papers related to allergenicity prediction}
\textbf{Agent:} \textit{Computational Biologist}
\vspace{0.4em}

\textbf{Thought:} \\
The Computational Biologist invokes the Literature Search tool to retrieve relevant publications and collect background information and data sources for protein allergenicity prediction.
\vspace{0.4em}

\textbf{Invoking:} \\
\texttt{literature\_search} with \{query: "Allergenic protein and computational-based prediction algorithms"\}
\vspace{0.2em}

\textit{Responded:} I will now call the \texttt{literature\_search} tool.

\vspace{0.8em}
\hrule
\vspace{0.8em}

\textbf{Output:}
\vspace{0.3em}

\begin{Verbatim}[
  fontsize=\small,
  breaklines=true,
  breakanywhere=true,
  breaksymbolleft={}
]
{"success": true,
 "references": [
  {"source": "arXiv",
   "title": "AlphaFold predicts the most complex protein knot and composite protein knots",
   "url": "https://arxiv.org/pdf/2207.07410v1",
   "authors": ["Maarten A. Brems", "et al."],
   "year": "2022-07-15T11:38:45Z",
   "abstract": "The computer artificial intelligence system AlphaFold has recently predicted previously ...",
   "doi": "10.1002/pro.4380"},
  {"source": "arXiv",
   "title": "Mass Balance Approximation of Unfolding Improves Potential-Like Methods for Protein Stability Predictions",
   "url": "https://arxiv.org/pdf/2504.06806v1",
   "authors": ["Ivan Rossi", "et al."],
   "year": "2025-04-09T11:53:02Z",
   "abstract": "The prediction of protein stability changes following single-point mutations plays ...",
   "doi": "10.1002/pro.70134"},
  {"source": "arXiv",
   "title": "Driving Accurate Allergen Prediction with Protein Language Models and Generalization-Focused Evaluation",
   "url": "https://arxiv.org/pdf/2508.10541v1",
   "authors": ["Brian Shing-Hei Wong", "et al."],
   "year": "2025-08-14T11:30:20Z",
   "abstract": "Allergens, typically proteins capable of triggering adverse immune responses, represent ...",
   "doi": ""},
  {"source": "arXiv",
   "title": "ISLAND: In-Silico Prediction of Proteins Binding Affinity Using Sequence Descriptors",
   "url": "https://arxiv.org/pdf/1711.10540v2",
   "authors": ["Wajid Arshad Abbasi", "et al."],
   "year": "2017-11-22T07:54:31Z",
   "abstract": "Determination of binding affinity of proteins in the formation of protein ...",
   "doi": "10.1186/s13040-020-00231-w"}
 ]}
\end{Verbatim}

\vspace{0.5em}
\hrule
\vspace{0.8em}

\textbf{Comment:} \\
Retrieved literature related to protein allergenicity prediction and used it as background information to support the response.

\end{agentstep}

\vspace{0.6em}

\begin{agentstep}{Step 2: Use web search to find datasets for allergen / non-allergen prediction.}
\textbf{Agent:} \textit{Computational Biologist}
\vspace{0.4em}

\textbf{Thought:} \\
The Computational Biologist invokes the Deep Search tool to identify available datasets and related resources from platforms such as Hugging Face and Kaggle.

\vspace{0.4em}

\textbf{Invoking:} \\
\texttt{deep\_search} with \{query: "search allergenic protein sequences datasets"\}

\vspace{0.2em}
\textit{Responded:} I will now call the \texttt{deep\_search} tool.

\vspace{0.8em}
\hrule
\vspace{0.8em}

\textbf{Output:}
\vspace{0.3em}

\begin{Verbatim}[
  fontsize=\small,
  breaklines=true,
  breakanywhere=true,
  breaksymbolleft={}
]
{"success": true, 
    "query": "search allergenic protein sequence datasets", 
    "results": [
        {"title": "Protein_allergenic from Hugging Face Database", 
        "url": 
        "description": "Protein data related to the computational task of allergenic protein identification were collected from public datasets.", 
        "source": "Web Search", 
        {"title": "SDAP 2.0 : Structural Database of Allergenic Proteins", 
        "url": "https://fermi.utmb.edu/", 
        "description": "We used the collected data on protein sequences of allergenic proteins ...", 
        "source": "Web Search", }, 
        {"title": "The updated Structural Database of Allergenic Proteins (SDAP 2.0 ...", 
        "url": "https://www.sciencedirect.com/science/article/pii/S2772829323000875", 
        "description": "We aim to present the features and use of the updated Structural ...", 
        "source": "Web Search"}, 
        {"title": "AllergenOnline", 
        "url": "http://www.allergenonline.org/", 
        "description": "AllergenOnline provides access to a peer reviewed allergen list and ...", 
        "source": "Web Search"}, 
        {"title": "Compare Database - Allergen Database", 
        "url": "https://comparedatabase.org/", 
        "description": "The COMPARE Allergen Database What it is - the COM prehensive P rotein A llergen ...", 
        "source": "Web Search"}
 ]}
\end{Verbatim}

\vspace{0.5em}
\hrule
\vspace{0.8em}

\textbf{Comment:} \\
Retrieved literature related to protein allergenicity prediction and used it as background information to support the response.

\end{agentstep}
\vspace{0.6em}

\begin{agentstep}
{Step 3: Split the downloaded training data into an 8:2 training and validation set.} 
\textbf{Agent:} \textit{Machine Learning Specialist}
\vspace{0.4em}

\textbf{Thought:} \\
The Machine Learning Specialist then retrieves the LrZhang/Protein\_allergenic dataset from Hugging Face and writes code to split the data into training and validation sets with an 8:2 ratio.
\vspace{0.4em}

\textbf{Invoking:} \\
\texttt{ai\_code\_execution} with \ {\texttt{
\{"input\_files": ["LrZhang/Protein\_allergenic"], "task\_description": "Download the dataset from Hugging Face. Then, shuffle the data and split it into a training set (80\%) and a test set (20\%). Save the new splits as 'train\_split.csv' and 'test\_split.csv'. The output should be a JSON object with keys 'train\_path' and 'test\_path' containing the full paths to the newly created files."\}
}}\\

\vspace{0.2em}

\textit{Responded:} I will now call the \texttt{ai\_code\_execution} tool.

\vspace{0.8em}
\hrule
\vspace{0.8em}

\textbf{Output:}
\vspace{0.3em}

\begin{Verbatim}[
  fontsize=\small,
  breaklines=true,
  breakanywhere=true,
  breaksymbolleft={}
]
{
  "success": true,
  "output_files": [
    {
      "name": "train_split.csv",
      "path": "temp_outputs/2026/01/02/11_23_03/train_split.csv"   
    },
    {
      "name": "test_split.csv",
      "path": "temp_outputs/2026/01/02/11_23_03/test_split.csv"    
    }
  ],
  "summary": "Successfully concatenated, shuffled, and split data into an 80% training set and a 20% test set.",
  "model_info": {},
  "details": {
    "train_path": "temp_outputs/2026/01/02/11_23_03/train_split.csv",
    "test_path": "temp_outputs/2026/01/02/11_23_03/test_split.csv",    
    "input_files": [
      "LrZhang/Protein_allergenic"
    ],
    "combined_data_shape": {
      "rows": 7100,
      "columns": 2
    },
    "new_train_set_shape": {
      "rows": 5680,
      "columns": 2
    },
    "new_test_set_shape": {
      "rows": 1420,
      "columns": 2
    }
  },
  "generated_code_path": "temp_outputs/2026/01/02/11_23_03/generated_scripts/generated_code_20260119_135809_1d7f55c5.py"
\end{Verbatim}

\vspace{0.5em}
\hrule
\vspace{0.8em}

\textbf{Comment:} \\
The execution results show that the dataset was successfully retrieved from \texttt{LrZhang/Protein\_allergenic} on Hugging Face and split into training and test sets. The processed files were saved under the directory \texttt{temp\_outputs/2026/01/02/11\_23\_03}.%

\end{agentstep}

\vspace{0.6em}
\begin{agentstep}{Step 4: Generate the training configuration file for the ESM2-8M model with LoRA.}
\textbf{Agent:} \textit{Machine Learning Specialist}
\vspace{0.4em}

\textbf{Thought:} \\
The Machine Learning Specialist then retrieves the LrZhang/Protein\_allergenic dataset from Hugging Face and writes code to split the data into training and validation sets with an 8:2 ratio.
\vspace{0.4em}

\textbf{Invoking:}\\
\texttt{generate\_training\_config} with
\texttt{\{csv\_file: temp\_outputs\_2026\_01\_02\_11\_23\_03\_ train\_split.csv,\;
output\_name: esm2\_8m\_lora\_allergen\_config,\;
test\_csv\_file: temp\_outputs/2026/01/02/11\_23\_03/test\_split.csv,\;
user\_requirements: Train an ESM2-8M model with LoRA for allergen prediction.\}}
\vspace{0.2em}

\textit{Responded:} I will now call the \texttt{generate\_training\_config} tool.

\vspace{0.8em}
\hrule
\vspace{0.8em}

\textbf{Output:}

\vspace{0.3em}

\begin{Verbatim}[
  fontsize=\small, 
  breaklines=true,
  breakanywhere=true,
  breaksymbolleft={}
]
{
  "dataset_selection": "Custom Dataset",
  "dataset_custom": "temp_outputs/2026/01/02/11_23_03",
  "problem_type": "single_label_classification",
  "num_labels": 2,
  "metrics": [
    "accuracy",
    "mcc",
    "f1",
    "precision",
    "recall",
    "auroc"
  ],
  "sequence_column_name": "seq",
  "label_column_name": "label",
  "plm_model": "ESM2-8M",
  "training_method": "plm-lora",
  "pooling_method": "mean",
  "batch_mode": "Batch Size Mode",
  "batch_size": 16,
  "learning_rate": 5e-05,
  "num_epochs": 2,
  "max_seq_len": 1098,
  "patience": 10,
  "gradient_accumulation_steps": 1,
  "warmup_steps": 100,
  "scheduler": "linear",
  "max_grad_norm": 1.0,
  "num_workers": 4,
  "monitored_metrics": "accuracy",
  "monitored_strategy": "max",
  "output_model_name": "model_train_split.pt",
  "output_dir": "ckpt/train_split",
  "wandb_enabled": false,
  "lora_r": 8,
  "lora_alpha": 32,
  "lora_dropout": 0.1,
  "lora_target_modules": "query,key,value",
  "test_file": "temp_outputs/2026/01/02/11_23_03/test_split.csv"
}
\end{Verbatim}

\vspace{0.5em}
\hrule
\vspace{0.8em}

\textbf{Comment:}
The generated training configuration file records the dataset paths and key training settings in a standardized parameter format, including the inputs for both the training and validation splits. 
The configuration is then used to drive the downstream training pipeline, supporting traceability and reproducibility of the training process.

\noindent The path of the training configuration file is:
\path{temp_outputs/2026/01/02/training_pipeline/configs/esm2_8m_lora_classification_config_1767324344.json}.

\end{agentstep}

\vspace{0.6em}
\begin{agentstep}{Step 5: Train the model using the generated configuration and report performance.}
\textbf{Agent:} \textit{Machine Learning Specialist}
\vspace{0.4em}

\textbf{Thought:} \\
The Machine Learning Specialist then launches and executes the model training script based on the generated training configuration file to carry out the downstream training process.
\vspace{0.4em}

\textbf{Invoking:} \\
\texttt{train\_protein\_model} with \texttt{\{config\_path: }\path{temp_outputs/2026/01/02/training_pipeline/configs/esm2_8m_lora_classification_config_1767324344.json}\texttt{ \}.}
\vspace{0.2em}

\textit{Responded:} I will now call the \texttt{train\_protein\_model} tool.

\vspace{0.8em}
\hrule
\vspace{0.8em}

\textbf{Output:}
\vspace{0.3em}

\begin{Verbatim}[
   fontsize=\small, 
   breaklines=true,
   breakanywhere=true,
   breaksymbolleft={}
]
{
  "success": true,
  "message": "Model training completed successfully!",
  "model_path": "ckpt/train_split/model_train_split.pt",
  "output_dir": "ckpt/train_split",
  "command": "python src/train.py 
            --plm_model facebook/esm2_t6_8M_UR50D 
            --pooling_method mean
            --training_method plm-lora 
            --dataset /home/tanyang/VenusFactory/temp_outputs/2026/01/02/11_23_03 
            --problem_type single_label_classification 
            --num_labels 2 
            --metrics accuracy,mcc,f1,precision,recall,auroc 
            --sequence_column_name seq 
            --label_column_name label 
            --learning_rate 5e-05 
            --num_epochs 2 
            --max_seq_len 1080 
            --gradient_accumulation_steps 1 
            --warmup_steps 100 
            --scheduler linear 
            --patience 10 
            --num_workers 4 
            --max_grad_norm 1.0 
            --monitor accuracy 
            --monitor_strategy max 
            --batch_size 16 
            --lora_r 8 
            --lora_alpha 32 
            --lora_dropout 0.1 
            --lora_target_modules query key value 
            --output_model_name model_train_split.pt 
            --output_dir ckpt/train_split",
  "logs":
2026-01-02 11:25:49 - training - INFO - Test Results:\n2026-01-02 11:25:49 - training - INFO - Test Loss: 0.3009\n2026-01-02 11:25:49 - training - INFO - Test accuracy: 0.8732\n2026-01-02 11:25:49 - training - INFO - Test mcc: 0.7465\n2026-01-02 11:25:49 - training - INFO - Test f1: 0.8727\n2026-01-02 11:25:49 - training - INFO - Test precision: 0.8764\n2026-01-02 11:25:49 - training - INFO - Test recall: 0.8690\n2026-01-02 11:25:499 - training - INFO - Test auroc: 0.9454\n "
}
\end{Verbatim}

\vspace{0.5em}
\hrule
\vspace{0.8em}

\textbf{Comment:}
After training, evaluation on the test set yielded a test accuracy of 0.8732, a Matthews correlation coefficient of 0.7465, and a test AUROC of 0.9454.

\end{agentstep}

\vspace{0.6em}
\begin{agentstep}{Step 6: Package the trained checkpoint as a new MCP service.}
\textbf{Agent:} \textit{Machine Learning Specialist}
\vspace{0.4em}

\textbf{Thought:} After training, the Machine Learning Specialist wraps the checkpoint into a standardized prediction interface and registers it into the system tool set, allowing the Computational Biologist to invoke it as a callable prediction tool in downstream tasks. Model usability is verified through configuration-based inference testing.
\vspace{0.4em}
\end{agentstep}

\vspace{0.6em}
\begin{agentstep}{Step 7: Use the new MCP tool to predict sequences in the FASTA file.}
\textbf{Agent:} \textit{Computational Biologist}
\vspace{0.4em}

\textbf{Thought:} \\
The Machine Learning Specialist then launches and executes the model training script based on the generated training configuration file to carry out the downstream training process.
\vspace{0.4em}

\textbf{Invoking:} \\
{\ttfamily\small
predict\_with\_protein\_model with \seqsplit{\{'config\_path': `temp\_outputs/2026/01/02/training\_pipeline/configs/esm2\_8m\_lora\_classification\_config\_1767324344.json',
'sequence': `MKVLILACLVALALARELDELNVPGEIVESLSSSEESITRINKKIEKFQTEEQQQQTEDELQDKIHPFAQTQSLVYPFPGPIPNSLPQNIPPLTQTPVVVPPFLQPEVMGVSKVKEAMAPKHKEMPFPKYPVEPFTESQSLTLTDVENLHLPLPLLQSWMHQPHQPLPPTVMFPPQSVLSLSQSKVLPVPQKAVPYPQRDMPIQAFLLYQEPVLGPVRGPFPIIV'\}}}

\vspace{0.2em}

\textit{Responded:} I will now call the \texttt{predict\_with\_protein\_model} tool.

\vspace{0.8em}
\hrule
\vspace{0.3em}

\textbf{Output:}
\vspace{0.3em}

\begin{Verbatim}[
  fontsize=\small, 
  breaklines=true, 
  breakanywhere=true,
  breaksymbolleft={}
]
{
  "success": true,
  "message": "Prediction completed successfully!",
  "output_file": "/tmp/tmp5tkee33s/predictions.csv",
  "command": "python src/predict_batch.py 
            --model_path ckpt/train_split/model_train_split.pt 
            --plm_model facebook/esm2_t6_8M_UR50D 
            --eval_method plm-lora 
            --pooling_method mean 
            --problem_type single_label_classification 
            --num_labels 2 
            --max_seq_len 1080 
            --batch_size 16 
            --input_file /tmp/tmp5tkee33s/temp_sequence.csv 
            --output_dir /tmp/tmp5tkee33s 
            --output_file predictions.csv",
  "logs": "Number of attention heads: 8\n---------- Reading input file: /tmp/tmp5tkee33s/temp_sequence.csv ----------\nFound 1 sequences in input file\n---------- Processing sequences ----------\nPredicting:   0%|          | 0/1 [00:00<?, ?it/s]\nPredicting: 100%|██████████| 1/1 [00:00<00:00,  1.89it/s]\nPredicting: 100%|██████████| 1/1 [00:00<00:00,  1.89it/s]\n---------- Saving results to /tmp/tmp5tkee33s/predictions.csv ----------\nSaved 1 prediction results\n---------- Batch prediction completed successfully ----------",
  "preview": [
    {
      "aa_seq": "MKVLILACLVALALARELDELNVPGEIVESLSSSEESITRINKKIEKFQTEEQQQQTEDELQDKIHPFAQTQSLVYPFPGPIPNSLPQNIPPLTQTPVVVPPFLQPEVMGVSKVKEAMAPKHKEMPFPKYPVEPFTESQSLTLTDVENLHLPLPLLQSWMHQPHQPLPPTVMFPPQSVLSLSQSKVLPVPQKAVPYPQRDMPIQAFLLYQEPVLGPVRGPFPIIV",
      "predicted_class": 1,
      "class_0_prob": 0.3118185102939605,
      "class_1_prob": 0.6881815195083618
    }
  ],
  "total_predictions": 1
}

\end{Verbatim}

\vspace{0.5em}
\hrule
\vspace{0.8em}

\textbf{Comment:}
Using the trained prediction model, the provided five-point Beta-casein mutant sequence was predicted as class 1 with a confidence score of 0.688, suggesting allergenicity-related characteristics.
\end{agentstep}

\vspace{0.6em}
\begin{agentstep}{Scientific Critic}

\noindent\textbf{Response:}
\begin{enumerate}
  \item \textbf{The modified food enzyme sequence was predicted as allergenic (class 1).}\\
  The prediction confidence score was \textbf{0.688}, indicating a higher probability for class 1 than class 0.

  \item \textbf{No allergenicity prediction tool was initially available in the platform, so a task-specific predictor was trained and then applied.}\\
  The workflow collected data, trained an ESM2-8M model with LoRA, and used the trained checkpoint for downstream inference.

  \item \textbf{The trained model achieved consistent performance on the held-out test set.}\\
  Evaluation reported a test accuracy of \textbf{0.8732}, a Matthews correlation coefficient of \textbf{0.7465}, and a test AUROC of \textbf{0.9454}.
\end{enumerate}

\noindent\textbf{Supporting Evidence}

\vspace{0.4em}
\noindent\textbf{For Conclusion 1}
\begin{itemize}
  \item Batch prediction output reported:
  \begin{itemize}
    \item "predicted\_class": 1
    \item "class\_1\_prob": 0.6881815195083618
    \item "class\_0\_prob": 0.3118185102939605
  \end{itemize}
\end{itemize}

\vspace{0.4em}
\noindent\textbf{For Conclusion 2}
\begin{itemize}
  \item Task planning explicitly noted that the platform did not support allergen prediction:
  \begin{itemize}
    \item \emph{``no module is available for allergenic protein prediction. Therefore, relevant data need to be collected, and a new protein function prediction tool must be trained\ldots''}
  \end{itemize}

  \item Dataset retrieval and preparation were performed using the Hugging Face dataset \textbf{LrZhang/Protein\_allergenic}.
  \begin{itemize}
    \item Training and test splits were created with an \textbf{8:2 ratio} and saved under:
    \begin{itemize}
      \item \code{temp\_outputs/2026/01/02/11\_23\_03/train\_split.csv}
      \item \code{temp\_outputs/2026/01/02/11\_23\_03/test\_split.csv}
    \end{itemize}
  \end{itemize}

  \item Model training was executed using the generated configuration file:
  \begin{itemize}
    \item \code{\path{temp_outputs/2026/01/02/training_pipeline/configs/esm2_8m_lora_classification_config_1767324344.json}}
    \item The trained checkpoint was saved to:
    \begin{itemize}
      \item \code{ckpt/train\_split/model\_train\_split.pt}
    \end{itemize}
  \end{itemize}
\end{itemize}

\vspace{0.4em}
\noindent\textbf{For Conclusion 3}
\begin{itemize}
  \item Training logs reported test performance:
  \begin{itemize}
    \item \code{Test accuracy: 0.8732}
    \item \code{Test mcc: 0.7465}
    \item \code{Test auroc: 0.9454}
  \end{itemize}
\end{itemize}

\noindent\textbf{Rationale}

\vspace{0.4em}
\noindent\textbf{For Conclusion 1}\\
The final allergenicity assessment directly follows the trained model's output, where class 1 probability (0.688) exceeds class 0 probability (0.312). The sequence is therefore classified as class 1 under the trained predictor.

\vspace{0.5em}
\noindent\textbf{For Conclusion 2}\\
The execution trace shows that allergenicity prediction was not supported by an existing tool in the platform. As a result, the workflow proceeded with dataset search, training data preparation, model training, and model invocation as a callable predictor for downstream evaluation.

\vspace{0.5em}
\noindent\textbf{For Conclusion 3}\\
The test set metrics indicate that the trained model can distinguish class 0 and class 1 in the held-out dataset. The reported AUROC of 0.9454 suggests good separability on this evaluation split, while accuracy and Matthews correlation coefficient provide additional evidence of predictive consistency.

\noindent\textbf{Confidence \& Caveats}
\begin{itemize}
  \item The prediction confidence score (0.688) is moderate and does not constitute experimental validation of allergenicity.
  \item The evaluation results are based on a single dataset split (80\% training, 20\% test). Performance may vary across datasets or enzyme families not represented in the training data.
  \item The current result reflects the model's classification behavior, not a clinical or regulatory conclusion. Additional validation is required for safety assessment.
\end{itemize}

\noindent\textbf{Practical Recommendations}
\begin{enumerate}
  \item Validate the prediction using established bioinformatics baselines, such as sequence similarity search against curated allergen databases.
  \item If available, test the model on additional independent datasets to check robustness across different food enzyme families.
  \item Perform sensitivity analysis by scoring closely related enzyme variants to assess prediction stability under small sequence changes.
  \item For downstream safety evaluation, consider experimental validation steps (for example, IgE binding assays), especially when predicted confidence is above a practical threshold.
\end{enumerate}

\noindent\textbf{References}
\begin{itemize}
  \item {[1]} Driving Accurate Allergen Prediction with Protein Language Models and Generalization-Focused Evaluation. Brian Shing-Hei Wong \emph{et al.} 2025. arXiv. \href{https://arxiv.org/pdf/2508.10541v1}{PDF}.
  \item {[2]} AllergenOnline. Food Allergy Research and Resource Program. Web resource. \href{http://www.allergenonline.org/}{Link}.
  \item {[3]} Compare Database -- Allergen Database (COMPARE). Web resource. \href{https://comparedatabase.org/}{Link}.
\end{itemize}

\end{agentstep}

\subsection*{Case Study 2: Directed Evolution of VHH}
\begin{humanprompt}
I have a specific VHH antibody with the protein sequence
\seqsplit{MQVQLVESGGGLAQAGGSLRLSCAVSGMPEFARAMGWFRQAPGKERELLAAIEGIGATTYYADSVKGRFTISRDDAANTVLLQMNSLKPDDTAVYYCAAAFSVTIPTRARHWVDWGPGTLVTVSSDDDDKSGGGGSHHHHHH}.
The current goal is to improve the binding affinity of this antibody.
\end{humanprompt}

\vspace{0.6em}
\begin{agentstep}{Task Planning}
\textbf{Agent:} \textit{Principal Investigator}
\vspace{0.4em}

\textbf{Thought:} \\
The task is to identify a single-point mutation design problem aimed at improving binding affinity. The user provides a VHH sequence and clearly states the goal of enhancing its binding performance. Therefore, the first step is to frame this as a structure-guided single-site mutation optimization task. This requires analyzing the protein structure, evaluating the potential impact of individual amino acid substitutions, and prioritizing mutations that are likely to increase affinity while maintaining structural stability.
\vspace{0.4em}

\noindent\textbf{Generate Plan:}
\begin{itemize}
	\item Step 1: Based on the provided VHH sequence, use AlphaFold to predict its three-dimensional structure to prepare for structure-based mutation prediction.
    \item Step 2: Based on the predicted structure, perform structural homology search using FoldSeek.
    \item Step 3: Use VenusREM to conduct zero-shot mutation prediction.
\end{itemize}

\textbf{Comment:}
Decomposes these requirements into executable sub-tasks and assigns them to the Computational Biologist for protein structure prediction, foldseek alignment, and zero-shot mutation prediction.
\end{agentstep}

\vspace{0.6em}
\begin{agentstep}{{Step 1: Based on the provided VHH sequence, use AlphaFold to predict its three-dimensional structure to prepare for structure-based mutation prediction.}}
\textbf{Agent:} \textit{Computational Biologist}
\vspace{0.4em}

\textbf{Thought:} \\
The Computational Biologist invokes the Protein Structure Prediction tool to predict the three-dimensional structure based on the provided sequence.
\vspace{0.4em}

\textbf{Invoking:}\\
\texttt{'protein\_structure\_prediction\_AlphaFold2' with \{'save\_path': 'vhh\_antibody\_predicted\_structure.pdb', 'sequence': '\seqsplit{MQVQLVESGGGLAQAGGSLRLSCAVSGMPEFARAMGWFRQAPGKERELLAAIEGIGATTYYADSVKGRFTISRDDAANTVLLQMNSLKPDDTAVYYCAAAFSVTIPTRARHWVDWGPGTLVTVSSDDDDKSGGGGSHHHHHH}'\}}\\
\vspace{0.2em}

\textit{Responded:} I will now call the \texttt{protein\_structure\_prediction\_AlphaFold} tool.

\vspace{0.8em}
\hrule
\vspace{0.8em}

\textbf{Output:}
\vspace{0.3em}
\begin{Verbatim}[
  fontsize=\small, 
  breaklines=true,
  breakanywhere=true,
  breaksymbolleft={}
]
{"success": true,
  "pdb_path": "vhh_antibody_predicted_structure.pdb\\pdb_8568.pdb",
  "result_info": {
    "file_name": "pdb_8568.pdb",
    "local_path": "vhh_antibody_predicted_structure.pdb\\pdb_8568.pdb",
    "sequence_length": 142
  }
}
]}
\end{Verbatim}

\vspace{0.5em}
\hrule
\vspace{0.8em}

\noindent\textbf{Comment:}
Predict the protein structure using AlphaFold, and use the predicted 3D model as input for downstream zero-shot mutation prediction.

\end{agentstep}

\vspace{0.6em}
\begin{agentstep}{{Step 2: Based on the predicted structure, perform structural homology search using FoldSeek.}}
\textbf{Agent:} \textit{Computational Biologist}
\vspace{0.4em}

\textbf{Thought:} \\
The Computational Biologist invokes the FoldSeek ClusterSearch tool to search for structural homologs in databases such as afdb50, afdb-proteome, cath50, and pdb100.
\vspace{0.4em}

\textbf{Invoking:} \\
\texttt{'foldseek\_search' with \{'pdb\_file\_path': 'predicted\_vhh\_structure.pdb/pdb\_6739.pdb', 'protect\_end': 142, 'protect\_start': 1\}}\\
\vspace{0.2em}

\textit{Responded:} I will now call the \texttt{foldseek\_search} tool.

\vspace{0.8em}
\hrule
\vspace{0.8em}

\textbf{Output:}
\vspace{0.3em}

\begin{Verbatim}[
  fontsize=\small, 
  breaklines=true,
  breakanywhere=true,
  breaksymbolleft={}
]
{"success": true,
"fasta_file":"FoldSeek/Download_data/pdb_6739_20260215002556.fasta", 
"total_sequences": 161
}
]}
\end{Verbatim}

\vspace{0.5em}
\hrule
\vspace{0.8em}

\textbf{Comment:}
Generate structural alignments using the FoldSeek API, and use the retrieved homology information to support VenusREM analysis and the final response.
\end{agentstep}

\vspace{0.6em}
\begin{agentstep}{Step 3: Use VenusREM to conduct zero-shot mutation prediction.}
\textbf{Agent:} \textit{Computational Biologist}
\vspace{0.4em}

\textbf{Thought:} \\
The Computational Biologist invokes the VenusREM (foldseek-based) for structure-based and foldseek-based zero-shot mutation scoring to tank candidate variants for directed evolution.
\vspace{0.4em}

\textbf{Invoking:} \\
\texttt{'zero\_shot\_structure\_prediction' with \{'model\_name': 'VenusREM (foldseek-based)', 'structure\_file': ' vhh\_antibody\_predicted\_structure.pdb\\pdb\_8568.pdb '\}}\\
\vspace{0.2em}

\textit{Responded:} I will now call the \texttt{zero\_shot\_structure\_prediction} tool.

\vspace{0.8em}
\hrule
\vspace{0.8em}

\textbf{Output:}
\vspace{0.3em}
\begin{Verbatim}[
   fontsize=\small, 
   breaklines=true,
   breakanywhere=true,
   breaksymbolleft={}
]
{"success": true,
{
  "headers": [
    "Mutant",
    "Prediction Rank",
    "Prediction Score"
  ],
  "data": [
    ["E7V", 1, 1.0], ["E7I", 2, 0.89], ["E7A", 3, 0.88],["E7L", 4, 0.87], ["E7T", 5, 0.79], ["R110P", 6, 0.76], ["R46P", 7, 0.75], ["S136H", 8, 0.73], ["L81T", 9, 0.73], ["E7M", 10, 0.72], ["S136G", 11, 0.72], ["F101E", 12, 0.71], ["F101D", 13, 0.71], ["M35V", 14, 0.69], ["E7S", 15, 0.69], ["E7P", 16, 0.68], ["M35W", 17, 0.68], ["I105D", 18, 0.67], ["M35I", 19, 0.67], ["S102P", 20, 0.67], ["M35Y", 21, 0.66], ["R33P", 22, 0.66], ["M84I", 23, 0.66], ["R33G", 24, 0.65], ["F101P", 25, 0.65], ["L12S", 26, 0.65], ["F101A", 27, 0.65], ["M35F", 28, 0.65], ["F101S", 29, 0.64], ["S86D", 30, 0.64], ["E7G", 31, 0.64], ["A32P", 32, 0.63], ["E7F", 33, 0.63], ["F101N", 34, 0.63], ["L81S", 35, 0.63], ["A34D", 36, 0.63], ["L12T", 37, 0.62], ["I52T", 38, 0.62], ["A13V", 39, 0.62], ["I105P", 40, 0.61], ["Q14R", 41, 0.61], ["R33V", 42, 0.61], ["F31D", 43, 0.61], ["N85S", 44, 0.61], ["M28L", 45, 0.61], ["F101T", 46, 0.6], ["R33L", 47, 0.6], ["L81V", 48, 0.6], ["R46A", 49, 0.6], ["E45P", 50, 0.6], [ "..."]
  ],
  "metadata": null,
  "total_mutations": 2698,
  "displayed_mutations": 50,
  "note": "Showing top 50 most beneficial mutations out of 2698 total to avoid long context. Results are separated by '...'.",
  "csv_path": "Zero_shot/HeatMap/mut_res_1771084951.csv",  "heatmap_path": "Zero_shot/HeatMap/mut_map_1771084951.html",
}
}
]}
\end{Verbatim}

\vspace{0.5em}
\hrule
\vspace{0.8em}

\noindent\textbf{Comment:}
For this VHH, I have completed single-point mutation experiments. The experimental data are in the uploaded folder. Please recommend higher-order mutation sites based on these results.
\end{agentstep}

\begin{agentstep}{Scientific Critic}

\noindent\textbf{Response:}
\begin{enumerate}
  \item \textbf{The VenusREM (foldseek-based) model predicts that position 7 is a key mutational hotspot for improving binding affinity.}\\
  The top-ranked mutations are all single substitutions at residue 7 of the VHH sequence.

  \item \textbf{The mutation E7V is the highest-ranked candidate.}\\
  It received the maximum possible prediction score of \textbf{1.0}, indicating the strongest predicted improvement among all 2,698 evaluated mutations.
\end{enumerate}

\noindent\textbf{Supporting Evidence}

\vspace{0.4em}
\noindent\textbf{For Conclusion 1}
\begin{itemize}
  \item The \code{zero\_shot\_structure\_prediction} tool using the \textbf{VenusREM} model evaluated \textbf{2,698} possible single-point mutations.
  \item The top five ranked mutations were:
  \begin{itemize}
    \item E7V
    \item E7I
    \item E7A
    \item E7L
    \item E7T
  \end{itemize}
  \item All five substitutions occur at the \textbf{7th amino acid position} of the VHH antibody.
\end{itemize}

\vspace{0.4em}
\noindent\textbf{For Conclusion 2}
\begin{itemize}
  \item The highest-ranked mutation:
  \begin{itemize}
    \item \code{Mutation: E7V}
    \item \code{Prediction Score: 1.0}
  \end{itemize}
  \item The next best candidate:
  \begin{itemize}
    \item \code{Mutation: E7I}
    \item \code{Prediction Score: 0.89}
  \end{itemize}
  \item E7V achieved the maximum score observed in the ranked mutation list.
\end{itemize}

\noindent\textbf{Rationale}

\vspace{0.4em}
\noindent\textbf{For Conclusion 1}\\
The computational workflow first generated a three-dimensional structural model of the VHH antibody using \textbf{ESMFold}. This predicted structure was subsequently used by the \textbf{VenusREM} model to assess all possible single amino acid substitutions for their potential impact on binding affinity.  

The consistent enrichment of top-ranked mutations at residue 7 indicates that this position is structurally or energetically influential in modulating antibody binding characteristics.

\vspace{0.5em}
\noindent\textbf{For Conclusion 2}\\
The mutation E7V received the maximum prediction score of 1.0, indicating the strongest predicted beneficial effect within the evaluated mutation space.  

Notably, the top substitutions (V, I, A, L, T) involve replacing the negatively charged glutamic acid (E) with predominantly hydrophobic residues, suggesting that modifying local physicochemical properties at position 7 may enhance structural stability or paratope conformation, thereby improving predicted binding affinity.

\noindent\textbf{Confidence \& Caveats}
\begin{itemize}
  \item \textbf{Computational Prediction Only:} These results are derived entirely from in silico structural modeling and zero-shot mutation scoring. Experimental validation is required.
  \item \textbf{Antigen-Agnostic Model:} VenusREM operates without explicit antigen information. Predicted improvements reflect general structural or energetic favorability rather than antigen-specific optimization.
  \item \textbf{Structure Dependency:} Mutation ranking depends on the accuracy of the ESMFold-predicted structure, which has not been experimentally resolved.
\end{itemize}

\noindent\textbf{Practical Recommendations}
\begin{enumerate}
  \item \textbf{Site-Directed Mutagenesis:} Prioritize experimental construction of the highest-ranked variants, beginning with E7V, E7I, and E7A.
  \item \textbf{Binding Affinity Assays:} Express and purify wild-type and mutant VHH proteins. Quantify binding affinity changes using ELISA, Surface Plasmon Resonance (SPR), or Bio-Layer Interferometry (BLI).
  \item \textbf{Structural Inspection:} Examine residue 7 in the predicted structure file (\code{vhh\_antibody\_structure.pdb}) to evaluate whether it resides in a framework region influencing CDR loop positioning or stability.
  \item \textbf{Combinatorial Optimization:} Review the complete mutation ranking and heatmap output to identify additional mutation hotspots for subsequent multi-site engineering strategies.
\end{enumerate}

\end{agentstep}

\begin{humanprompt}
For this VHH, I have completed single-point mutation experiments. The experimental data are in the uploaded folder. Please recommend higher-order mutation sites based on these results.
\end{humanprompt}

\vspace{0.6em}

\begin{agentstep}{Task Planning}

\textbf{Agent:} \textit{Principal Investigator}

\vspace{0.4em}

\textbf{Thought:} \\
The task is to identify a combinatorial mutation recommendation problem aimed at further improving binding affinity based on experimental single-point results. The user has already performed single-point mutation experiments for this VHH and uploaded the measured data. Therefore, the goal of this round is to leverage these experimental signals to predict and prioritize higher-order mutation sites, such as double, triple, or quadruple mutants. This requires building a data-driven model that can learn from the single-point effects, estimate the contribution of mutation combinations, and recommend promising multi-site variants that are likely to achieve stronger affinity improvements.

\vspace{0.4em}

\textbf{Generate Plan:}
\begin{itemize}
	\item Step 1: Build a ridge regression model to recommend combinatorial mutation
\end{itemize}

\noindent\textbf{Comment:}
Decomposes these requirements into executable sub-tasks and assigns them to the Machine Learning Specialist for model training.
\end{agentstep}

\vspace{0.6em}
\begin{agentstep}{Step 1: Build a ridge regression model to recommend combinatorial mutation sites.}
\textbf{Agent: } \textit{Machine Learning Specialist}

\textbf{Thought:} \\
The Machine Learning Specialist leverages the code generation tool to generate task-specific code and execute it within the workflow.
\vspace{0.4em}

\textbf{Invoking:} \\
\begin{Verbatim}[
    fontsize=\small,
    breaklines=true,
    breakanywhere=true,
    breakbytoken=false
]
'ai_code_execution' with {
  'input_files': ['01_42_25/mut_res_1771084951.csv'],
  'task_description': "Please write a Python script to perform the following tasks:
1. Read the CSV file named 'mut_res_1771084951.csv'. The file should contain 'variant' and 'score' columns.
2. Preprocess the data: remove rows with missing values in either the 'variant' or 'score' columns, and ensure that the 'score' column is of numeric type.
3. Use all single-point mutation data as the training set. Apply One-Hot Encoding to the 'variant' column as features X, and use the 'score' column as the target y.
4. Train a Scikit-learn Ridge Regression model.
5. Extract all unique single-point mutations from the dataset.
6. Use itertools.combinations to generate all possible double, triple, and quadruple mutation combinations.
7. For each combination, create its feature vector by summing the one-hot encoded vectors of the included single-point mutations.
8. Use the trained Ridge Regression model to predict the score for each combination.
9. For each mutation category (double, triple, quadruple), sort the combinations in descending order based on predicted scores.
10. Print the top 5 mutation combinations with the highest predicted scores in each category."
}
\end{Verbatim}

\textit{Responded:} I will now call the \texttt{ai\_code\_execution} tool.

\vspace{0.8em}
\hrule
\vspace{0.8em}

\textbf{Output:}
\begin{Verbatim}[
  fontsize=\small,
  breaklines=true,
  breakanywhere=true,
  breaksymbolleft={}
]
{
  "success": true,
  "output_files": [
    
" VenusFactory/01_42_25/trained_models/ridge_model_20260215_014339/metadata.json"
  ],
  "summary": "Successfully trained Ridge model and predicted scores for mutation combinations.",
  "model_info": {
    "name": "ridge_model_20260215_014339",
    "path": " VenusFactory/01_42_25/trained_models/ridge_model_20260215_014339",
    "metrics": {
      "r2_score_train": 0.8277412444214898,
      "rmse_train": 0.2896573462897795
    }
  },
  "details": {
    "top_predicted_combinations": {
      "2_point_mutations": [
        {
          "variant": "A13G,A5G",
          "predicted_score": 2.783035714285714
        },
        {
          "variant": "A5G,T8C",
          "predicted_score": 2.6660714285714286
        },
        {
          "variant": "A5G,T16C",
          "predicted_score": 2.6330357142857137
        },
        {
          "variant": "A13G,T8C",
          "predicted_score": 2.5830357142857143
        },
        {
          "variant": "A5G,C10T",
          "predicted_score": 2.583035714285714
        }
      ],
      "3_point_mutations": [
        {
          "variant": "A13G,A5G,T8C",
          "predicted_score": 3.215178571428571
        },
        {
          "variant": "A13G,A5G,T16C",
          "predicted_score": 3.182142857142857
        },
        {
          "variant": "A13G,A5G,C10T",
          "predicted_score": 3.132142857142857
        },
        {
          "variant": "A13G,A5G,C2T",
          "predicted_score": 3.081845238095238
        },
        {
          "variant": "A5G,T16C,T8C",
          "predicted_score": 3.065178571428571
        }
      ],
      "4_point_mutations": [
        {
          "variant": "A13G,A5G,T16C,T8C",
          "predicted_score": 3.614285714285714
        },
        {
          "variant": "A13G,A5G,C10T,T8C",
          "predicted_score": 3.564285714285714
        },
        {
          "variant": "A13G,A5G,C10T,T16C",
          "predicted_score": 3.5312499999999996
        },
        {
          "variant": "A13G,A5G,C2T,T8C",
          "predicted_score": 3.513988095238095
        },
        {
          "variant": "A13G,A5G,C2T,T16C",
          "predicted_score": 3.480952380952381
        }
      ]
    }
  },
  "generated_code_path": " VenusFactory/01_42_25/generated_scripts/generated_code_20260215_014338_dc27e10a.py"
}
]}
\end{Verbatim}

\vspace{0.5em}
\hrule
\vspace{0.8em}

\textbf{Comment:} \\
Based on the provided experimental data, recommendations for combinatorial multi-site mutations have been successfully derived.
\end{agentstep}

\vspace{0.6em}
\begin{agentstep}{Scientific Critic}

\noindent\textbf{Response:}
\begin{enumerate}
  \item \textbf{A Ridge Regression model was successfully constructed and trained using the provided single-point mutation dataset.}\\
  The model achieved a strong goodness-of-fit on the training data, with a training $R^2$ of \textbf{0.828}, and was subsequently applied to predict the performance of multi-point mutation combinations.

  \item \textbf{The model identified the top-ranked double, triple, and quadruple mutation combinations.}\\
  The best predicted variants were:
  \begin{itemize}
    \item Double mutant: \textbf{A13G,A5G} (predicted score $\sim$2.78)
    \item Triple mutant: \textbf{A13G,A5G,T8C} (predicted score $\sim$3.22)
    \item Quadruple mutant: \textbf{A13G,A5G,T16C,T8C} (predicted score $\sim$3.61)
  \end{itemize}

  \item \textbf{Mutations A13G and A5G consistently appeared across the highest-ranked combinations.}\\
  Their frequent occurrence suggests that they may represent key beneficial mutations driving improved target performance.
\end{enumerate}

\noindent\textbf{Supporting Evidence}

\vspace{0.4em}
\noindent\textbf{For Conclusion 1}
\begin{itemize}
  \item Training logs reported successful model fitting:
  \begin{itemize}
    \item \emph{``Successfully trained Ridge model and predicted scores for mutation combinations.''}
  \end{itemize}
  \item The Ridge model achieved:
  \begin{itemize}
    \item Training $R^2$: \textbf{0.8277}
    \item Training RMSE: \textbf{0.2897}
  \end{itemize}
\end{itemize}

\vspace{0.4em}
\noindent\textbf{For Conclusion 2}
\begin{itemize}
  \item \textbf{Top 5 double mutants:}
  \begin{itemize}
    \item \code{A13G,A5G} (2.783)
    \item \code{A5G,T8C} (2.666)
    \item \code{A5G,T16C} (2.633)
    \item \code{A13G,T8C} (2.583)
    \item \code{A5G,C10T} (2.583)
  \end{itemize}

  \item \textbf{Top 5 triple mutants:}
  \begin{itemize}
    \item \code{A13G,A5G,T8C} (3.215)
    \item \code{A13G,A5G,T16C} (3.182)
    \item \code{A13G,A5G,C10T} (3.132)
    \item \code{A13G,A5G,C2T} (3.082)
    \item \code{A5G,T16C,T8C} (3.065)
  \end{itemize}

  \item \textbf{Top 5 quadruple mutants:}
  \begin{itemize}
    \item \code{A13G,A5G,T16C,T8C} (3.614)
    \item \code{A13G,A5G,C10T,T8C} (3.564)
    \item \code{A13G,A5G,C10T,T16C} (3.531)
    \item \code{A13G,A5G,C2T,T8C} (3.514)
    \item \code{A13G,A5G,C2T,T16C} (3.481)
  \end{itemize}
\end{itemize}

\vspace{0.4em}
\noindent\textbf{For Conclusion 3}
\begin{itemize}
  \item Across all 15 recommended mutation combinations, \textbf{A13G} and \textbf{A5G} consistently appeared in the top-ranked double, triple, and quadruple mutants.
  \item This enrichment highlights their likely role as core beneficial mutations within the mutation landscape.
\end{itemize}

\noindent\textbf{Rationale}

\vspace{0.4em}
The present analysis employed a linear Ridge Regression model under the assumption that the combined effect of multiple mutations can be approximated as the sum of their individual contributions.  

Each single-point mutation was encoded into numerical features using one-hot encoding. The model learned the weight associated with each mutation and its contribution to the overall performance score. The relatively high training $R^2$ value suggests that this additive linear approximation provides a reasonable explanation of the observed single-point mutation effects.

For multi-point variants, feature vectors were constructed by summing the corresponding single-mutation encodings. The trained Ridge model was then applied to predict scores for unseen double, triple, and quadruple combinations, enabling prioritization of theoretically optimal mutation sets.

\noindent\textbf{Confidence \& Caveats}
\begin{itemize}
  \item \textbf{Confidence:} The model achieved strong training performance ($R^2 \approx 0.83$), indicating that it captured dominant linear relationships within the dataset. Predictions may therefore be informative when mutation effects are close to additive.
  \item \textbf{Additivity Assumption:} The model does not explicitly account for epistasis. Synergistic or antagonistic mutation interactions may not be captured, leading to deviations from experimental outcomes.
  \item \textbf{Extrapolation Risk:} The model was trained only on single-point mutation data and extrapolated to multi-point combinations, which has not yet been experimentally validated.
  \item \textbf{Data Limitation:} Predictions are restricted to combinations of mutations present in the original dataset. Mutations outside the single-point library cannot be evaluated.
\end{itemize}

\noindent\textbf{Practical Recommendations}
\begin{enumerate}
  \item \textbf{Experimental Validation:} Construct and experimentally test the highest-ranked predicted variants, especially \code{A13G,A5G,T16C,T8C} and \code{A13G,A5G,T8C}, to verify model accuracy and confirm performance improvements.
  \item \textbf{Assess Epistasis:} Compare measured multi-point mutation outcomes with predicted scores. Large discrepancies may indicate strong non-additive epistatic interactions.
  \item \textbf{Iterative Model Improvement:} If multi-point experimental data become available, incorporate them into training and explore more advanced nonlinear models (e.g., gradient boosting or interaction-aware regression) to better capture epistatic effects.
\end{enumerate}
\end{agentstep}

\subsection*{Case Study 3: Protein Discovery on PETase}
\begin{humanprompt}
Please search the metagenomic databases to discover a novel PET hydrolase that exhibits significantly higher thermostability and catalytic activity than the standard IsPETase.
\end{humanprompt}

\vspace{0.6em}

\begin{agentstep}{Task Planning}

\textbf{Agent:} \textit{Principal Investigator}

\vspace{0.4em}

\textbf{Thought:} \\
 The task is enzyme discovery. The user wants to identify naturally occurring PETase enzymes that are both thermostable and highly soluble. Therefore, the goal is to search protein databases for candidate PETase homologs and prioritize sequences with desirable stability and solubility properties. The first step is to frame this as a database-driven enzyme mining problem, which requires retrieving PETase-like enzymes from large sequence or structure repositories, analyzing their predicted thermostability and solubility, and selecting promising candidates for further experimental validation.
\vspace{0.4em}

\textbf{Generate Plan:}
\begin{itemize}
	\item Step 1: Search for scientific literature on PETase to provide an overview and background information.
    \item Step 2: Retrieve the protein sequence and information for the well-studied PETase from Ideonella sakaiensis using its UniProt ID.
    \item Step 3: Download the predicted protein structure for PETase from the AlphaFold database using its UniProt ID.
    \item Step 4: Conduct enzyme mining using VenusMine.
    \item Step 5: Screen candidates for thermostability.
\item Step 6: Screen candidates for solubility.
\item Step 7: Rank the surviving candidates and compile the final enzyme discovery report
\end{itemize}
\end{agentstep}

\vspace{0.6em}
\begin{agentstep}{Step 1: Search for scientific literature on PETase to provide an overview and background information.}
\textbf{Agent:} \textit{Computational Biologist}
\vspace{0.4em}

\textbf{Thought:} \\
The Computational Biologist invokes the Deep Search tool to perform literature retrieval and analysis.
\vspace{0.4em}
\textbf{Invoking:} \\
\texttt{' `literature\_search` with `{'max\_results': 5, 'query': 'PETase protein', 'source': 'pubmed'}`\}}\\
\vspace{0.2em}

\textit{Responded:} I will now call the 
\texttt{literature\_search} tool.

\vspace{0.8em}
\hrule
\vspace{0.8em}

\textbf{Output:}
\vspace{0.3em}

\begin{Verbatim}[
   fontsize=\small, 
   breaklines=true,
   breakanywhere=true,
   breaksymbolleft={}
]
{
  "success": true,
  "references": [
    {
      "source": "pubmed",
      "title": "A novel LPS-dependent outer membrane-anchoring mechanism for T9SS substrates enables engineered enzyme display and whole-cell PET degradation in ",
      "url": "https://pubmed.ncbi.nlm.nih.gov/41631190/",
      "authors": [],
      "year": "2026-01-01",
      "abstract": "Robust surface display systems are crucial for engineering Gram-negative ... "
    },
    {
      "source": "pubmed",
      "title": "Engineering ",
      "url": "https://pubmed.ncbi.nlm.nih.gov/41630917/",
      "authors": [],
      "year": "2026-01-01",
      "abstract": "Deep computationally guided protein design enabled the introduction of a Ser-His-Asp ... "
    },
    {
      "source": "pubmed",
      "title": "Structural and functional characterization of dsPETase05 for the degradation of Polyethylene terephthalate.",
      "url": "https://pubmed.ncbi.nlm.nih.gov/41548779/",
      "authors": [],
      "year": "2026-01-01",
      "abstract": "The enzymatic degradation of polyethylene terephthalate (PET) represents a promising sustainable strategy to address global..."
    },
    {
      "source": "pubmed",
      "title": "Rapid Identification of Superior Endogenous Signal Peptides for Heterologous Protein Secretion by Corynebacterium glutamicum Through Modular Cloning and Automation.",
      "url": "https://pubmed.ncbi.nlm.nih.gov/41543490/",
      "authors": [],
      "year": "2026-01-01",
      "abstract": "Secretory protein production by microbial hosts simplifies product recovery and is therefore preferred over intracellular..."
    },
    {
      "source": "pubmed",
      "title": "The effect of ionic strength on PETase enzymes: An experimental and computational study.",
      "url": "https://pubmed.ncbi.nlm.nih.gov/41454787/",
      "authors": [],
      "year": "2026-01-01",
      "abstract": "Over recent decades, various enzymes capable of breaking down polyethylene terephthalate (PET) have emerged as..."
    }
  ]
}
\end{Verbatim}

\vspace{0.5em}
\hrule
\vspace{0.8em}
\noindent\textbf{Comment:}
Based on the retrieved literature, IsPETase is selected as the seed enzyme for PETase mining and homolog discovery. 

\end{agentstep}

\vspace{0.6em}
\begin{agentstep}{Step 2: Retrieve the protein sequence and information for the well-studied PETase from \textit{Ideonella sakaiensis} using its UniProt ID.}

\textbf{Agent:} \textit{Computational Biologist}
\vspace{0.4em}

\textbf{Thought:} \\
IsPETase has the UniProt ID A0A0K8P6T7. The Computational Biologist invokes the UniProt Sequence tool to retrieve and download the corresponding protein sequence.
\vspace{0.4em}

\textbf{Invoking:} \texttt{'UniProt\_query' with \{'uniprot\_id': 'A0A0K8P6T7'\}}
\vspace{0.2em}

\textit{Responded:} I will now call the \texttt{UniProt\_query} tool.

\vspace{0.8em}
\hrule
\vspace{0.8em}

\textbf{Output:}
\vspace{0.3em}
\begin{Verbatim}[
  fontsize=\small,
  breaklines=true,
  breakanywhere=true,
  breaksymbolleft={}
]
{'success': True,
 'uniprot_id': 'A0A0K8P6T7',
 'sequence': 'MNFPRASRLMQAAVLGGLMAVSAAATAQTNPYARGPNPTAASLEASAGPFTVRSFTVSRPSGYGAGTVYYPTNAGGTVGAIAIVPGYTARQSSIKWWGPRLASHGFVVITIDTNSTLDQPSSRSSQQMAALRQVASLNGTSSSPIYGKVDTARMGVMGWSMGGGGSLISAANNPSLKAAAPQAPWDSSTNFSSVTVPTLIFACENDSIAPVNSSALPIYDSMSRNAKQFLEINGGSHSCANSGNSNQALIGKKGVAWMKRFMDNDTRYSTFACENPNSTRVSDFRTANCS'}
\end{Verbatim}

\vspace{0.5em}
\hrule
\vspace{0.8em}

\textbf{Comment:} \\
The protein sequence corresponding to UniProt ID A0A0K8P6T7 is retrieved from UniProt and used as input for downstream tool execution.
\end{agentstep}

\vspace{0.6em}
\begin{agentstep}{Step 3: Download the predicted protein structure for PETase from the AlphaFold database using its UniProt ID.}

\textbf{Agent:} \textit{Computational Biologist}
\vspace{0.4em}

\textbf{Thought:} \\
The Computational Biologist invokes the AlphaFold Structure tool to retrieve the predicted protein structure from the AlphaFold Database based on the given UniProt ID.
\vspace{0.4em}

\textbf{Invoking:} \\
\texttt{'alphafold\_structure\_download' with \{'uniprot\_id': 'A0A0K8P6T7'\}}
\vspace{0.2em}

\textit{Responded:} I will now call the \texttt{alphafold\_structure\_download} tool.

\vspace{0.8em}
\hrule
\vspace{0.8em}

\textbf{Output:}
\vspace{0.3em}
\begin{Verbatim}[    
  fontsize=\small,
  breaklines=true,
  breakanywhere=true,
  breaksymbolleft={}
]
{"success": true,
 "uniprot_id": "A0A0K8P6T7",
 "format": "pdb",
 "pdb_file": "temp_outputs\\2026\\02\\15\\Download_Data\\AlphaFold\\A0A0K8P6T7.pdb",
 "confidence_info": {
   "mean_confidence": 94.21,
   "min_confidence": 42.47,
   "max_confidence": 98.94,
   "high_confidence_residues": 263,
   "total_residues": 290
 },
 "message": "AlphaFold structure downloaded successfully and saved to: temp_outputs\\2026\\02\\15\\Download_Data\\AlphaFold\\A0A0K8P6T7.pdb",
 "script_output": "A0A0K8P6T7 successfully downloaded\n"
}
\end{Verbatim}

\vspace{0.5em}
\hrule
\vspace{0.8em}

\textbf{Comment:} \\
The retrieved protein structure is used as the initial seed input for VenusMine.

\end{agentstep}

\vspace{0.6em}
\begin{agentstep}{Step 4: Conduct enzyme mining using VenusMine.}
\textbf{Agent:} \textit{Computational Biologist}
\vspace{0.4em}

\textbf{Thought:} \\
The Computational Biologist invokes the structure-driven enzyme mining and engineering tool, VenusMine, to discover new candidate enzymes.
\vspace{0.4em}

\textbf{Invoking:} \\
\begin{Verbatim}[breaklines=true,breakanywhere=true]
'enzyme_mine_VenusMine' with {
  'pdb_file': 'temp_outputs\2026\02\15\Download_Data\AlphaFold\A0A0K8P6T7.pdb',
  'protect_end': -1,
  'protect_start': 1
}
\end{Verbatim}
\vspace{0.2em}

\textit{Responded:} I will now call the \texttt{enzyme\_mine\_VenusMine} tool.

\vspace{0.8em}
\hrule
\vspace{0.8em}

\textbf{Output:}
\vspace{0.3em}
\begin{Verbatim}[
  fontsize=\small,
  breaklines=true,
  breakanywhere=true,
  breaksymbolleft={}
]
{
  "success": true,
  "tree_plot": "temp_output/venusmine/venusmine_20260215_115531_828/output/venusmine_516b2b91-8ab3-4a30-9008-df64362a5795/VenusMine_Run_20260215_120019/phylogenetic_tree.png",
  "candidates_fasta": "temp_output/venusmine/venusmine_20260215_115531_828/output/venusmine_516b2b91-8ab3-4a30-9008-df64362a5795/VenusMine_Run_20260215_120019/candidate_sequences.fasta"
}
\end{Verbatim}

\vspace{0.5em}
\hrule
\vspace{0.8em}

\textbf{Comment:} \\
The \texttt{candidates\_fasta} file generated by VenusMine contains all mined candidate sequences. These proteins are then further filtered using a protein function predictor.
\end{agentstep}

\vspace{0.6em}
\begin{agentstep}{Step 5: Screen candidates for thermostability.}
\textbf{Agent:} \textit{Computational Biologist}
\vspace{0.4em}

\textbf{Thought:} \\
The Computational Biologist applies a sequence-based thermostability prediction model to evaluate the thermal stability of each candidate enzyme obtained from VenusMine. This step ensures that only enzymes with high predicted stability at elevated temperatures are retained.
\vspace{0.4em}

\textbf{Invoking:} \\
\begin{Verbatim}[fontsize=\small,breaklines=true,breakanywhere=true]
'predict_protein_function' with {
  'fasta_file': 'temp_output/venusmine/venusmine_20260215_115531_828/output/venusmine_516b2b91-8ab3-4a30-9008-df64362a5795/VenusMine_Run_20260215_120019/candidate_sequences.fasta',
  'task': 'Stability',
  'model_name': 'ESM2-650M'
}
\end{Verbatim}
\vspace{0.2em}

\textit{Responded:} I will now call the \texttt{predict\_protein\_function} tool.

\vspace{0.8em}
\hrule
\vspace{0.8em}

\textbf{Output:}
\vspace{0.3em}
\begin{Verbatim}[
  fontsize=\small,
  breaklines=true,
  breakanywhere=true,
  breaksymbolleft={}
]
{
  'headers': ['Protein Name', 'Dataset', 'Predicted Class'],
  'data': [
    ['A0ACB8U1W3|type:discovered|cluster:1', 'Thermostability', 61.47],
    ['A0A9P7FY65|type:discovered|cluster:1', 'Thermostability', 61.02],
    ['A0A0C9TA35|type:discovered|cluster:1', 'Thermostability', 60.75],
    ['A0A8H3B615|type:discovered|cluster:1', 'Thermostability', 58.92],
    ['A0A067MIJ2|type:discovered|cluster:1', 'Thermostability', 58.87],
    ['A0A5C3QRD1|type:discovered|cluster:1', 'Thermostability', 58.85],
    ['A0ABP0ZJV0|type:discovered|cluster:1', 'Thermostability', 55.74],
    ['A0AAI9WZN8|type:discovered|cluster:1', 'Thermostability', 55.38],
    ['A5DWM8|type:discovered|cluster:1', 'Thermostability', 55.09],
    ['G8B8H1|type:discovered|cluster:1', 'Thermostability', 54.36],
    ...
  ],
  'metadata': None,
  'csv_path': 'temp_output/2026/2/15/Protein_Function/sequences_1773917172.csv'
}
\end{Verbatim}

\vspace{0.5em}
\hrule
\vspace{0.8em}

\textbf{Comment:} \\
Prediction completed. The prediction results for the first 10 proteins are shown.

\end{agentstep}

\vspace{0.6em}
\begin{agentstep}{Step 6: Screen candidates for solubility.}
\textbf{Agent:} \textit{Computational Biologist}
\vspace{0.4em}

\textbf{Thought:} \\
The Computational Biologist applies a sequence-based solubility prediction model to evaluate the solubility of each candidate enzyme obtained from VenusMine. This step ensures that only enzymes with high predicted solubility are retained.
\vspace{0.4em}

\textbf{Invoking:}
\begin{Verbatim}[fontsize=\small,breaklines=true,breakanywhere=true]
'predict_protein_function' with {
  'fasta_file': 'temp_output/venusmine/venusmine_20260215_115531_828/output/venusmine_516b2b91-8ab3-4a30-9008-df64362a5795/VenusMine_Run_20260215_120019/candidate_sequences.fasta',
  'task': 'Solubility',
  'model_name': 'ESM2-650M'
}
\end{Verbatim}
\vspace{0.2em}

\textit{Responded:} I will now call the \texttt{predict\_protein\_function} tool.

\vspace{0.8em}
\hrule
\vspace{0.8em}

\textbf{Output:}
\vspace{0.3em}
\begin{Verbatim}[
  fontsize=\small,
  breaklines=true,
  breakanywhere=true,
  breaksymbolleft={}
]
{
  'headers': ['Protein Name', 'Predicted Class', 'Confidence Score'],
  'data': [
    ['G8B8H1|type:discovered|cluster:1', 'Soluble', 0.6562981009483337],
    ['A0ACB8U1W3|type:discovered|cluster:1', 'Soluble', 0.5969512561957041],
    ['A0A8H3B615|type:discovered|cluster:1', 'Soluble', 0.5639507174491882],
    ['A5DWM8|type:discovered|cluster:1', 'Soluble', 0.5602330168088278],
    ['A0A5C3QRD1|type:discovered|cluster:1', 'Insoluble', 0.5404851635297140],
    ['A0A067MIJ2|type:discovered|cluster:1', 'Soluble', 0.5331143438816071],
    ['A0ABP0ZJV0|type:discovered|cluster:1', 'Soluble', 0.5227015018463135],
    ['A0A9P7FY65|type:discovered|cluster:1', 'Insoluble', 0.5200452109177908],
    ['A0AAI9WZN8|type:discovered|cluster:1', 'Soluble', 0.5057884752750397],
    ['A0A0C9TA35|type:discovered|cluster:1', 'Soluble', 0.5057393014431000],
    ...
  ],
  'metadata': None,
  'csv_path': 'temp_output/2026/2/15/Protein_Function/sequences_1773917814.csv'
}
\end{Verbatim}

\vspace{0.5em}
\hrule
\vspace{0.8em}

\textbf{Comment:} Prediction completed. The prediction results for the first 10 proteins are shown.
\end{agentstep}

\vspace{0.6em}
\begin{agentstep}{Step 7: Rank the surviving candidates and compile the final enzyme discovery report.}
\textbf{Agent:} \textit{Machine Learning Specialist}
\vspace{0.4em}

\textbf{Thought:} \\
The Machine Learning Specialist writes a short Python script to post-process the prediction outputs. Specifically, the script sorts the thermostability and solubility prediction results by the third column, saves the sorted results as two new CSV files, extracts the top three candidates from each file, and compiles a final enzyme discovery summary.
\vspace{0.4em}

\textbf{Invoking:}
\begin{Verbatim}[fontsize=\small,breaklines=true,breakanywhere=true]
'ai_code_execution' with {
  'input_files': [
    'temp_output/2026/2/15/Protein_Function/sequences_1773917172.csv',
    'temp_output/2026/2/15/Protein_Function/sequences_1773917814.csv'
  ],
  'task_description': "Please write a Python script to perform the following tasks:
1. Read the thermostability prediction CSV file and the solubility prediction CSV file.
2. Sort each CSV by the third column in descending order.
3. Save the sorted results as two new CSV files.
4. Extract the top 3 entries from each sorted CSV.
5. Compile a final enzyme discovery summary table containing the selected candidates."
}
\end{Verbatim}
\vspace{0.2em}

\textit{Responded:} I will now call the \texttt{ai\_code\_execution} tool.

\vspace{0.8em}
\hrule
\vspace{0.8em}

\textbf{Output:}
\begin{Verbatim}[
  fontsize=\small,
  breaklines=true,
  breakanywhere=true,
  breaksymbolleft={}
]
{
  "success": true,
  "output_files": [
    "temp_output/2026/2/15/Protein_Function/sequences_1773917172_sorted.csv",
    "temp_output/2026/2/15/Protein_Function/sequences_1773917814_sorted.csv"
  ],
  "summary": "Successfully sorted the thermostability and solubility prediction files, extracted the top 3 candidates from each file, and compiled the final enzyme discovery summary.",
  "details": {
    "top3_thermostability": [
      {
        "candidate_id": "A0ACB8U1W3",
        "sequence_type_cluster": "discovered (Cluster 1)",
        "predicted_thermostability": 61.47,
      },
      {
        "candidate_id": "A0A9P7FY65",
        "sequence_type_cluster": "discovered (Cluster 1)",
        "predicted_thermostability": 61.02,
      },
      {
        "candidate_id": "A0A1W5DEF2",
        "sequence_type_cluster": "discovered (Cluster 2)",
        "predicted_thermostability": 60.84,
      }
    ],
    "top3_solubility": [
      {
        "candidate_id": "A0A835FLS2",
        "sequence_type_cluster": "discovered (Cluster 21)",
        "predicted_solubility": "Soluble",
        "confidence_score": 0.950,
      },
      {
        "candidate_id": "A0A1A8D432",
        "sequence_type_cluster": "discovered (Cluster 5)",
        "predicted_solubility": "Soluble",
        "confidence_score": 0.920,
      },
      {
        "candidate_id": "M1EKL4",
        "sequence_type_cluster": "discovered (Cluster 5)",
        "predicted_solubility": "Soluble",
        "confidence_score": 0.919,
      }
    ],
    "final_summary_table": [
      {
        "Candidate ID": "A0A835FLS2",
        "Sequence Type & Cluster": "discovered (Cluster 21)",
        "Predicted Solubility": "Soluble",
        "Confidence Score": 0.950,
        "Predicted Thermostability (Tm / °C)": "-",
      },
      {
        "Candidate ID": "A0A1A8D432",
        "Sequence Type & Cluster": "discovered (Cluster 5)",
        "Predicted Solubility": "Soluble",
        "Confidence Score": 0.920,
        "Predicted Thermostability (Tm / °C)": "-",
      },
      {
        "Candidate ID": "M1EKL4",
        "Sequence Type & Cluster": "discovered (Cluster 5)",
        "Predicted Solubility": "Soluble",
        "Confidence Score": 0.919,
        "Predicted Thermostability (Tm / °C)": "-",
      },
      {
        "Candidate ID": "A0ACB8U1W3",
        "Sequence Type & Cluster": "discovered (Cluster 1)",
        "Predicted Solubility": "Insoluble (tendency)",
        "Confidence Score": "-",
        "Predicted Thermostability (Tm / °C)": 61.47,
      },
      {
        "Candidate ID": "A0A9P7FY65",
        "Sequence Type & Cluster": "discovered (Cluster 1)",
        "Predicted Solubility": "Insoluble (tendency)",
        "Confidence Score": "-",
        "Predicted Thermostability (Tm / °C)": 61.02,
      },
      {
        "Candidate ID": "A0A1W5DEF2",
        "Sequence Type & Cluster": "discovered (Cluster 2)",
        "Predicted Solubility": "Insoluble (tendency)",
        "Confidence Score": "-",
        "Predicted Thermostability (Tm / °C)": 60.84,
      }
    ]
  },
  "generated_code_path": "temp_output/2026/2/15/generated_scripts/rank_and_compile_candidates.py"
}
\end{Verbatim}

\vspace{0.5em}
\hrule
\vspace{0.8em}
\textbf{Comment:} The Machine Learning Specialist sorted the two prediction CSV files by the third column, saved the sorted outputs, extracted the top three candidates from each result file, and compiled the final enzyme discovery report by integrating the most promising soluble and thermostable enzymes.
\end{agentstep}

\begin{agentstep}{Scientific Critic}

\noindent\textbf{Response:}
\begin{enumerate}
  \item \textbf{The enzyme mining workflow successfully identified multiple candidate PETase-like enzymes from VenusMine, starting from the known IsPETase structure as the seed.}\\
  The workflow retrieved the PETase sequence and AlphaFold-predicted structure, performed structure-driven mining using VenusMine, and generated a candidate sequence set for downstream screening.

  \item \textbf{The final shortlisted candidates exhibit two complementary desirable properties: high predicted solubility and high predicted thermostability.}\\
  In particular, \code{A0A835FLS2}, \code{A0A1A8D432}, and \code{M1EKL4} were prioritized for solubility, whereas \code{A0ACB8U1W3}, \code{A0A9P7FY65}, and \code{A0A1W5DEF2} were prioritized for thermostability.
\end{enumerate}

\noindent\textbf{Supporting Evidence}

\vspace{0.4em}
\noindent\textbf{For Conclusion 1}
\begin{itemize}
  \item The workflow used the experimentally studied PETase from \textit{Ideonella sakaiensis} with UniProt ID \code{A0A0K8P6T7} as the starting point.
  \item The corresponding protein sequence was retrieved from UniProt.
  \item The AlphaFold-predicted structure for \code{A0A0K8P6T7} was downloaded successfully, with a reported mean confidence of \textbf{94.21}.
  \item The predicted structure was used as the seed input for \textbf{VenusMine}.
  \item VenusMine produced:
  \begin{itemize}
    \item a phylogenetic tree file (\code{tree\_plot})
    \item a mined candidate FASTA file (\code{candidates\_fasta})
  \end{itemize}
  \item The mined candidate FASTA file was subsequently used for downstream functional screening.
\end{itemize}

\vspace{0.4em}
\noindent\textbf{For Conclusion 2}
\begin{itemize}
  \item \textbf{Top solubility candidates:}
  \begin{itemize}
    \item \code{A0A835FLS2} --- Soluble, confidence score \textbf{0.950}
    \item \code{A0A1A8D432} --- Soluble, confidence score \textbf{0.920}
    \item \code{M1EKL4} --- Soluble, confidence score \textbf{0.919}
  \end{itemize}

  \item \textbf{Top thermostability candidates:}
  \begin{itemize}
    \item \code{A0ACB8U1W3} --- predicted thermostability \textbf{61.47}
    \item \code{A0A9P7FY65} --- predicted thermostability \textbf{61.02}
    \item \code{A0A1W5DEF2} --- predicted thermostability \textbf{60.84}
  \end{itemize}

  \item These candidates were selected by sorting the two screening result files and extracting the top three entries from each category.
\end{itemize}

\noindent\textbf{Rationale}

\vspace{0.4em}
\noindent\textbf{For Conclusion 1}\\
The workflow was designed as a structure-guided enzyme mining pipeline. First, a well-characterized PET-degrading enzyme, IsPETase, was selected as the reference protein. Its sequence was retrieved from UniProt, and its predicted structure was downloaded from the AlphaFold database. This structure served as the seed input to \textbf{VenusMine}, which searched for structurally related candidate enzymes and returned a FASTA file containing candidate sequences.

\vspace{0.5em}
\noindent\textbf{For Conclusion 2}\\
Because mined candidates may differ substantially in experimental tractability, the workflow further filtered them using sequence-based developability predictors. Two complementary criteria were considered: \textbf{solubility} and \textbf{thermostability}.  

The top solubility candidates are promising for protein expression and purification, whereas the top thermostability candidates are promising for robustness under elevated temperatures. Taken together, these results provide a prioritized shortlist for downstream characterization and engineering.

\noindent\textbf{Confidence \& Caveats}
\begin{itemize}
  \item \textbf{Computational Screening Only:} All identified candidates were ranked using in silico mining and sequence-based prediction models. Experimental validation is still required.
  \item \textbf{Seed Dependence:} VenusMine results depend on the initial PETase structure used as the mining seed.
  \item \textbf{Predicted Rather Than Measured Properties:} Solubility and thermostability scores are model-based estimates, not experimentally measured quantities.
  \item \textbf{Function Not Yet Confirmed:} Structural similarity and predicted developability do not guarantee PET hydrolysis activity. Functional assays are necessary.
\end{itemize}

\noindent\textbf{Practical Recommendations}
\begin{enumerate}
  \item \textbf{Experimental Validation:} Prioritize cloning and expression of the top-ranked soluble and thermostable candidates, especially \code{A0A835FLS2}, \code{A0A1A8D432}, \code{M1EKL4}, \code{A0ACB8U1W3}, and \code{A0A9P7FY65}.
  \item \textbf{Expression and Purification Testing:} Evaluate soluble expression in a heterologous host and compare yield and purification behavior across candidates.
  \item \textbf{Thermal Stability Assays:} Measure apparent melting temperature or residual activity after heat treatment to validate the predicted thermostability ranking.
  \item \textbf{PET Degradation Assays:} Perform activity assays on PET or PET-like substrates to determine whether the mined candidates retain or improve PET-degrading function.
  \item \textbf{Further Optimization:} Use the shortlisted candidates as templates for subsequent protein engineering, including mutagenesis, activity optimization, and multi-objective screening.
\end{enumerate}

\end{agentstep}

\end{appendices}


\end{document}